\documentclass[10pt,twocolumn,journal]{IEEEtran}

\usepackage{epsfig}
\usepackage{epstopdf}
\usepackage{citesort}
\usepackage{amsmath}
\usepackage{amssymb}
\usepackage{color}
\usepackage{array}
\usepackage{multirow}
\usepackage{algorithm}
\usepackage{algorithmic}
\usepackage{rotating}
\usepackage{graphicx}
\usepackage{subfigure}
\usepackage[colorlinks,linkcolor=blue]{hyperref}
\newlength{\figurewidth}
\newlength{\smallfigurewidth}

\begin{document}

\title
{
Cross-domain Hyperspectral Image Classification based on Bi-directional Domain Adaptation
}

\author{%
Yuxiang Zhang,
Wei Li,~\IEEEmembership{Senior Member,~IEEE},
Wen Jia, 
Mengmeng Zhang,\\
Ran Tao,~\IEEEmembership{Senior Member,~IEEE},
Shunlin Liang,~\IEEEmembership{Fellow,~IEEE}
\thanks{%
	This work was supported in part by funding of the National Natural Science Foundation of China (Grant No. 42101403 and 6247012790), in part by the Beijing Natural Science Foundation (Grant No. 4232013), and in part by the National Key Research and Development Program of China (Grant No. 2023YFD2200804 and 2017YFD0600404). This work was conducted in the JC STEM Lab of Quantitative Remote sensing funded by The Hong Kong Jockey Club Charities Trust. (corresponding author: Wei Li; liwei089@ieee.org).  }
\thanks{%
	Y. Zhang and S. Liang are with the Jockey Club STEM Laboratory of Quantitative Remote Sensing, Department of Geography, the University of Hong Kong, Hong Kong, China (e-mail: yxzhang7@hku.hk, shunlin@hku.hk).
}
\thanks{%
W. Li, M. Zhang and R. Tao are with the School of Information and Electronics, Beijing Institute of Technology, and Beijing Key Laboratory of Fractional Signals and Systems, 100081 Beijing, China (e-mail: liwei089@ieee.org, mengmengzhang@bit.edu.cn, rantao@bit.edu.cn).
}
\thanks{%
	Wen Jia is with Institute of Forest Resource Information Techniques, Chinese Academy of Forestry, and Key Laboratory of Forestry Remote Sensing and Information System, National Forestry and Grassland Administration, Beijing 100091, China (e-mail: jiawen@ifrit.ac.cn)}
}

\maketitle
\thispagestyle{empty}
\pagestyle{empty}


\begin{abstract}
Utilizing hyperspectral remote sensing technology enables the extraction of fine-grained land cover classes. Typically, satellite or airborne images used for training and testing are acquired from different regions or times, where the same class has significant spectral shifts in different scenes. In this paper, we propose a Bi-directional Domain Adaptation (BiDA) framework for cross-domain hyperspectral image (HSI) classification, which focuses on extracting both domain-invariant features and domain-specific information in the independent adaptive space, thereby enhancing the adaptability and separability to the target scene. In the proposed BiDA, a triple-branch transformer architecture (the source branch, target branch, and coupled branch) with semantic tokenizer is designed as the backbone. Specifically, the source branch and target branch independently learn the adaptive space of source and target domains, a Coupled Multi-head Cross-attention (CMCA) mechanism is developed in coupled branch for feature interaction and inter-domain correlation mining. Furthermore, a bi-directional distillation loss is designed to guide adaptive space learning using inter-domain correlation. Finally, we propose an Adaptive Reinforcement Strategy (ARS) to encourage the model to focus on specific generalized feature extraction within both source and target scenes in noise condition. Experimental results on cross-temporal/scene airborne and satellite datasets demonstrate that the proposed BiDA performs significantly better than some state-of-the-art domain adaptation approaches. In the cross-temporal tree species classification task, the proposed BiDA is more than 3\%$\sim$5\% higher than the most advanced method. The codes will be available from the website: \href{https://github.com/YuxiangZhang-BIT/IEEE_TCSVT_BiDA}{https://github.com/YuxiangZhang-BIT/IEEE\_TCSVT\_BiDA}.
\end{abstract}


\section{Introduction}
Hyperspectral images (HSIs) with high spectral resolution and rich spatial information use subtle spectral information to distinguish different materials. It has been widely used in many fields, including but not limited to resource inventories, analyzing urban living environments, monitoring and evaluating biodiversity. HSI classification is one of the key techniques in remote sensing image interpretation, and has achieved a series of remarkable achievements in recent years \cite{xie2020multiscale,gong2021deep,10520291}. The HSI single-scene classification assumes that the training and test data follow the same distribution, meaning the current scene is constrained \cite{10495370,10341254}. However, in practical tasks, the scene to be predicted is often uncertain, leading to insufficient adaptability of single-scene classification models to new scenes. For example, investigators conduct field surveys over large-scale areas (100 $km^2$) and collect airborne imagery using airborne imaging systems. Typically, the airborne push-broom hyperspectral imager are used to obtain airborne HSIs. The data collection for large-scale areas often requires multiple flight lines of airborne HSI, resulting in a collection process spanning several days or weeks. This implies that the acquired data may vary in terms of illumination (i.e., clear or partially cloudy conditions), resulting in a significant difference in spectral characteristics of the same land cover classes, the phenomenon known as spectral shift \cite{zhang2023single,li2024scformer,10659915}. When land cover classes labels are available only for a specific local region (source domain, SD), existing HSI single-scene classification methods trained on the SD and transferred to other regions (target domain, TD) suffer from high generalization errors, leading to poor interpretability performance. Due to differences in the collection region or acquisition time between training and testing data, it is challenging to meet the assumption of independent and identically distributed data. This challenge is referred to as the cross-scene or cross-temporal classification task, where the training data and testing data correspond to the labeled SD and the unlabeled TD, respectively. The objective of this task is to transfer shared knowledge from the SD to the TD through transfer learning, enabling the classification of the TD across different regions or time periods.

Benefiting from the latest advances in deep learning, deep domain adaptation (DA) methods have been applied to cross-scene tasks \cite{10804219,10742340,10726587,li2025multiscale}. At present, most DA methods design an adaptive layer to complete the adaptation the SD and the TD, which makes the data distribution between domains closer and improves the classification effect of the model on the TD. To solve the feature heterogeneous of multimodal data, Xu et al. \cite{Xu2024} developed invariant risk minimization (IRM) into multimodal classification for the first time, which can effectively improve the generalization ability of classification model. In the aspect of cross-domain ship detection, Zhang et al. \cite{10962255} proposed a SAR image cross-sensor target detection method based on dynamic feature discrimination and center-aware calibration. Zhou et al. \cite{10579867} proposed a dual-stream branch feature alignment extraction network with weight sharing, which can achieve knowledge extraction and sharing between the optical and SAR domains. In the aspect of HSI cross-scene classification, Zhou et al. utilized a deep convolutional recursive neural network to extract discriminative features from the source and target domains, and mapped the features of each layer to the shared subspace transformed by each layer for layer-wise feature alignment \cite{zhou2018deep}. Wang et al. proposed a DA method based on manifold embedding space to learn discriminative information between domains \cite{wang2019domain}. Liu et al. designed a domain-invariant feature generation approach based on generative adversarial strategies \cite{liu2020class}. The method involves adversarial learning between a feature extractor and multiple domain discriminators to generate domain-invariant features. Qu et al. introduced a HSI classification method based on physically constrained shared abundance space \cite{qu2021physically}. By projecting data from the SD and the TD onto a shared abundance space according to their respective physical characteristics, the approach mitigates spectral shift between domains. In order to achieve cross-scene wetland mapping, Huang et al. proposed a spatial-spectral weighted adversarial domain adaptation (SSWADA) network \cite{huang2023cross}. 
Considering the variability of domain shift from different SDs to TD, Ding et al. proposed Consistency-Aware Customized Learning (CACL)\cite{10659915}, which leverages adversarial training to achieve domain-level alignment. The spectral spatial prototypes are dynamically extracted from SD and TD, and the prototype labels are assigned to TD samples based on cosine similarity, so as to achieve fine-grained class-level joint distribution alignment. To focus on both the global domain structure of SD and TD as well as the subdomain structure within each class, Feng et al. proposed Pseudo-Label-Assisted Subdomain Adaptation (PASDA) \cite{10309248}. Based on subdomain alignment, high-quality pseudo-labeled samples are selected from the TD, and a reweighted pruning label propagation strategy is designed to re-weight the outputs of the TD. Huang et al. proposed adversarial DA framework based on calibrated prototype and dynamic instance convolution (CPDIC). While aligning the domain distribution, this method pays attention to the class separability of the aligned target features and the information of the intra-domain samples \cite{10497695}. Cai et al. proposed the multilevel unsupervised domain adaptation (MLUDA) framework, which includes image-level, feature-level, and logic-level alignment between domains to fully explore comprehensive spatial-spectral information \cite{10543066}. Fang et al., considering the difficulty of extracting semantic information from unlabeled TD, introduced the Masked Self-distillation Domain Adaptation (MSDA) \cite{10620320}. This method enhances the discriminability of features by integrating masked self-distillation into DA.


Existing DA-based HSI cross-scene/temporal classification methods rely on unidirectional adaptation from SD to TD, utilizing shared feature extractors and alignment strategies to forcibly project SD and TD into a shared adaptive space. These methods have never considered the difficulty in obtaining an optimal solution for the shared adaptive space in the presence of significant spectral shifts, lacking a bi-directional DA strategy for independent learning of adaptive spaces. Furthermore, while focusing primarily on extracting inter-domain invariant representations, these approaches often overlook the importance of capturing generalized intra-domain features. As a result, even if spectral shift is reduced, the inter-class separability of the learned TD features remains low, leading to suboptimal adaptability to TD data. This limitation is especially pronounced in fine-grained cross-domain classification tasks where inter-class spectral similarity is high.

To address the above issues, we propose a Bi-directional Domain Adaptation (BiDA) framework with a triple-branch architecture (the source branch, target branch, and coupled branch) for cross-domain classification from airborne and satellite HSIs. Different from the token calculation method in Vision Transformer (ViT), BiDA constructs semantic tokens for SD and TD using spatial-spectral characteristics. In the coupled branch, a Coupled Multi-head Cross Attention (CMCA) is designed, where source tokens and target tokens serve as queries to accomplish bi-directional cross-attention, perceiving inter-domain correlations. Additionally, a bi-directional distillation loss is devised to leverage the coupled branch in guiding the training of both the source branch and target branch. Finally, based on the teacher-student structure, an Adaptability Reinforcement Strategy (ARS) is designed to improve the ability of BiDA to extract intra-domain generalized features. The main contributions of this paper are as follows:

\begin{itemize}

\item BiDA framework is proposed, in which the source branch and target branch of encoder learn the adaptive space independently, and the CMCA for the bi-directional inter-domains correlations is proposed in coupled branch to realize the bi-directional adaptation of the feature layer. 

\item Designing bi-directional distillation loss by treating the predicted probability distributions of the SD and TD output by the coupled branch as soft labels. This approach achieves bi-directional supervision for the source and target branches, reinforcing the extraction of independent adaptive spaces.

\item ARS suitable for DA models is designed. The intra-domain consistency constraints are performed on SD and TD respectively, and the source branch and the target branch are encouraged to learn the intra-domain generalized features in the noise data distribution.
\end{itemize}

The rest of the paper is organized as follows. Section \ref{sec:proposed} elaborates on the proposed BiDA. The extensive experiments and analyses on cross-temporal/scene airborne and satellite datasets are presented in Section \ref{sec:results}. Finally, conclusions are drawn in Section \ref{sec:conclusions}.


\section{Bi-directional Domain Adaptation (BiDA)}
\label{sec:proposed}
\definecolor{myblue}{RGB}{46,117,182}
\definecolor{myred}{RGB}{192,0,0}
\definecolor{mypurple}{RGB}{135,98,232}
\begin{figure*}[tp] \small
	\begin{center}
		\epsfig{width=2\figurewidth,file=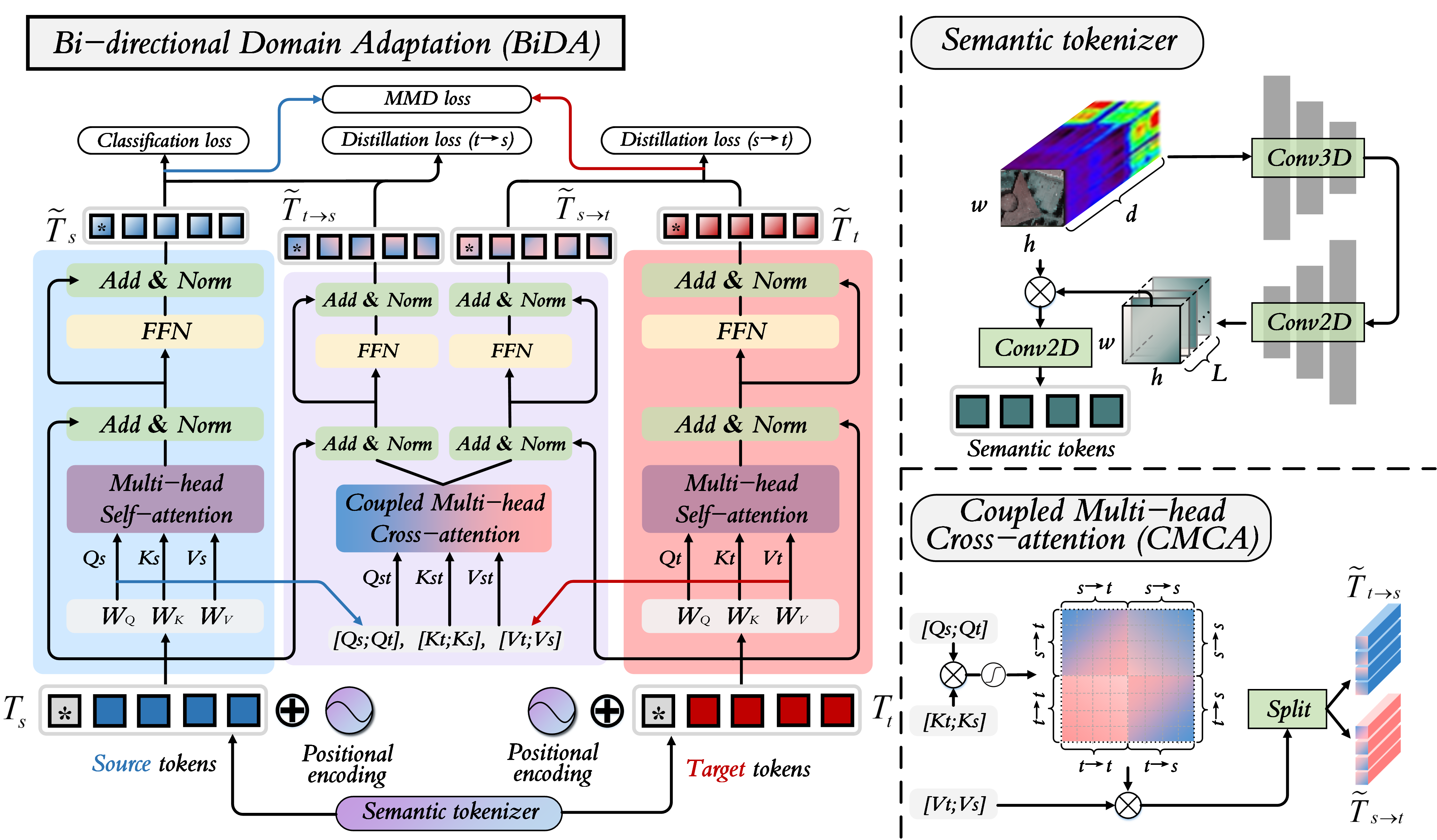}
		\caption{\label{fig:BiT}
			The main framework of BiDA is composed of a triple-branch transformer with semantic tokenizer (source branch/\textcolor{myblue}{\textbf{blue}}, target branch/\textcolor{myred}{\textbf{red}}, and coupled branch/\textcolor{mypurple}{\textbf{purple}}). The Multi-head Self-attention (MSA) is used in the source branch and the target branch, and the Coupled Multi-head Cross-attention (CMCA) is designed in the coupled branch.}
	\end{center}
\end{figure*}

Abbreviations and Notations used in this paper are summarized in Tables \ref{table:Abbreviation}-\ref{table:Notations}. Assuming that ${{\bf{X}}_s} = \left\{ {{\bf{x}}_i^s} \right\}_{i = 1}^{{n_s}}\in\mathbb{R}{^d}$ and ${{\bf{X}}_t} = \left\{ {{\bf{x}}_i^t} \right\}_{i = 1}^{{n_t}}\in\mathbb{R}{^d}$ are data from SD and TD, respectively, and ${{\bf{Y}}_s}$ and ${{\bf{Y}}_t}$ are the corresponding class labels. Note that, there will be no ${{\bf{Y}}_t}$ in the unsupervised case. Here, $d$, $n_s$ and $n_t$ denote the dimension of data, the number of source samples and the number of target samples, respectively. The main framework of proposed BiDA is shown in Fig.~\ref{fig:BiT}. The sample of $w \times h \times d$ ($w$ and $h$ are set to 13 in the experiment, i.e., 13$\times $13$\times d$) spatial patch in HSI is selected from SD and sent to semantic tokenizer to obtain semantic tokens by spatial-spectral projection. The transformer encoder of BiDA is composed of source branch, target branch, and coupled branch. In the source and target branches, Multi-head Self-attention (MSA) is employed to explore intra-domain correlations, while the CMCA in the coupled branch is proposed to perceive bi-directional inter-domains correlations. The uncertainty-based pseudo-labeling strategy is used to construct class-wise sample pairs between SD and TD for correlation mining of the three branches. Furthermore, the bi-directional distillation loss is used to impose inter-domain correlation supervision on the source branch and target branch, after obtaining representations from each branch and calculating probability distributions. Maximum Mean Discrepancy (MMD) is introduced to calculate the distribution difference of token representations in the adaptive spaces of SD and TD. Finally, ARS is designed based on the teacher-student model concept. Different noise is introduced to the input, and intra-domain consistency constraints are applied to both the SD and TD, enhancing the domain adaptation performance.

\begin{table}[]
	\caption{\label{table:Abbreviation}
		Summary of abbreviations.}
	\begin{center}
		\begin{tabular}{ll}
			\hline
			\multicolumn{1}{c}{Abbreviation}                      & Description                                                 \\ \hline
			HSI                                                   & Hyperspectral image                                     \\
			DA                                                   & Domain adaptation                                     \\
			SD                                                    & Source domain                                             \\
			TD                                                    & Target domain                                             \\
			MSA                                                    & Multi-head Self-attention                                            \\
			CMCA                                                    &  Coupled Multi-head Cross-attention                                      \\
			ARS                                                  & Adaptability Reinforcement Strategy                                      \\
			 \hline
		\end{tabular}
	\end{center}
\end{table}

\begin{table}[]
	\caption{\label{table:Notations}
		Notations of variables.}
	\begin{center}
			\begin{tabular}{ll}
			\hline
			Notations                               & Description                                                 \\ \hline
			${{\bf{X}}_s}$ and ${{\bf{X}}_t}$       & Source data and target data                \\
			${{\bf{T}}_s}$ and ${{\bf{T}}_t}$                              & Source and target semantic tokens                      \\
			${{\bf{T}}_{cls}}$       & Classification token\\
			${{\bf{\tilde T}}_s}$ and ${{\bf{\tilde T}}_t}$       & Intra-domain correlation token representations   \\
			${{\bf{\tilde T}}_{cls}}$      & Classification token representation  \\
			${{{{\bf{\tilde T}}}_{t \to s}}}$       & The coupled token representation of TD relative to SD    \\
			${{{{\bf{\tilde T}}}_{s \to t}}}$       & The coupled token representation of SD relative to TD  \\
			${\bf{\tilde T}}^{{o_1}}$ and ${\bf{\tilde T}}^{{o_2}}$       & The semantic tokens applying noise $o_1$ and noise $o_2$ \\ \hline
			&                                                             \\
			&
		\end{tabular}
	\end{center}
\end{table}

\subsection{Semantic Tokenizer}

Given that HSI is a data collection with strong spatial recognition and multi-band spectral data, it is essential to design a tokenizer capable of jointly encoding spatial-spectral information, as illustrated in Fig.~\ref{fig:BiT}. The original HSI data, with dimensions 13$\times$13$\times d$, is fed into a semantic tokenizer to generate source tokens and target tokens. Different from the method of ViT, which divides patches in the spatial dimension of the original data to create tokens, the semantic tokenizer utilizes a learned spatial-spectral projection to generate tokens.

Initially, the original HSI data is input into a spatial-spectral extractor to obtain a spatial-spectral projection, with dimensions 13$\times$13$\times L$, where $L$ represents the number of tokens ($L$ is set to 4 in all experiments). This extractor consists of a Conv3d-ReLU-MaxPool2d block followed by a Conv2d-ReLU-MaxPool2d block. Subsequently, applying a softmax function on the projection matrix calculates spatial-spectral attention maps, which are applied to each pixel in the spatial domain. Finally, a Conv2d layer with 1$\times$1 convolutional kernels is added to complete dimension mapping, resulting in compact vocabulary sets of size $L$, namely semantic tokens ${{\bf{T}}_{s/t}}\in\mathbb{R}{^{L \times d_{map}}}$. Formally, 
\begin{equation}\label{eq:1}
{{\bf{T}}_{s/t}} = \left[ {{\bf{T}}_{s/t}^{cls};M\left( {softmax {{\left( {f\left( {{{\bf{X}}_{s/t}};{\bf{W}}} \right)} \right)}^T}{{\bf{X}}_{s/t}}} \right)} \right]
\end{equation}
where $M\left( \cdot \right)$ denotes the Conv2d layer for dimension mapping, the softmax function is used to normalize the spatial-spectral projection to obtain attention maps, $f\left( \cdot \right)$ represent the spatial-spectral extractor composed of Conv3d and Conv2d with learnable kernels $\bf{W}$. ${\bf{T}}_{s/t}^{cls}$ is a learnable classification token using for the classification task and uncertainty-based pseudo-label learning.

\subsection{Triple-branch Encoder}
After obtaining two semantic tokens ${{\bf{T}}_{s}}$ and ${{\bf{T}}_{t}}$ for the SD and TD, a triple-branch encoder are used to model the context between tokens. Firstly, the source branch and target branch are composed of $N$ layers of MSA and feed-forward network (FFN) blocks, which are used to mine the context information of SD and TD and learn the intra-domain correlation token representation of the adaptive space, i.e., ${{\bf{\tilde T}}_s}$ and ${{\bf{\tilde T}}_t}$. Semantic tokens are input to three linear layers to calculate the triplet sets (query $\bf{Q}$, key $\bf{K}$, and value $\bf{V}$). Note that the learnable positional embeddings (PE) are added to tokens before feeding tokens to the transformer encoder. 
\begin{equation}\label{eq:2}
\begin{array}{l}
{{\bf{Q}}_{s/t}} = {{\bf{T}}_{s/t}}{\bf{W}}_{s/t}^q + PE,\\
{{\bf{K}}_{s/t}} = {{\bf{T}}_{s/t}}{\bf{W}}_{s/t}^k + PE,\\
{{\bf{V}}_{s/t}} = {{\bf{T}}_{s/t}}{\bf{W}}_{s/t}^v + PE
\end{array}
\end{equation}
where ${\bf{W}}_{s/t}^q$, ${\bf{W}}_{s/t}^k$ and ${\bf{W}}_{s/t}^v$ are the learnable parameters of three linear layers.

MSA consists of multiple independent attentions, with each head generating a separate representation. Its advantage lies in the ability to collectively focus on context information from different positions in the token sequence. In a single self-attention (SA), the attention map is calculated using all ${\bf{Q}}_{s/t}$, ${\bf{K}}_{s/t}$ and ${\bf{V}}_{s/t}$, and the attention score is obtained using the softmax function. Formally, 
\begin{equation}\label{eq:3}
\begin{array}{l}
SA = Attention\left( {{{\bf{Q}}_{s/t}},{{\bf{K}}_{s/t}},{{\bf{V}}_{s/t}}} \right)\\
= softmax\left( {\frac{{{{\bf{Q}}_{s/t}}{{\bf{K}}_{s/t}}^T}}{{\sqrt {{d_k}} }}} \right){{\bf{V}}_{s/t}}
\end{array}
\end{equation}
where ${\sqrt {{d_k}} }$ represents the scaling factor, which is equal to the channel dimension of ${{\bf{K}}_{s/t}}$, to avoid the variance effect caused by the dot product. MSA contains multiple SAs to calculate multi-head attention value and merge the attention scores of each head. MSA is formulated as follows,
\begin{equation}\label{eq:4}
MSA\left( {{{\bf{Q}}_{s/t}},{{\bf{K}}_{s/t}},{{\bf{V}}_{s/t}}} \right) = Concat\left( {S{A_1},S{A_2}...S{A_h}} \right){{\bf{W}}^o}
\end{equation}
where $h$ is the number of attention heads (the default value is 8), ${{\bf{W}}^o}$ is the parameter matrix. Next, the attention scores learned in the previous step are input into a skip-connected FFN to calculate the intra-domain correlation token representations ${{\bf{\tilde T}}_s}$ and ${{\bf{\tilde T}}_t}$. The FFN consists of two fully connected layers, between which there is a nonlinear activation function, Gaussian Error Linear Unit (GELU). Following the multilayer perceptron (MLP) layers is a Layer Normalization (LN), which addresses gradient explosion and mitigates the issue of gradient vanishing.

The above source and target branches, based on the SA mechanism, primarily focus on representing the intra-domain correlations within token sequences. Furthermore, it is crucial to consider how to discover inter-domain correlations between different scenes for alleviating spectral shift. 
To obtain a unified representation across different domains, the CMCA is designed within the coupled branch based on the cross-attention mechanism, as shown in Fig.~\ref{fig:BiT}. This attention method integrates information from all tokens in both the SD and TD, promoting the exploration of bi-directional inter-domain correlations. We merge the ${\bf{Q}}_{s/t}$, ${\bf{K}}_{s/t}$ and ${\bf{V}}_{s/t}$ from both SD and TD and input them into CMCA to calculate the attention scores of the coupled branch,
\begin{equation}\label{eq:5.1}
\begin{array}{l}
\left( {{{\bf{Q}}_{st}},{{\bf{K}}_{st}},{{\bf{V}}_{st}}} \right) = \left[ {{{\bf{Q}}_s};{{\bf{Q}}_t},{{\bf{K}}_t};{{\bf{K}}_s},{{\bf{V}}_t};{{\bf{V}}_s}} \right]
\end{array}
\end{equation}
\begin{equation}\label{eq:5}
\begin{array}{l}
CMCA\left( {{{\bf{Q}}_{st}},{{\bf{K}}_{st}},{{\bf{V}}_{st}}} \right) \\
= softmax\left( {\frac{{\left[ {{{\bf{Q}}_s};{{\bf{Q}}_t}} \right]{{\left[ {{{\bf{K}}_t};{{\bf{K}}_s}} \right]}^T}}}{{\sqrt {{d_k}} }}} \right)\left[ {{{\bf{V}}_t};{{\bf{V}}_s}} \right]
\end{array}
\end{equation}

\begin{equation}\label{eq:6}
\left[ {{{{\bf{\tilde T}}}_{s \to t}},{{{\bf{\tilde T}}}_{t \to s}}} \right] = Split\left( {CMCA\left( {{{\bf{Q}}_{st}},{{\bf{K}}_{st}},{{\bf{V}}_{st}}} \right)} \right)
\end{equation}
The attention scores are split and input to skip-connected FFN to obtain inter-domain token coupled representations, i.e., ${{{{\bf{\tilde T}}}_{t \to s}}}$ (the coupled token representation of TD relative to SD), and ${{{{\bf{\tilde T}}}_{s \to t}}}$ (the coupled token representation of SD relative to TD).

The SD class token representations ${\bf{\tilde T}}_s^{cls}$ are used to calculate the classification loss,
\begin{equation}\label{eq:6.1}
{{\cal L}_{cls}} =  - \frac{1}{{{n_s}}}\sum\limits_{i = 1}^{{n_s}} {\sum\limits_{c = 1}^C {y_{s,i}^c\log q_{s,i}^c} } 
\end{equation}
where ${\bf{y}}_{s,i}$ is the one-hot encoding of the label information of ${{\bf{x}}_{s,i}}$, $c$ is the index of class, $C$ is the number of classes, and ${\bf{q}}_{s,i}$ is the predicted probability. The three branches of encoder aim to capture intra-domain and inter-domain correlations of the same class within SD and TD. Thus, an uncertainty-based pseudo-labeling strategy is introduced to construct class-wise sample pairs between SD and TD. In each training iteration, the probability distribution ${{{\bf{q}}_{t,i}}}$ is computed using the TD class token representations ${\bf{\tilde T}}_t^{cls}$ from the target branch, followed by uncertainty analysis on ${{{\bf{q}}_{t,i}}}$,
\begin{equation}\label{eq:6.2}
H\left( {{{\bf{q}}_{t,i}}} \right) =  - \frac{1}{{{n_t}}}\sum\limits_{i = 1}^{{n_t}} {\sum\limits_{c = 1}^C {q_{t,i}^c\log q_{t,i}^c} }
\end{equation}
TD samples meeting the uncertainty criteria $H\left( {{{\bf{q}}_{t,i}}} \right) \le \left[ {0.5 \times log\left( C \right)} \right]$ are retained, and their predicted results are treated as pseudo-labels. These pseudo-labeled samples are then randomly paired with  SD samples from the same class as input for the next training iteration.

In order to make the coupled branch play a role of bi-directional supervision in the training of source branch and target branch, we use the inter-domain correlation to assist the independent adaptive space learning. Based on the idea of distillation learning, bi-directional distillation loss is designed. We calculate the probability distribution $\textbf{p}_{s \to t,i}$ and $\textbf{p}_{t \to s,i}$ of class token representations ${\bf{\tilde T}}_{s \to t}^{cls}$ and ${\bf{\tilde T}}_{t \to s}^{cls}$ output from the coupled branch and utilize it as soft labels. The distillation loss of SD is calculated as follows,
\begin{equation}\label{eq:10}
Distill\left( {{\bf{\tilde T}}_{t \to s}^{cls},{\bf{\tilde T}}_s^{cls}} \right) =  - \frac{1}{{{n_s}}}\sum\limits_{i = 1}^{{n_s}} {\sum\limits_{c = 1}^C {p_{t \to s,i}^c\log q_{s,i}^c} } 
\end{equation}
The distillation loss of TD is calculated as follows,
\begin{equation}\label{eq:11}
Distill\left( {{\bf{\tilde T}}_{s \to t}^{cls},{\bf{\tilde T}}_t^{cls}} \right) =  - \frac{1}{{{n_t}}}\sum\limits_{i = 1}^{{n_t}} {\sum\limits_{c = 1}^C {p_{s \to t,i}^c\log q_{t,i}^c} } 
\end{equation}
Bi-directional distillation loss is as follows,
\begin{equation}\label{eq:12}
{{\cal L}_{Bi - distill}} = Distill\left( {{\bf{\tilde T}}_{t \to s}^{cls},{\bf{\tilde T}}_s^{cls}} \right) + Distill\left( {{\bf{\tilde T}}_{s \to t}^{cls},{\bf{\tilde T}}_t^{cls}} \right)
\end{equation}

Apart from utilizing the above four types of class tokens, we introduce the MMD metric to project the token representations ${{{\bf{\tilde T}}}_s}$ and ${{{\bf{\tilde T}}}_t}$ into a Hilbert space, and calculate the marginal distribution discrepancy between adaptive spaces of SD and TD,
\begin{equation}\label{eq:7}
MMD({{{\bf{\tilde T}}}_s},{{{\bf{\tilde T}}}_t}) = \left\| {\frac{1}{{{n_s}}}\sum\limits_{i = 1}^{{n_s}} {\phi ({\bf{\tilde t}}_i^s)}  - \frac{1}{{{n_t}}}\sum\limits_{j = 1}^{{n_t}} {\phi ({\bf{\tilde t}}_j^t)} } \right\|_{\rm{H}}^2
\end{equation}
where ${\bf{\tilde t}}_i^s$ and ${\bf{\tilde t}}_j^t$ are the $i$-th and $j$-th vector of ${{\bf{\tilde T}}_s}$ and ${{\bf{\tilde T}}_t}$. Similarly, ${{\bf{\tilde T}}_{s \to t}}$ and ${{\bf{\tilde T}}_{t \to s}}$ are substituted into the Eq. \ref{eq:7} to calculate the inter-domain similarity difference in the coupled branch. Finally, the MMD loss is calculated as follows,
\begin{equation}\label{eq:9}
{{\cal L}_{MMD}} = MMD({{\bf{\tilde T}}_s},{{\bf{\tilde T}}_t}) + MMD({{\bf{\tilde T}}_{s \to t}},{{\bf{\tilde T}}_{t \to s}})
\end{equation}

\subsection{Adaptability Reinforcement Strategy}

\begin{figure}[tp] \small
	\vspace{-2em}
	\begin{center}
		\centering
		\epsfig{width=1.1\figurewidth,file=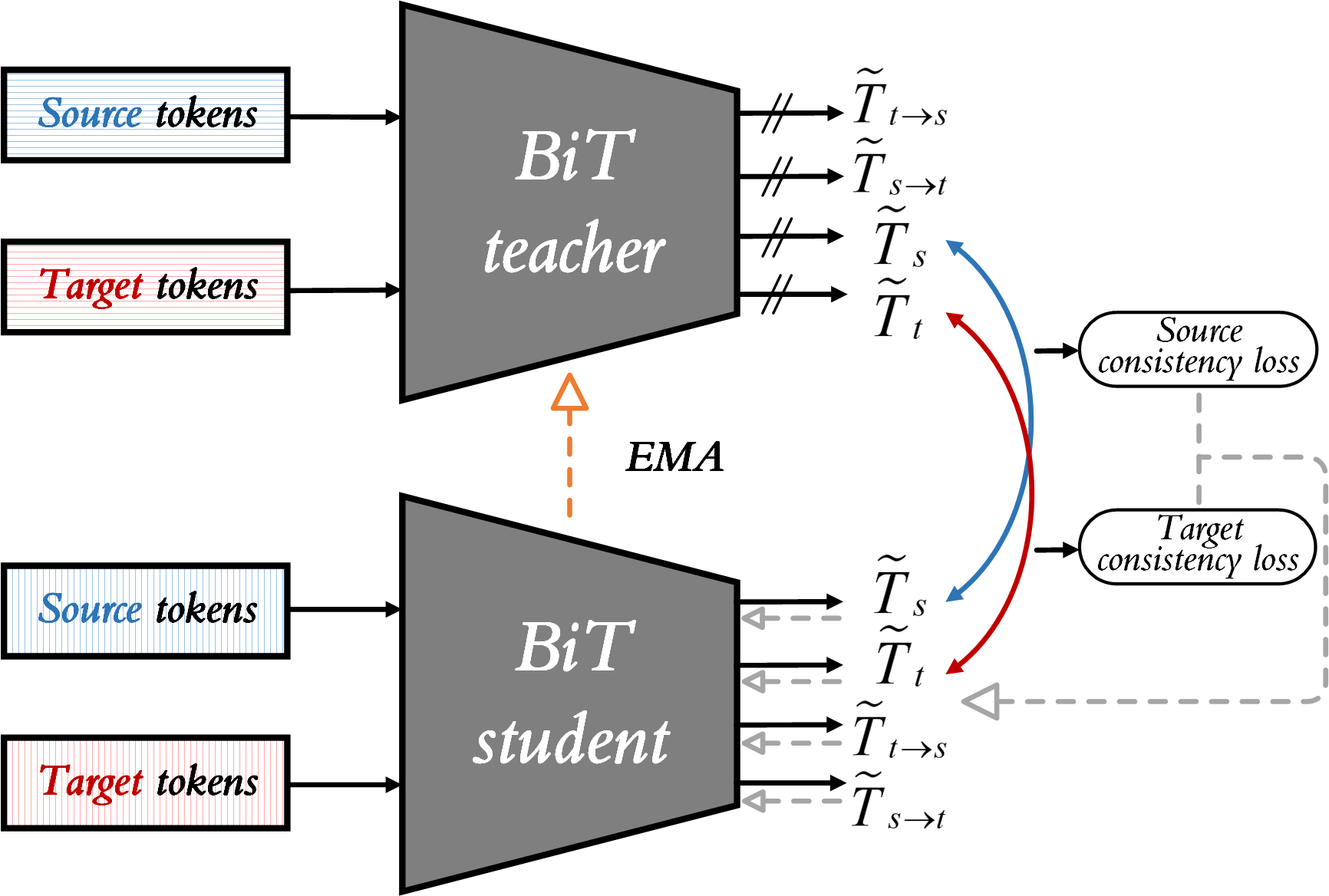}
		\caption{\label{fig:ARS}
			The flowchart of ARS. Applying different types of noise to the original HSI data respectively, we obtain two types of semantic tokens for SD and TD. These tokens are then fed into the BiDA teacher and BiDA student models. After computing the token representations, the intra-domain consistency loss is applied to SD and TD to update the student model. The teacher model is updated using EMA.}
	\end{center}
	\vspace{-2em}
\end{figure}

In the cross-domain interpretation tasks, the TD often possesses specific internal features, while current DA methods focus solely on learning domain-invariant representations, potentially failing to adequately explore and retain these specific features. Therefore, it is necessary to enhance the model's ability to extract domain-generalized features effectively while reducing spectral shift. The proposed ARS aims to make the model to better adapt to the internal structure and characteristics of the TD, as illustrated in Fig.~\ref{fig:ARS}. Different noises are added to the input (source tokens and target tokens) of the teacher model and the student model, and the intra-domain consistency constraint is applied. The teacher model is updated by exponential moving average (EMA) aggregating all the knowledge of the forward time. In the testing phase, we directly use the student model for inference. The intra-domain consistency constraint is the mean square error between the teacher model and student model. The SD is calculated as follows,
\begin{equation}\label{eq:13}
\!\!\!\!Consistency\left( {{\bf{\tilde T}}_{s,i}^{{o_1}},{\bf{\tilde T}}_{s,i}^{{o_2}}} \right) = \frac{1}{{{n_s}}}\sum\limits_{i = 1}^{{n_s}} {\left( {{f_{tec}}\left( {{\bf{\tilde T}}_{s,i}^{{o_1}}} \right) - {f_{std}}\left( {{\bf{\tilde T}}_{s,i}^{{o_2}}} \right)} \right)} 
\end{equation}
where ${\bf{\tilde T}}_{s,i}^{{o_1}}$ and ${\bf{\tilde T}}_{s,i}^{{o_2}}$ are the semantic tokens calculated after applying noise $o_1$ and noise $o_2$ to the original HSI data, the random rotation, random clipping and Gaussian noise $N\left({0, \sigma^2} \right), \sigma=0.05$, are selected for noise in the experiment. $f_{tec}$ and $f_{std}$ denote the teacher and student triple-branch encoder of BiDA.

The TD is calculated as follows,
\begin{equation}\label{eq:14}
\!\!\!\!Consistency\left( {{\bf{\tilde T}}_{t,i}^{{o_1}},{\bf{\tilde T}}_{t,i}^{{o_2}}} \right) = \frac{1}{{{n_t}}}\sum\limits_{i = 1}^{{n_t}} {\left( {{f_{tec}}\left( {{\bf{\tilde T}}_{t,i}^{{o_1}}} \right) - {f_{std}}\left( {{\bf{\tilde T}}_{t,i}^{{o_2}}} \right)} \right)}  
\end{equation}

The intra-domain consistency loss is as follows,
\begin{equation}\label{eq:15}
{{\cal L}_{con}} = Consistency\left( {{\bf{\tilde T}}_{s,i}^{{o_1}},{\bf{\tilde T}}_{s,i}^{{o_2}}} \right) + Consistency\left( {{\bf{\tilde T}}_{t,i}^{{o_1}},{\bf{\tilde T}}_{t,i}^{{o_2}}} \right)
\end{equation}

Integrating above loss functions, the total loss of BiDA is defined as follows,
\begin{equation}\label{eq:17}
{{\cal L}_{total}} = {{\cal L}_{cls}} + {\lambda _1}\left( {{{\cal L}_{MMD}} + {{\cal L}_{Bi - distill}}} \right) + {\lambda _2}{{\cal L}_{con}}
\end{equation}

Note that ${{\cal L}_{total}}$ is used to update the student model, and the teacher model is updated by EMA.
\section{Experimental results and analysis}
\label{sec:results}
In this section, the MFF cross-temporal airborne dataset, Houston cross-temporal satellite dataset and HyRANK cross-scene satellite dataset, are conducted to verify the effectiveness of the proposed BiDA (\textbf{MFF and HyRANK are fine-grained cross-domain classification tasks}). Several classic and state-of-the-art transformer-based algorithms and unsupervised deep domain adaptation algorithms are employed for comparison algorithms, including Group-Aware Hierarchical Transformer (GAHT) \cite{GAHT}, MLUDA \cite{10543066}, MSDA \cite{10620320}, Multisource Domain Generalization Two-branch network (MDGTnet) \cite{10410893}, Topological Structure and Semantic Information Transfer Network (TSTnet) \cite{TSTnet}, Confident Learning-based Domain Adaptation (CLDA) \cite{CLDA}, Supervised Contrastive Learning-Based Unsupervised Domain Adaptation (SCLUDA) \cite{SCLUDA}, Spatial–spectral Weighted Adversarial Domain Adaptation (SSWADA) \cite{SSWADA} and CACL\cite{10659915}. All comparative algorithms as well as BiDA are trained using SD data with labels and TD data, without utilizing TD label information. Classification Accuracy (CA), Overall Accuracy (OA), and Kappa Coefficient (KC) are employed to evaluate the classification performance.

\subsection{Experimental Data}

\textbf{MFF cross-temporal airborne dataset}:
Cross-temporal tree species classification is conducted at Mengjiagang Forest Farm, Heilongjiang Province, Northeast China (MFF; 130° 32' 00'' E–130° 52' 06'' E, 46° 26' 20'' N–45° 30'16'' N). The MFF covers about 155 $km^2$ and is located in the western fringe of the Wangda Mountains. The entire area of MFF contains five tree species \cite{WANG2022100032}. The AISA Eagle II diffraction grating push-broom hyperspectral imager carried by the airborne LiCHy system \cite{Pang_2016} of the Chinese Academy of Forestry Sciences was used for collecting the airborne hyperspectral data. The LiCHy was mounted on a Yun-5 aircraft flying at an altitude of 750 meters. The flights occurred four times between May 31st and June 15th, 2017. The data covers a spectral range from 400 to 1000 nm with 64 bands. The spectral resolution is 9.6nm, corresponding to a spatial resolution of 2m.

\begin{figure*}[tp]
	\begin{tabular}{cc}
		\epsfig{width=1\figurewidth,file=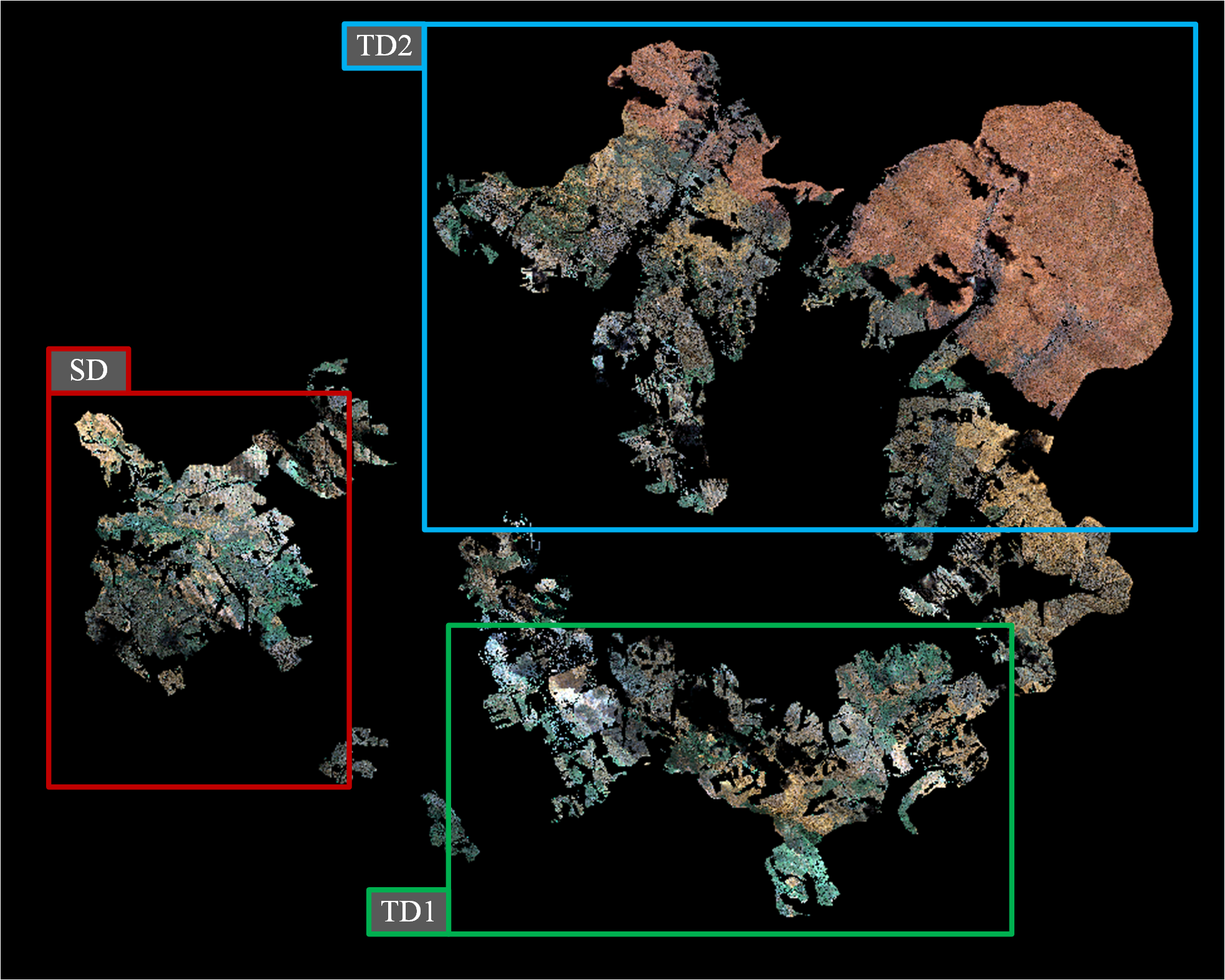}
		&
		\epsfig{width=1\figurewidth,file=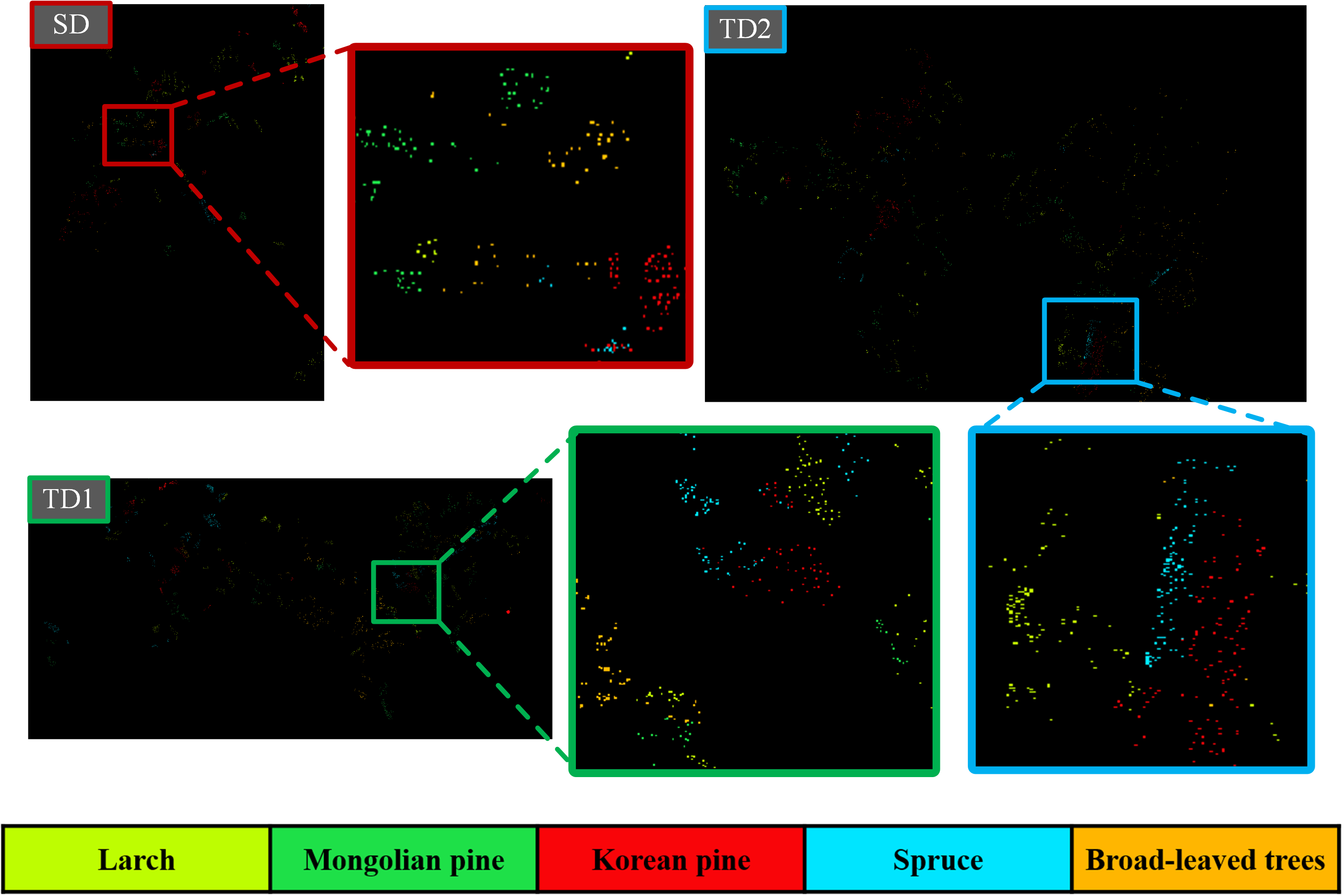}  \\
		(a) The division of MFF-SD, MFF-TD1, and MFF-TD2 & (b) Ground truth maps of MFF-SD, MFF-TD1, and MFF-TD2
	\end{tabular}
	\vspace*{0in}
	\caption{\label{fig:MFF_cross_scene}
		MFF cross-temporal airborne dataset: (a) Pseudo-color images of MFF-SD, MFF-TD1, and MFF-TD2 regions, (b) Ground truth maps.}
\end{figure*}
\begin{figure}[tp] \small
	\begin{center}
		\centering
		\epsfig{width=1\figurewidth,file=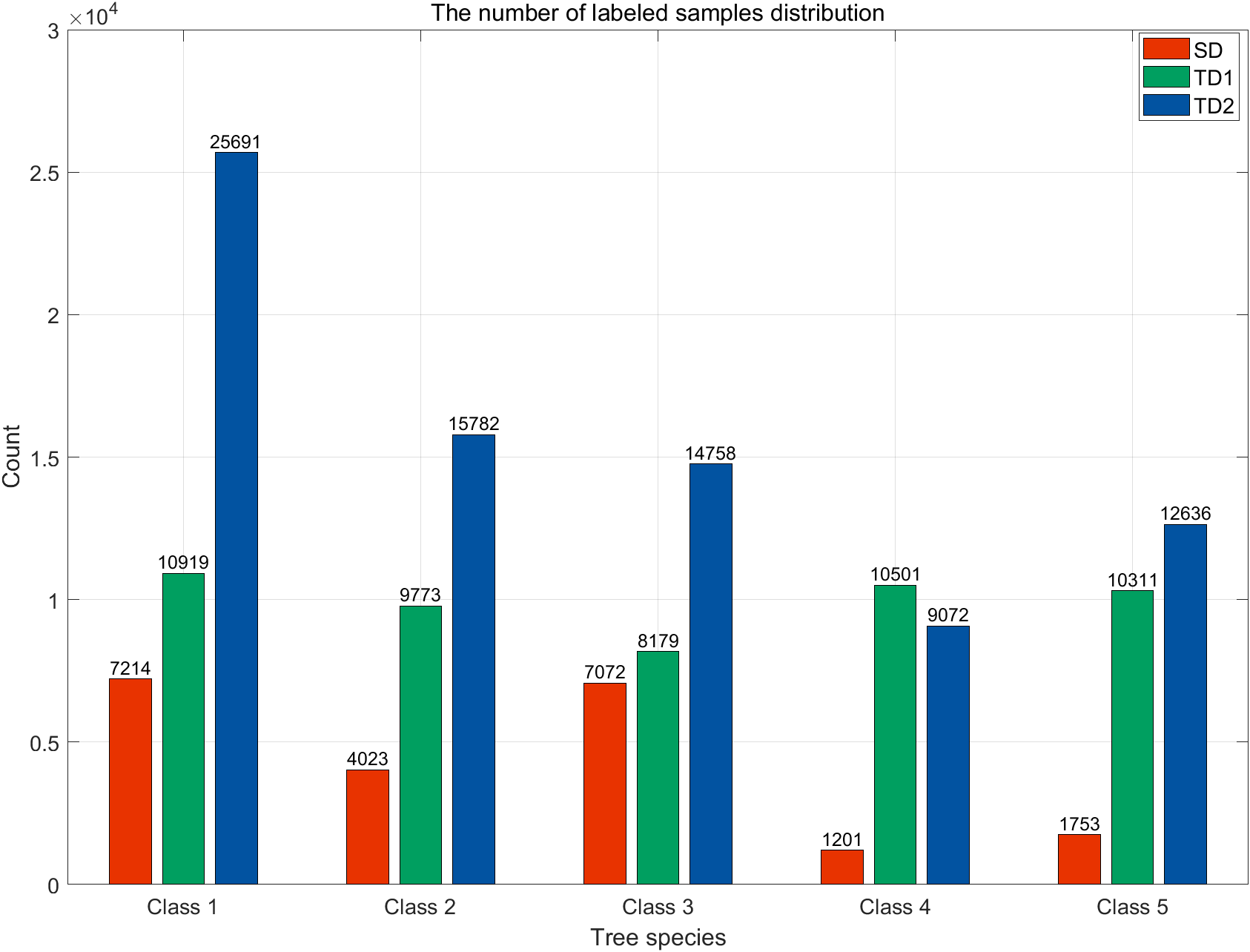}
		\caption{\label{fig:MFF_labeled_distribution}
			The number of labeled samples distribution of  MFF-SD, MFF-TD1, and MFF-TD2, where Class1 to Class5 corresponded to Larch, Mongolian pine, Korean pine, Spruce and Broad-leaved trees, respectively}
	\end{center}
	\vspace{-2em}
\end{figure}

The multi-flightline airborne HSIs collected for large-area forest farms have a large time span (\textbf{the flights occurred four times between May 31st and June 15th, 2017, a period of rapid growth relative to tree species}), resulting in obvious differences in spectral characteristics of the same tree species, even if they are geographically close. To construct the MFF cross-temporal airborne dataset, we initially chose a small region in the southwest of MFF as the MFF-SD, highlighted by the red box in Fig.~\ref{fig:MFF_cross_scene} (a). This area mostly contains five tree species, including Larch, Mongolian pine, and Korean pine. Subsequently, we selected two regions in the south and north of MMF as TD1 and TD2, marked by the green and blue boxes in Fig.~\ref{fig:MFF_cross_scene} (b). The number of labeled samples for each region is detailed in Table \ref{table:MFF_labeled}, and the distribution of labeled samples for  MFF-SD, MFF-TD1, and MFF-TD2 is visualized in Fig.~\ref{fig:MFF_labeled_distribution}. It is obvious that the selected SD region has significantly fewer samples per class compared to TD1 and TD2, especially for Spruce and Broad-leaved trees.

\begin{table}[]
	\setlength\tabcolsep{2pt}
	\caption{\label{table:MFF_labeled}
		The number of SD, TD1 and TD2 samples for the MFF cross-temporal airborne dataset.}
	\begin{center}
		\begin{tabular}{|c|c|c|c|c|c|}
			\hline \hline
			\multirow{2}{*}{No.} &\multirow{2}{*}{Class}  & MFF-SD & MFF-TD1 & { MFF-TD2}  & \multirow{2}{*}{Total} \\ {~}  & {~}  & {(Source)} &  {(Target)} &  {(Target)} & {~}\\\hline 
			1                       & { Larch}              & { 7214}  & { 10919} & { 25691} & { 43824}  \\
			2                       & { Mongolian pine}     & { 4023}  & { 9773}  & { 15782} & { 29578}  \\
			3                       & { Korean pine}        & { 7072}  & { 8179}  & { 14758} & { 30009}  \\
			4                       & { Spruce}             & { 1201}  & { 10501} & { 9072}  & { 20774}  \\
			5                       & { Broad-leaved trees} & { 1753}  & { 10311} & { 12636} & { 24700}  \\ \hline
			\multicolumn{2}{|c|}{Total}                                           & { 21263} & { 49683} & { 77939} & { 148885} \\ \hline \hline
		\end{tabular}
	\end{center}
\end{table}


\begin{figure*}[tp]
	\begin{center}
		\begin{tabular}{cc}
			\epsfig{width=0.9\figurewidth,file=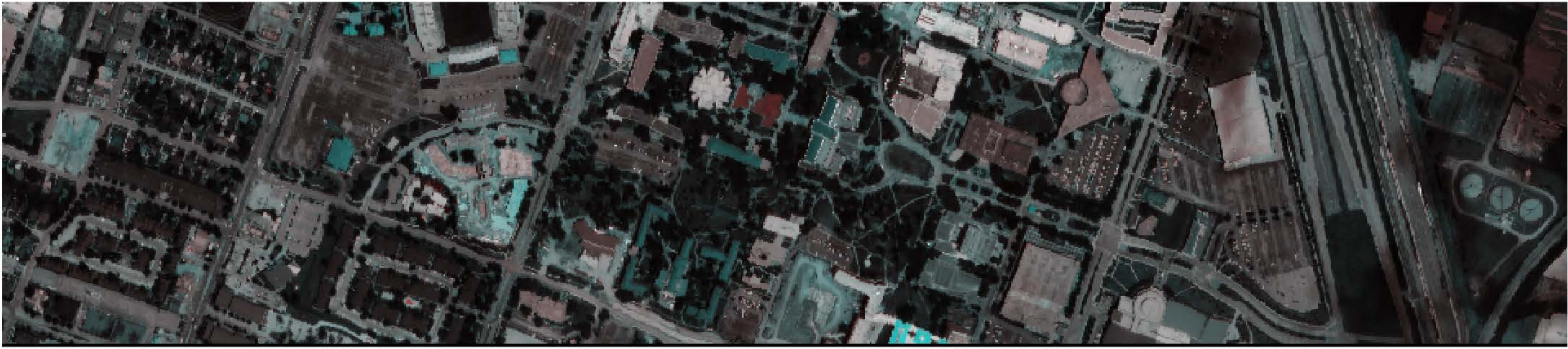} &
			\epsfig{width=0.9\figurewidth,file=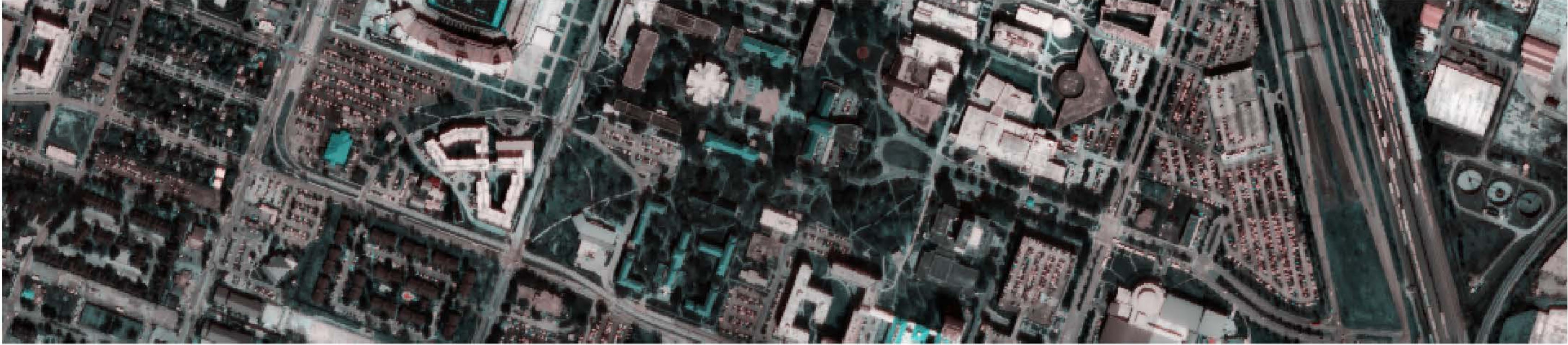} \\
			(a) & (b) \\
			\epsfig{width=0.9\figurewidth,file=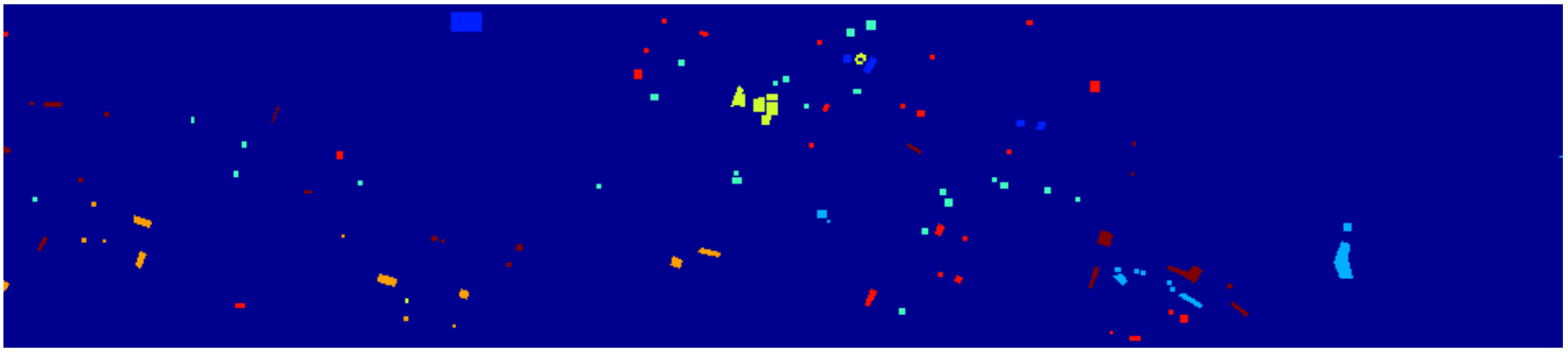} &
			\epsfig{width=0.9\figurewidth,file=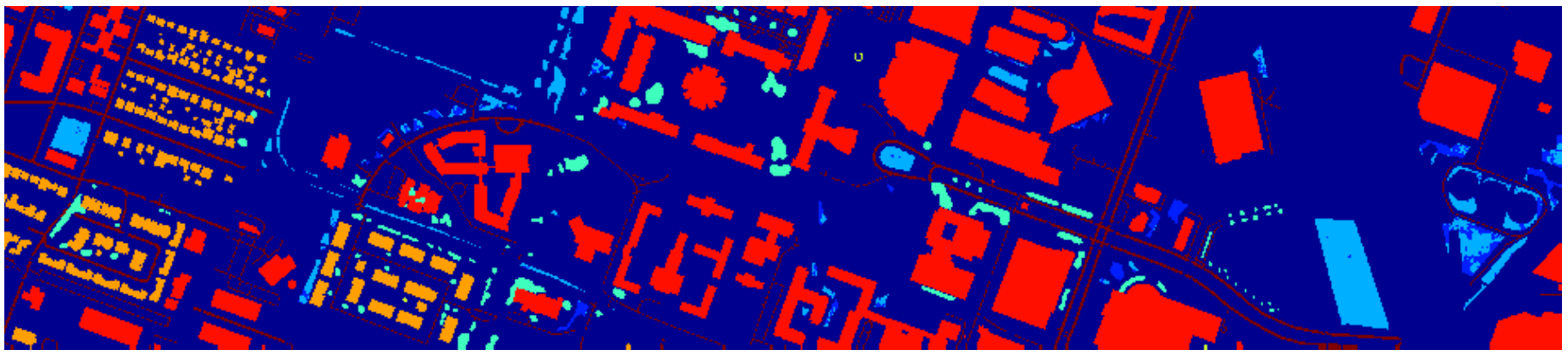} \\
			(c) & (d)   \\
		\end{tabular}
		\begin{tabular}{cc}
			\epsfig{width=1\figurewidth,file=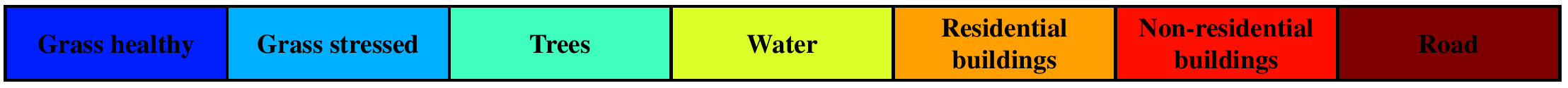}
		\end{tabular}
	\end{center}
	\caption{\label{fig:Houston_fg}
		Houston cross-temporal satellite dataset: (a) Pseudo-color image of Houston 2013, (b) Pseudo-color image of Houston 2018, (c) Ground truth map of Houston 2013, (d) Ground truth map of Houston 2018.}
\end{figure*}

\textbf{Houston cross-temporal satellite dataset}: The dataset includes Houston 2013 \cite{2014Hyperspectral} and Houston 2018 \cite{20182018} scenes, which were obtained by different sensors on the University of Houston campus and its vicinity in different years. The Houston 2013 dataset is composed of 349$\times$1905 pixels, including 144 spectral bands, the wavelength range is 380-1050nm, and the image spatial resolution is 2.5m. The Houston 2018 dataset has the same wavelength range but contains 48 spectral bands, and the image has a spatial resolution of 1m. There are seven consistent classes in their scene. We extract 48 spectral bands (wavelength range 0.38$\sim$1.05um) from Houston 2013 scene corresponding to Houston 2018 scene, and select the overlapping area of 209$\times$955. The classes and the number of samples are listed in Table \ref{table:Houston_samples}. Additionally, their false-color and ground truth maps are shown in Fig. \ref{fig:Houston_fg}.

\textbf{HyRANK cross-scene satellite dataset}:
In order to verify the effectiveness of the method on satellite platforms, we chose the HyRANK dataset, which was developed in the framework of the International Society for Photogrammetry and Remote Sensing (ISPRS) Scientific Initiatives \cite{HyRANK}. The satellite HSI collected by the Hyperion sensor (EO-1, USGS) has 176 spectral bands. The two labeled scenes are Dioni and Loukia, which are composed of 250$\times$1376 pixels and 249$\times$945 pixels, respectively. There are 12 consistent classes, which are listed in Table \ref{table:HyRANK_samples}, including multiple tree species, such as Fruit Trees, Olive Groves, Coniferous Forest and Sderophyllous Vegetation, and other classes, such as non Irrigated Arable Land and water. 

\begin{table}[tp]
	\caption{\label{table:Houston_samples}
		Number of source and target samples for the Houston dataset.}
	{
		\begin{center}
			\begin{tabular}{|c|c|c|c|}
				\hline \hline
				\multicolumn{2}{|c}{Class} &\multicolumn{2}{|c|}{Number of Samples} \\
				\hline
				\multirow{2}{*}{No.} &\multirow{2}{*}{Name}  & Houston 2013 & Houston 2018\\
				{~}  & {~}  & {(Source)} &  {(Target)} \\
				\hline
				1     & Grass healthy & 345   & 1353 \\
				2     & Grass stressed & 365   & 4888 \\
				3     & Trees & 365   & 2766 \\
				4     & Water & 285   & 22 \\
				5     & Residential buildings & 319   & 5347 \\
				6     & Non-residential buildings & 408   & 32459 \\
				7     & Road  & 443   & 6365 \\
				\hline
				\multicolumn{2}{|c|}{Total} & 2530& 53200\\
				\hline \hline
			\end{tabular}
	\end{center}}
\end{table}

\begin{table}[tp]
	\caption{\label{table:HyRANK_samples}
		Number of source and target samples for the HyRANK dataset.}
	{
		\begin{center}
			\begin{tabular}{|c|c|c|c|}
				\hline \hline
				\multirow{2}{*}{No.} &\multirow{2}{*}{Class}  & Dioni & Loukia\\
				{~}  & {~}  & {(Source)} &  {(Target)} \\
				\hline
				1     & Dense Urban Fabric & 1262  & 206 \\
				2     & Mineral Extraction Sites & 204   & 54 \\
				3     & Non Irrigated Arable Land & 614   & 426 \\
				4     & Fruit Trees& 150   & 79 \\
				5     & Olive Groves & 1768  & 1107 \\
				6     & Coniferous Forest& 361   & 422 \\
				7     & Dense Sderophyllous Vegetation & 5035  & 2996 \\
				8     & Sparce Sderophyllous Vegetation & 6374  & 2361 \\
				9     & Sparcely Vegetated Areas & 1754  & 399 \\
				10    & Rocks and Sand & 492   & 453 \\
				11    & Water & 1612  & 1393 \\
				12    & Coastal Water & 398   & 421 \\
				\hline
				\multicolumn{2}{|c|}{Total} & 20024 & 10317\\
				\hline \hline
			\end{tabular}
	\end{center}}
\end{table}

\subsection{Parameter tuning}

\begin{figure*}[tp]
	\begin{center}
		\centering
		\begin{tabular}{cc}
			\hspace{-1cm}
			\epsfig{width=0.52\figurewidth,file=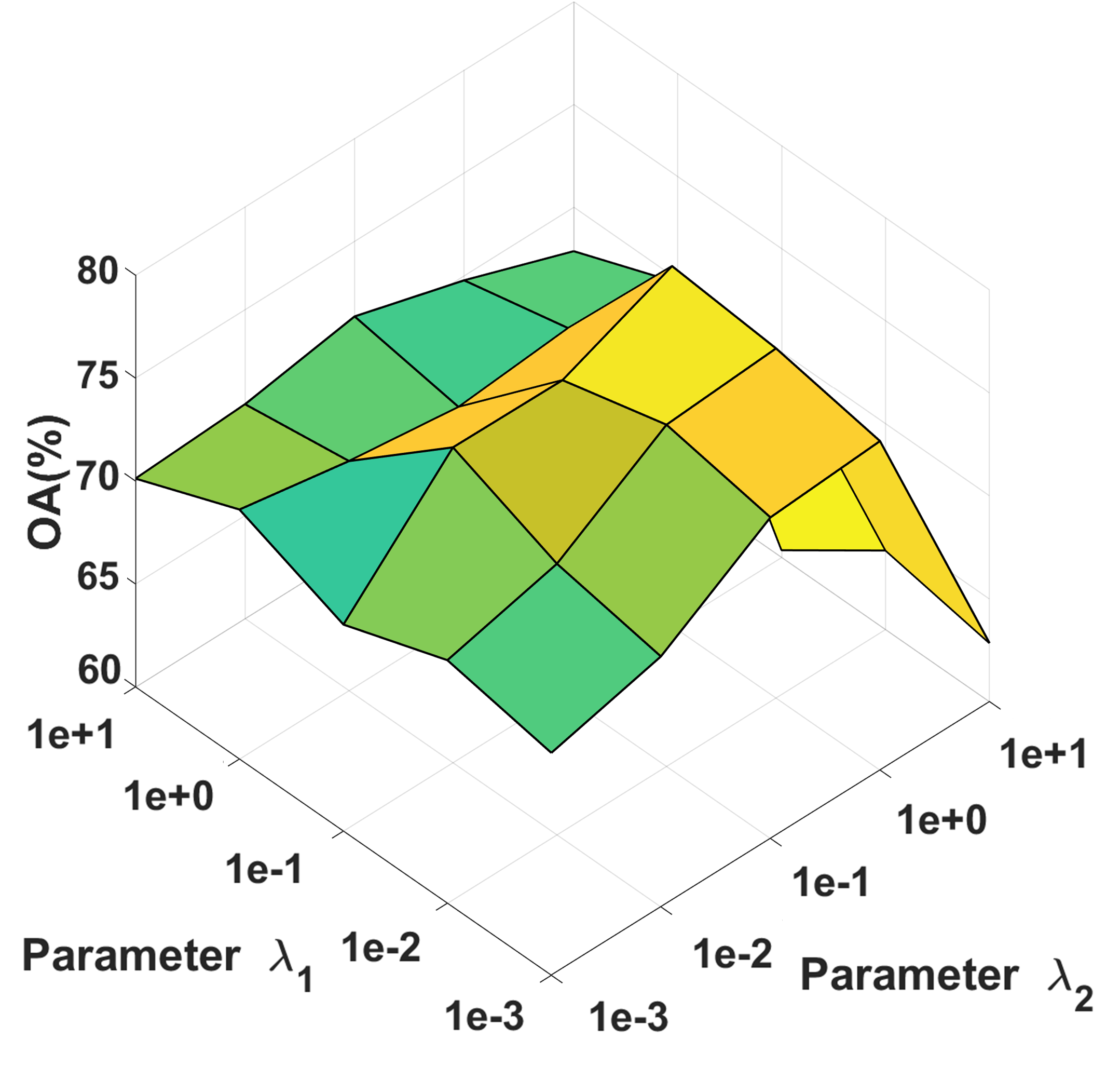}
			&
			\epsfig{width=0.52\figurewidth,file=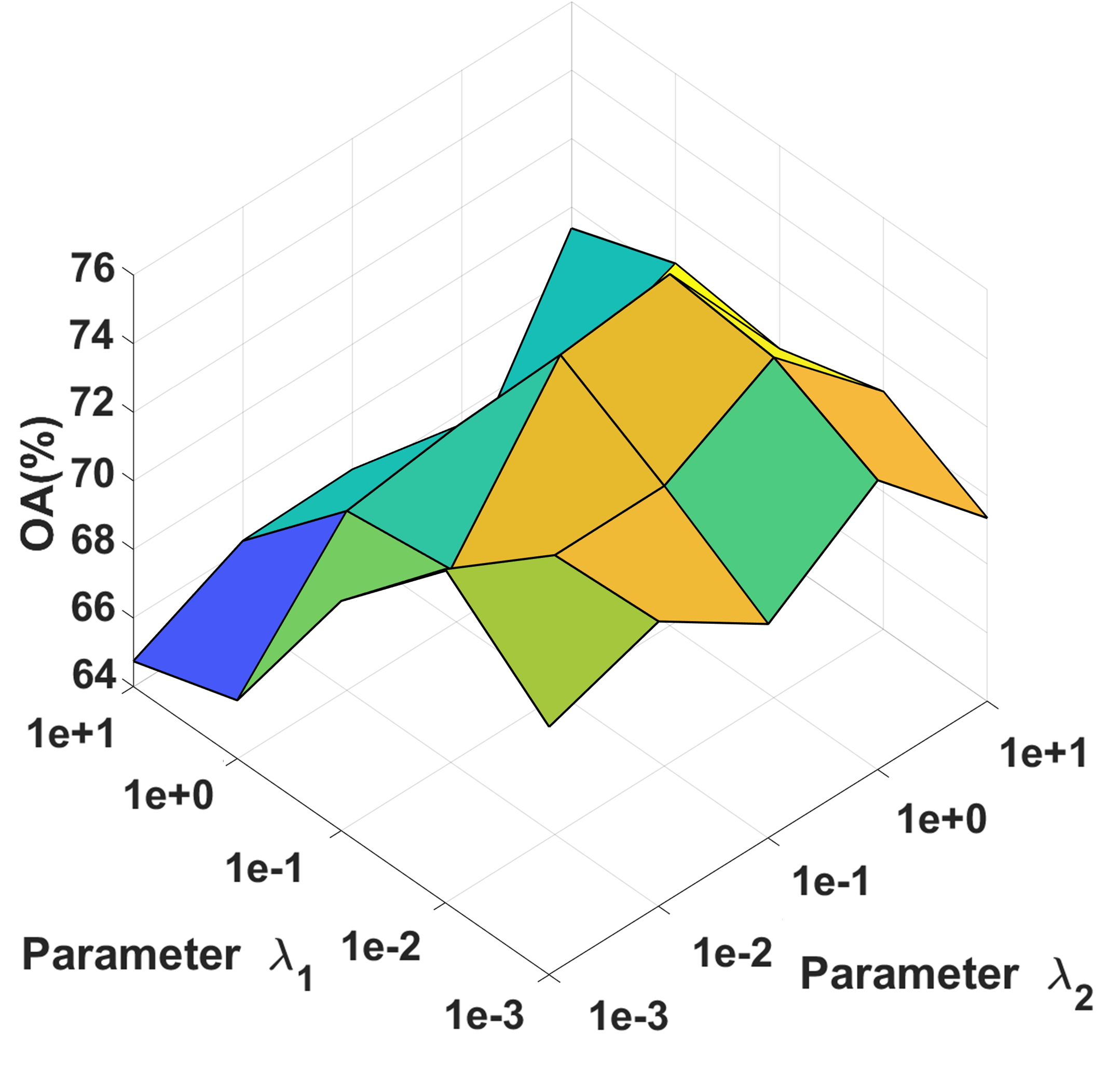} \\
			(a) MFF-TD1  & (b) MFF-TD2 \\  [0.5em]
		\end{tabular}
		\begin{tabular}{cc}
			\epsfig{width=0.52\figurewidth,file=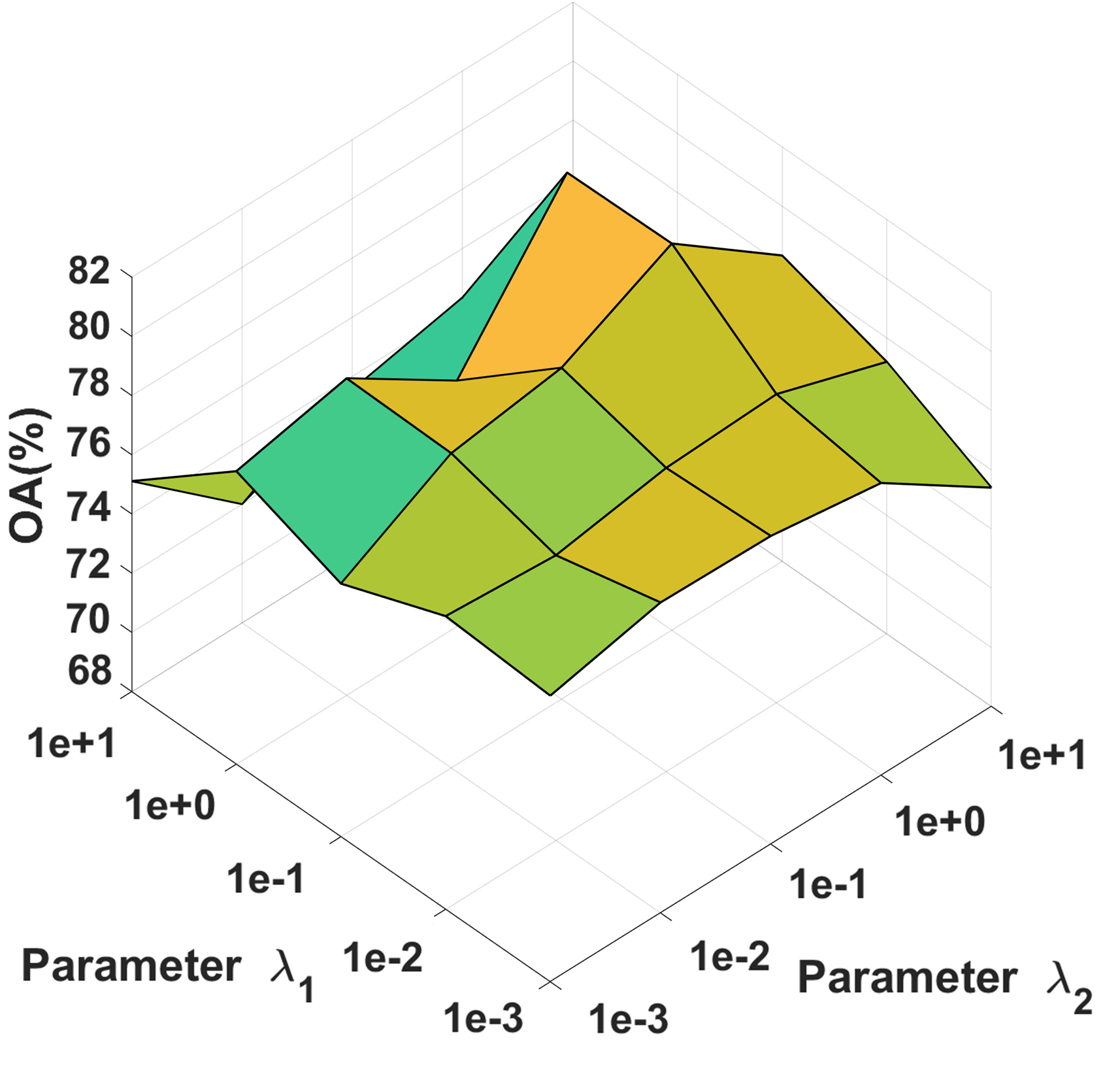} &
			\epsfig{width=0.52\figurewidth,file=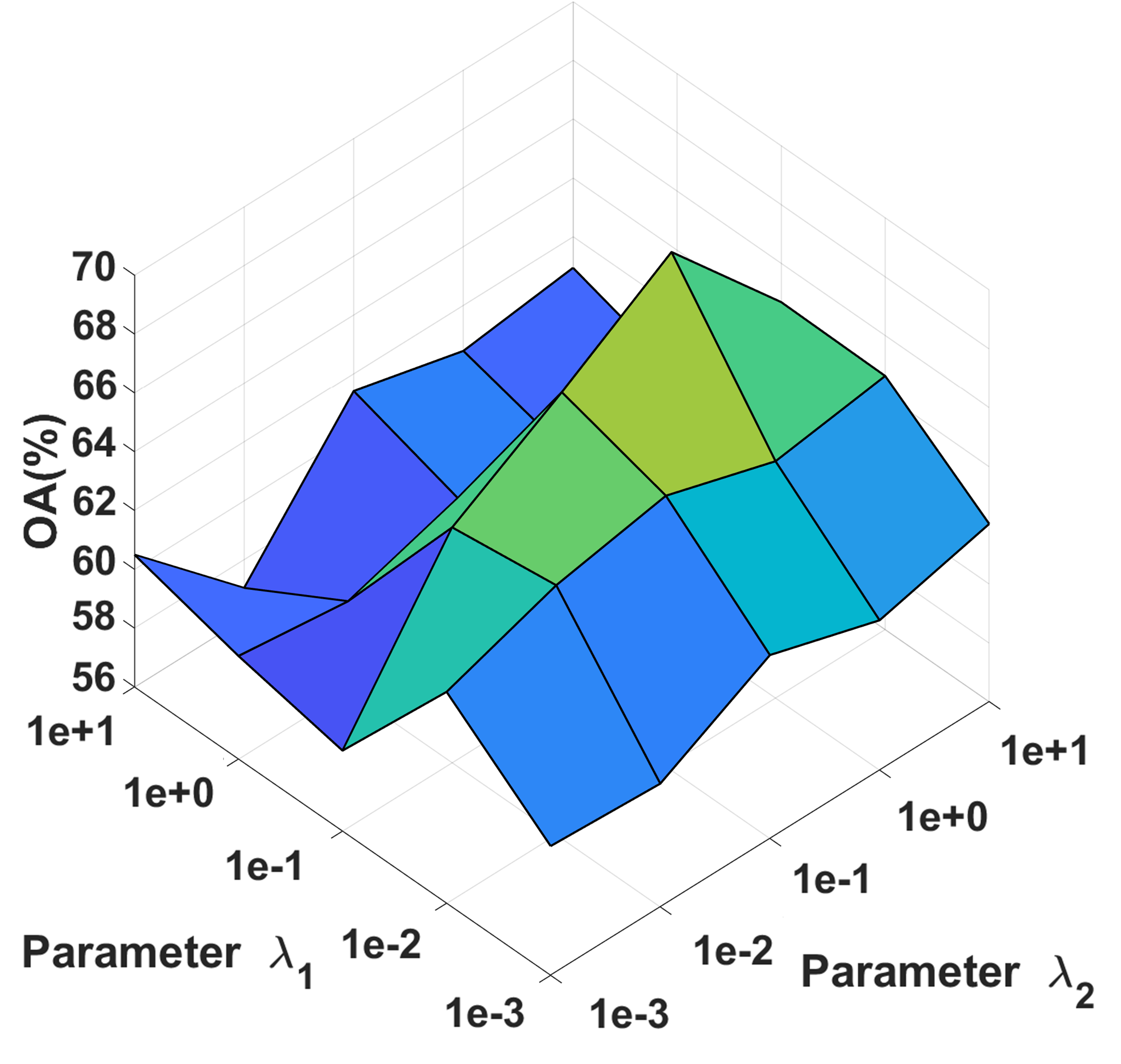} \\
			(c) Houston 2018  & (d) Loukia
		\end{tabular}
		\vspace*{0in}
		\caption{\label{fig:Parameter Tuning}
		Parameter tuning of $\lambda_1$ and $\lambda_2$ for the proposed BiDA using all the four experimental data. The optimal values of parameters $\lambda_1$ and $\lambda_2$ for the MFF-TD1, MFF-TD2, Houston 2018 and Loukia are determined to be $1e-1$ and $1e+0$, respectively.}
	\end{center}
\end{figure*}

According to the proposed BiDA, adjustable parameters are the regularization parameters, i.e., $\lambda_1$ and $\lambda_2$, and are selected from \{$1e-3$, $1e-2$, $1e-1$, $1e+0$, $1e+1$\}. The $\lambda_1$ and $\lambda_2$ are important hyperparameters of BiDA, which control the contribution of domain alignment and ARS to domain adaptation. Fig. \ref{fig:Parameter Tuning} shows the changing trend of the classification accuracy of BiDA in all experimental datasets with different combinations of $\lambda_1$ and $\lambda_2$ (indicated by OA). The optimal values of parameters $\lambda_1$ and $\lambda_2$ for the MFF-TD1, MFF-TD2, Houston 2018 and Loukia are determined to be $1e-1$ and $1e+0$, respectively.

To analyze the impact of the number of tokens \( L \) on BiDA, as shown in Table \ref{tab:L}, we present the classification accuracy across all datasets for different values of \( L \). In BiDA, a small number of tokens is sufficient to effectively represent HSI. Increasing the number of tokens tends to degrade classification performance. In traditional ViT used for image classification, the number of tokens is typically set to 192 for images of size \( 224 \times 224 \times 3 \). In BiDA, the input image size is \( 13 \times 13 \times d \). Although \( d \) is much larger than 3, our designed semantic tokenizer comprehensively considers spatial and spectral multidimensional information to serialize high-dimensional images.

\begin{table}[]
	\caption{\label{tab:L}
		The number of tokens $L$ for the proposed BiDA using the four experimental data.}
	\centering
	\begin{tabular}{|c|c|c|c|c|}
		\hline\hline
		{\begin{tabular}[c]{@{}c@{}}Target scene\\ OA (\%)\end{tabular}}& 4 & 10          & 20  & 30\\ \hline
		MFF-TD1 & \textbf{77.40} & 76.96 & 75.83 & 76.01  \\ \hline
		MFF-TD2 & \textbf{75.08} & 74.62 & 74.85 & 72.62 \\ \hline
		Houston 2018       & \textbf{81.11} & 80.05 & 79.43 & 79.83 \\  \hline
		Loukia       & \textbf{68.89} & 68.11 & 67.68 & 67.27 \\ \hline\hline
	\end{tabular}
	\vspace{-0.1em}
\end{table}

\subsection{Ablation study}

We conducted ablation analysis on important components of BiDA. Reviewing the overall loss Eq. \ref{eq:14}, the loss function of BiDA mainly consists of classification loss, MMD loss, Bi-directional distillation loss, and ARS intra-domain consistency loss. The ablation results of each loss are listed in Table \ref{table:Ablation}. In the first row, only classification loss is retained to evaluate the transfer performance of the backbone of BiDA on four TDs. Referring to Tables \ref{tab:accuracy_SD2-TD1}-\ref{tab:accuracy_SD2-TD2}, compared to other transformer-based methods, the backbone of BiDA can improve classification accuracy on TDs by 0.7\% to 3\%. The second and third rows assess the effectiveness of MMD loss and Bi-directional distillation loss. MMD significantly alleviates the negative effects of potential spectral shifts in the adaptive space of SD and TD on the domain-invariant representation learning, improving classification accuracy by 5.86\%, 4.22\%, 2.84\% and 2.96\% on four TDs compared to the backbone. Building upon MMD, Bi-directional distillation loss contributes to adaptive space learning, and achieves around 2\% accuracy improvement. Furthermore, we analyze the effectiveness of ARS, which brings about a slight improvement of approximately 0.5\% as shown in the third row of Table \ref{table:Ablation}.

We conducted ablation experiments on each branch of the triple-branch encoder and the CMCA to analyze their contributions to the overall framework of BiDA. The ablation results are shown in Table \ref{tab:Ablation_branch}. SD-b denotes retaining only the source branch (including classification loss), (SD+TD)-b represents retaining both the source branch and target branch (including classification loss and MMD loss). MCA indicates replacing the CMCA in the coupled branch with MCA (including classification loss, MMD loss, and unidirectional distillation loss). CMCA indicates coupled multi-head cross attention. \textbf{Note that all ablation settings mentioned above have removed ARS}. When only the source branch is used, the classification accuracy remains at a low level, failing to effectively recognize the TD. In (SD+TD)-b, adding the target branch and employing MMD loss improves the classification accuracy by 2\% to 4\% across all datasets. Clearly, extracting prior information from the unlabeled data of TD and using it for feature alignment is effective. In MCA, while maintaining the source branch and target branch, the coupled branch with traditional multi-head cross-attention is added. Compared to (SD+TD)-b, the classification performance improves by 4\% to 9\%. This is attributed to the coupled branch obtaining domain-invariant features by mining inter-domain correlations. The bi-directional interaction in CMCA promotes the acquisition of optimal domain-invariant features, leading to an improvement of about 2\% in classification accuracy across all datasets when MCA is replaced with CMCA.

We conducted ablation experiments on different noises in ARS, including random cropping and scaling (RandomRC), Gaussian noise (Gauss), and radiation noise (Radiation). The radiation noise is simulated by adjusting the radiation values of the original data and adding Gaussian noise, which simulates the radiation noise that may be encountered during the acquisition of HSI. The ablation results are shown in  Table \ref{tab:Ablation_noise}. Without using any noise, ARS shows almost no improvement. When adding one or more types of noise, such as RandomRC, Gauss, and Radiation, to the inputs of SD and TD, the classification accuracy of BiDA improves to varying degrees across all datasets. The improvement ranged from 2\% to 4\% when both RandomRC and Gauss are added simultaneously. Since Radiation adjusts the radiation values of the original data, it introduces significant noise interference. Therefore, compared to using two types of noise (RandomRC and Gauss), using all three types of noise only demonstrated optimal performance in the urban scene Houston 2018, which contains coarse-grained land cover classes.

We conducted ablation analysis using different tokenizers. Table \ref{table:tokenizer} displays the OA corresponding to using Patch tokenizer and semantic tokenizer in BiDA, where patch tokenizer is a commonly used token construction method in ViT. The proposed semantic tokenizer demonstrates significant advantages, with classification accuracy higher than patch tokenizer by 8.34\%, 7.75\%, 7.3\% and 8.18\% on four TDs, respectively. This indicates the importance of designing tokenizers that align with the data characteristics of HSI, which has strong spatial recognition and multi-band spectral characteristics, in transformer-based methods.

\begin{table}[tp]
	\setlength\tabcolsep{2pt}
	\caption{\label{table:Ablation}
		Ablation comparison of each loss of BiDA.}
	{
		\begin{center}
			\begin{tabular}{|c|c|c|c|c|c|c|}
				\hline \hline
				\multicolumn{3}{|c|}{Ablation Setting} & \multicolumn{4}{c|}{OA (\%)}                      \\ \hline 
				MMD       & Bi-distill   & ARS       & MFF-TD1        & MFF-TD2     & Houston 2018     & Loukia   \\ \hline
				×          & ×            & ×         & 68.82          & 69.11      & 75.07     &  61.02        \\
				$\surd$    & ×            & ×         & 74.68          & 73.33      & 77.91     & 63.98          \\
				$\surd$    & $\surd$      & ×         & 76.32          & 74.85      & 80.27    & 67.23          \\
				$\surd$    & $\surd$      & $\surd$   & \textbf{77.40} & \textbf{75.08} & \textbf{81.11}& \textbf{68.89} \\ \hline \hline 
			\end{tabular}
	\end{center}}
\end{table}

\begin{table}[tp]
	\setlength\tabcolsep{1pt}
	\caption{\label{tab:Ablation_branch}
		Ablation comparison of each branch and CMCA of BiDA.}
	{
		\begin{center}
			\begin{tabular}{|c|c|c|c|c|c|c|c|}
				\hline \hline
				\multicolumn{4}{|c|}{Ablation Setting} & \multicolumn{4}{c|}{OA (\%)}                      \\ \hline 
				SD-b       & (SD+TD)-b   & MCA  & CMCA     & MFF-TD1        & MFF-TD2     & Houston 2018     & Loukia   \\ \hline
				$\surd$          & ×            & ×   & ×    & 61.33          & 60.93      & 70.62     &  58.08        \\
				×    & $\surd$            & ×   & ×      & 65.70          & 63.44      & 72.85     & 60.57          \\
				×   & $\surd$      & $\surd$    & ×     & 74.92          & 73.01      & 78.30    & 64.60          \\
				×    & $\surd$      & ×  & $\surd$  & \textbf{76.32} & \textbf{74.85} & \textbf{80.27}& \textbf{67.23} \\ \hline \hline 
			\end{tabular}
	\end{center}}
\end{table}

\begin{table}[tp]
	\setlength\tabcolsep{2pt}
	\caption{\label{tab:Ablation_noise}
		Ablation comparison of different noises in ARS.}
	\vspace{-2em}
	{
		\begin{center}
			\begin{tabular}{|c|c|c|c|c|c|c|}
				\hline \hline
				\multicolumn{3}{|c|}{Ablation Setting} & \multicolumn{4}{c|}{OA (\%)}                      \\ \hline 
				RandomRC        & Gauss    & Radiation       & MFF-TD1        & MFF-TD2     & Houston 2018     & Loukia   \\ \hline
				×          & ×            & ×         & 75.77          & 72.60      & 78.86     &  65.22        \\
				$\surd$    & ×            & ×         & 76.58          & 73.92      & 79.73     & 67.68          \\
				$\surd$    & $\surd$      & ×         & \textbf{77.40}          & \textbf{75.08}      & 81.11    & \textbf{68.89}          \\
				$\surd$    & $\surd$      & $\surd$   & 76.94 & 74.76 & \textbf{81.27}& 67.11 \\ \hline \hline 
			\end{tabular}
	\end{center}}
\end{table}

\begin{table}[]
	\setlength\tabcolsep{2pt}
	\caption{\label{table:tokenizer}
		Ablation comparison of tokenizer for the proposed BiDA using the four experimental data.}
	\centering
	\begin{tabular}{|c|c|c|c|c|}
		\hline \hline
		{\begin{tabular}[c]{@{}c@{}}tokenizer\\ OA (\%)\end{tabular}}& MFF-TD1 & MFF-TD2    & Houston 2018   & Loukia \\ \hline
		Patch tokenizer & 69.06 & 67.33 &  73.81  &  60.71 \\ \hline
		Semantic tokenizer & \textbf{77.40} & \textbf{75.08}& \textbf{81.11} & \textbf{68.89} \\ \hline \hline
	\end{tabular}
	\vspace{-0.1em}
\end{table}

%

%

\begin{table*}[tp]
	\caption{\label{tab:accuracy_SD2-TD1}
		Class-specific and overall classification accuracy (\%) of different methods for the target scene MFF-TD1 data.}
	\begin{center}
	\begin{tabular}{|c|cccccccccc|}
		\hline \hline
		\multirow{2}{*}{Class} &
		\multicolumn{10}{c|}{Classification algorithms} \\ \cline{2-11} 
		&
		\multicolumn{1}{c|}{GAHT} &
		\multicolumn{1}{c|}{MLUDA} &
		\multicolumn{1}{c|}{MSDA} &
		\multicolumn{1}{c|}{TSTnet} &
		\multicolumn{1}{c|}{MDGTnet} &
		\multicolumn{1}{c|}{CLDA} &
		\multicolumn{1}{c|}{SCLUDA} &
		\multicolumn{1}{c|}{SSWADA} &
		\multicolumn{1}{c|}{CACL} &
		BiDA \\ \hline
		1 &
		\multicolumn{1}{c|}{84.88} &
		\multicolumn{1}{c|}{88.25} &
		\multicolumn{1}{c|}{66.87} &
		\multicolumn{1}{c|}{73.37} &
		\multicolumn{1}{c|}{68.63} &
		\multicolumn{1}{c|}{76.58} &
		\multicolumn{1}{c|}{76.31} &
		\multicolumn{1}{c|}{70.63} &
		\multicolumn{1}{c|}{79.22} &
		97.69 \\
		2 &
		\multicolumn{1}{c|}{82.67} &
		\multicolumn{1}{c|}{98.98} &
		\multicolumn{1}{c|}{90.00} &
		\multicolumn{1}{c|}{85.47} &
		\multicolumn{1}{c|}{78.97} &
		\multicolumn{1}{c|}{92.61} &
		\multicolumn{1}{c|}{93.62} &
		\multicolumn{1}{c|}{91.54} &
		\multicolumn{1}{c|}{83.17} &
		95.26 \\
		3 &
		\multicolumn{1}{c|}{68.38} &
		\multicolumn{1}{c|}{65.98} &
		\multicolumn{1}{c|}{36.58} &
		\multicolumn{1}{c|}{52.46} &
		\multicolumn{1}{c|}{49.24} &
		\multicolumn{1}{c|}{51.91} &
		\multicolumn{1}{c|}{54.46} &
		\multicolumn{1}{c|}{61.36} &
		\multicolumn{1}{c|}{70.08} &
		75.11 \\
		4 &
		\multicolumn{1}{c|}{43.95} &
		\multicolumn{1}{c|}{41.16} &
		\multicolumn{1}{c|}{55.79} &
		\multicolumn{1}{c|}{49.86} &
		\multicolumn{1}{c|}{41.65} &
		\multicolumn{1}{c|}{42.65} &
		\multicolumn{1}{c|}{13.85} &
		\multicolumn{1}{c|}{20.92} &
		\multicolumn{1}{c|}{39.75} &
		44.35 \\
		5 &
		\multicolumn{1}{c|}{57.46} &
		\multicolumn{1}{c|}{69.25} &
		\multicolumn{1}{c|}{85.77} &
		\multicolumn{1}{c|}{77.59} &
		\multicolumn{1}{c|}{85.35} &
		\multicolumn{1}{c|}{75.41} &
		\multicolumn{1}{c|}{75.40} &
		\multicolumn{1}{c|}{40.16} &
		\multicolumn{1}{c|}{55.20} &
		74.46 \\ \hline
		OA (\%) &
		\multicolumn{1}{c|}{67.38} &
		\multicolumn{1}{c|}{72.80} &
		\multicolumn{1}{c|}{68.01} &
		\multicolumn{1}{c|}{68.21} &
		\multicolumn{1}{c|}{65.24} &
		\multicolumn{1}{c|}{68.26} &
		\multicolumn{1}{c|}{62.72} &
		\multicolumn{1}{c|}{56.39} &
		\multicolumn{1}{c|}{66.86} &
		\textbf{77.40} \\ \hline
		KC ($\kappa$) &
		\multicolumn{1}{c|}{59.22} &
		\multicolumn{1}{c|}{65.98} &
		\multicolumn{1}{c|}{59.93} &
		\multicolumn{1}{c|}{60.24} &
		\multicolumn{1}{c|}{56.51} &
		\multicolumn{1}{c|}{60.31} &
		\multicolumn{1}{c|}{53.55} &
		\multicolumn{1}{c|}{45.78} &
		\multicolumn{1}{c|}{58.62} &
		\textbf{71.76} \\ \hline \hline
	\end{tabular}
\end{center}
\end{table*}

\begin{table*}[tp]
	\caption{\label{tab:accuracy_SD2-TD2}
		Class-specific and overall classification accuracy (\%) of different methods for the target scene MFF-TD2 data.}
	\begin{center}
	\begin{tabular}{|c|cccccccccc|}
		\hline \hline
		\multirow{2}{*}{Class} &
		\multicolumn{10}{c|}{Classification algorithms} \\ \cline{2-11} 
		&
		\multicolumn{1}{c|}{GAHT} &
		\multicolumn{1}{c|}{MLUDA} &
		\multicolumn{1}{c|}{MSDA} &
		\multicolumn{1}{c|}{TSTnet} &
		\multicolumn{1}{c|}{MDGTnet} &
		\multicolumn{1}{c|}{CLDA} &
		\multicolumn{1}{c|}{SCLUDA} &
		\multicolumn{1}{c|}{SSWADA} &
		\multicolumn{1}{c|}{CACL} &
		BiDA \\ \hline
		1 &
		\multicolumn{1}{c|}{73.05} &
		\multicolumn{1}{c|}{73.18} &
		\multicolumn{1}{c|}{88.94} &
		\multicolumn{1}{c|}{78.22} &
		\multicolumn{1}{c|}{79.38} &
		\multicolumn{1}{c|}{76.41} &
		\multicolumn{1}{c|}{82.32} &
		\multicolumn{1}{c|}{88.12} &
		\multicolumn{1}{c|}{82.49} &
		90.76 \\
		2 &
		\multicolumn{1}{c|}{82.51} &
		\multicolumn{1}{c|}{92.75} &
		\multicolumn{1}{c|}{93.84} &
		\multicolumn{1}{c|}{80.84} &
		\multicolumn{1}{c|}{93.31} &
		\multicolumn{1}{c|}{80.93} &
		\multicolumn{1}{c|}{73.53} &
		\multicolumn{1}{c|}{66.43} &
		\multicolumn{1}{c|}{71.03} &
		81.11 \\
		3 &
		\multicolumn{1}{c|}{80.65} &
		\multicolumn{1}{c|}{82.21} &
		\multicolumn{1}{c|}{45.15} &
		\multicolumn{1}{c|}{78.88} &
		\multicolumn{1}{c|}{48.89} &
		\multicolumn{1}{c|}{77.03} &
		\multicolumn{1}{c|}{84.08} &
		\multicolumn{1}{c|}{78.61} &
		\multicolumn{1}{c|}{76.45} &
		85.02 \\
		4 &
		\multicolumn{1}{c|}{23.23} &
		\multicolumn{1}{c|}{11.59} &
		\multicolumn{1}{c|}{10.22} &
		\multicolumn{1}{c|}{13.82} &
		\multicolumn{1}{c|}{10.03} &
		\multicolumn{1}{c|}{26.66} &
		\multicolumn{1}{c|}{8.12} &
		\multicolumn{1}{c|}{8.05} &
		\multicolumn{1}{c|}{20.12} &
		24.71 \\
		5 &
		\multicolumn{1}{c|}{60.72} &
		\multicolumn{1}{c|}{68.06} &
		\multicolumn{1}{c|}{88.53} &
		\multicolumn{1}{c|}{80.31} &
		\multicolumn{1}{c|}{48.75} &
		\multicolumn{1}{c|}{67.42} &
		\multicolumn{1}{c|}{40.84} &
		\multicolumn{1}{c|}{42.35} &
		\multicolumn{1}{c|}{54.87} &
		60.23 \\ \hline
		OA (\%) &
		\multicolumn{1}{c|}{68.61} &
		\multicolumn{1}{c|}{70.86} &
		\multicolumn{1}{c|}{72.41} &
		\multicolumn{1}{c|}{71.72} &
		\multicolumn{1}{c|}{66.01} &
		\multicolumn{1}{c|}{70.19} &
		\multicolumn{1}{c|}{66.21} &
		\multicolumn{1}{c|}{65.19} &
		\multicolumn{1}{c|}{68.49} &
		\textbf{75.08} \\ \hline
		KC ($\kappa$) &
		\multicolumn{1}{c|}{59.26} &
		\multicolumn{1}{c|}{62.16} &
		\multicolumn{1}{c|}{63.49} &
		\multicolumn{1}{c|}{63.97} &
		\multicolumn{1}{c|}{57.91} &
		\multicolumn{1}{c|}{61.27} &
		\multicolumn{1}{c|}{54.66} &
		\multicolumn{1}{c|}{53.44} &
		\multicolumn{1}{c|}{59.23} &
		\textbf{67.25} \\ \hline \hline
	\end{tabular}
\end{center}
\end{table*}


\begin{figure}[htp]
	\centering
	\setlength{\tabcolsep}{0.5em}
	\begin{tabular}{ccccccccccccc}
		\multicolumn{3}{c}{\epsfig{width=0.5\figurewidth,file=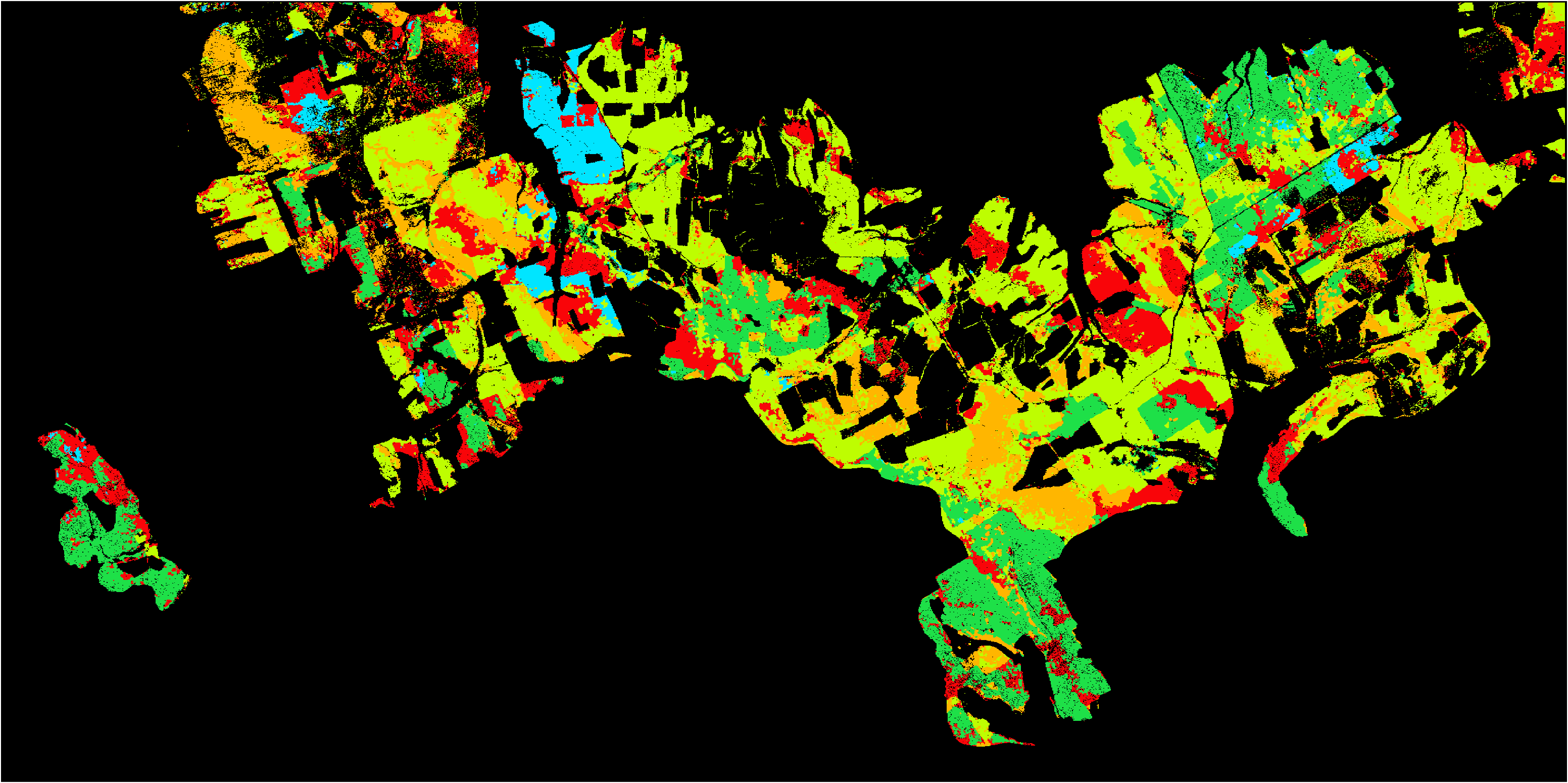}}     &
		\multicolumn{3}{c}{\epsfig{width=0.5\figurewidth,file=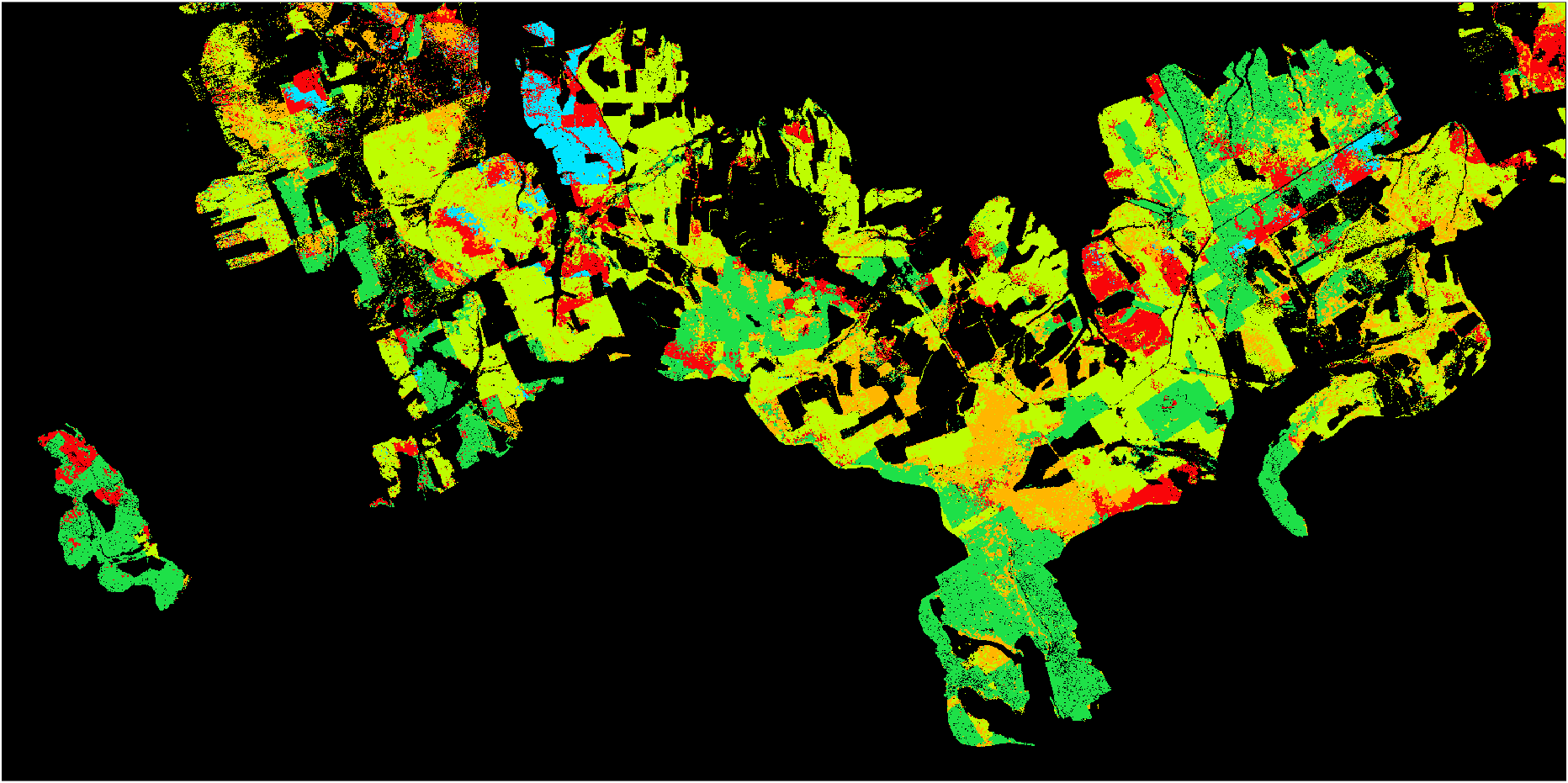}}  \\
		\multicolumn{3}{c}{(a)} & \multicolumn{3}{c}{(b)} & \\
		\multicolumn{3}{c}{\epsfig{width=0.5\figurewidth,file=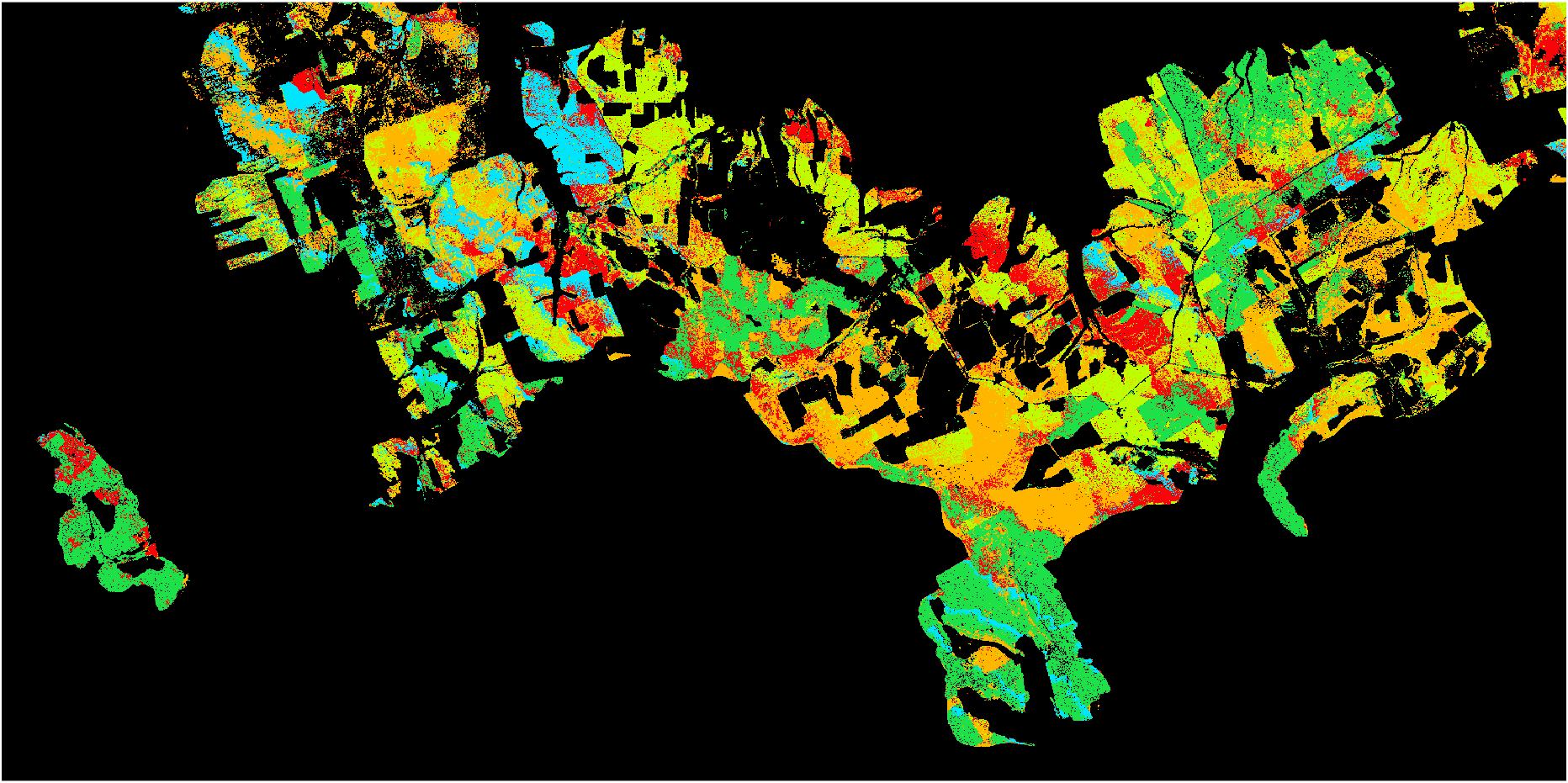}}    &
		\multicolumn{3}{c}{\epsfig{width=0.5\figurewidth,file=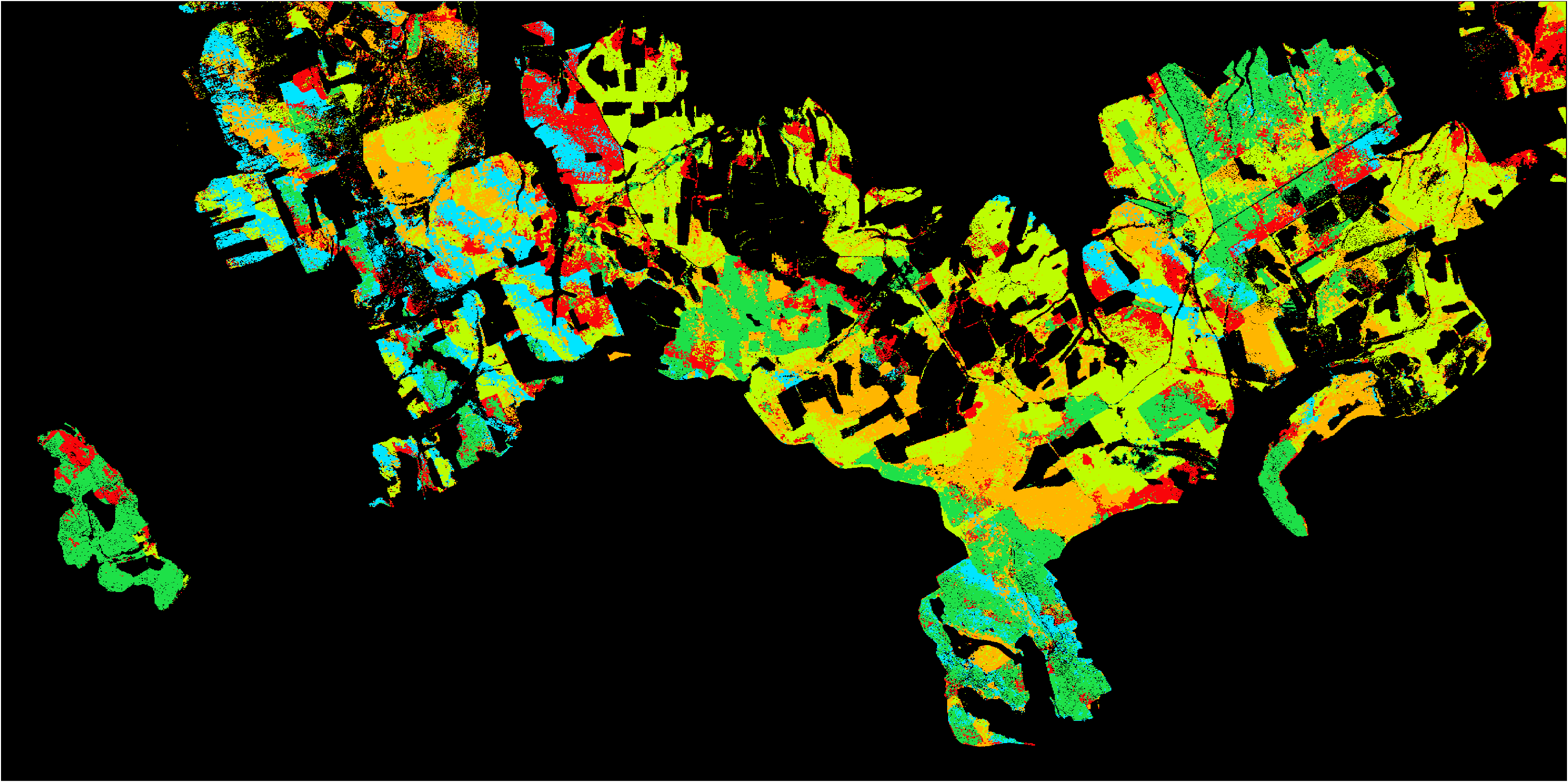}} \\				
		\multicolumn{3}{c}{(c)}  & \multicolumn{3}{c}{(d)} &    \\
		\multicolumn{3}{c}{\epsfig{width=0.5\figurewidth,file=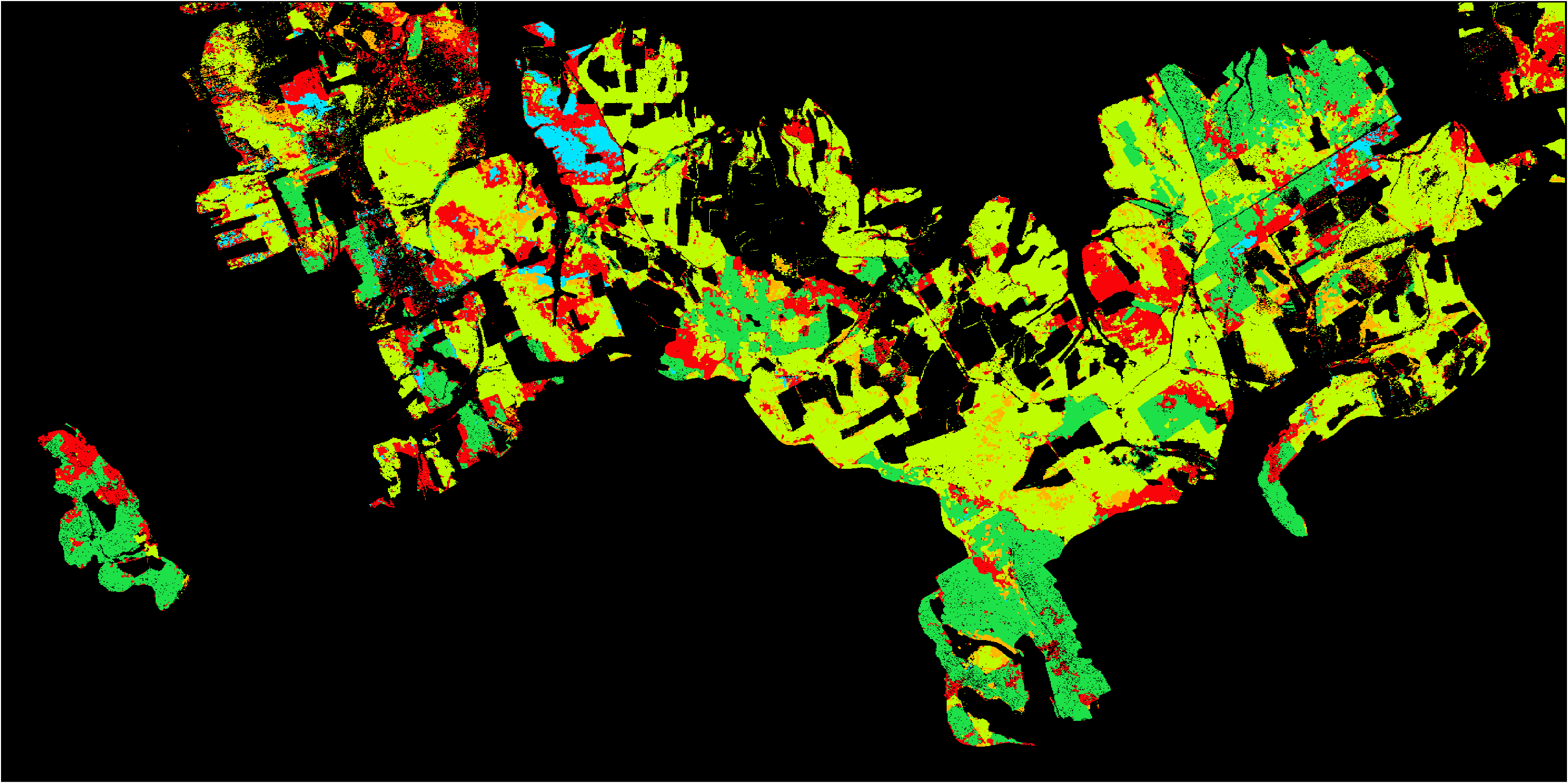}} &
		\multicolumn{3}{c}{\epsfig{width=0.5\figurewidth,file=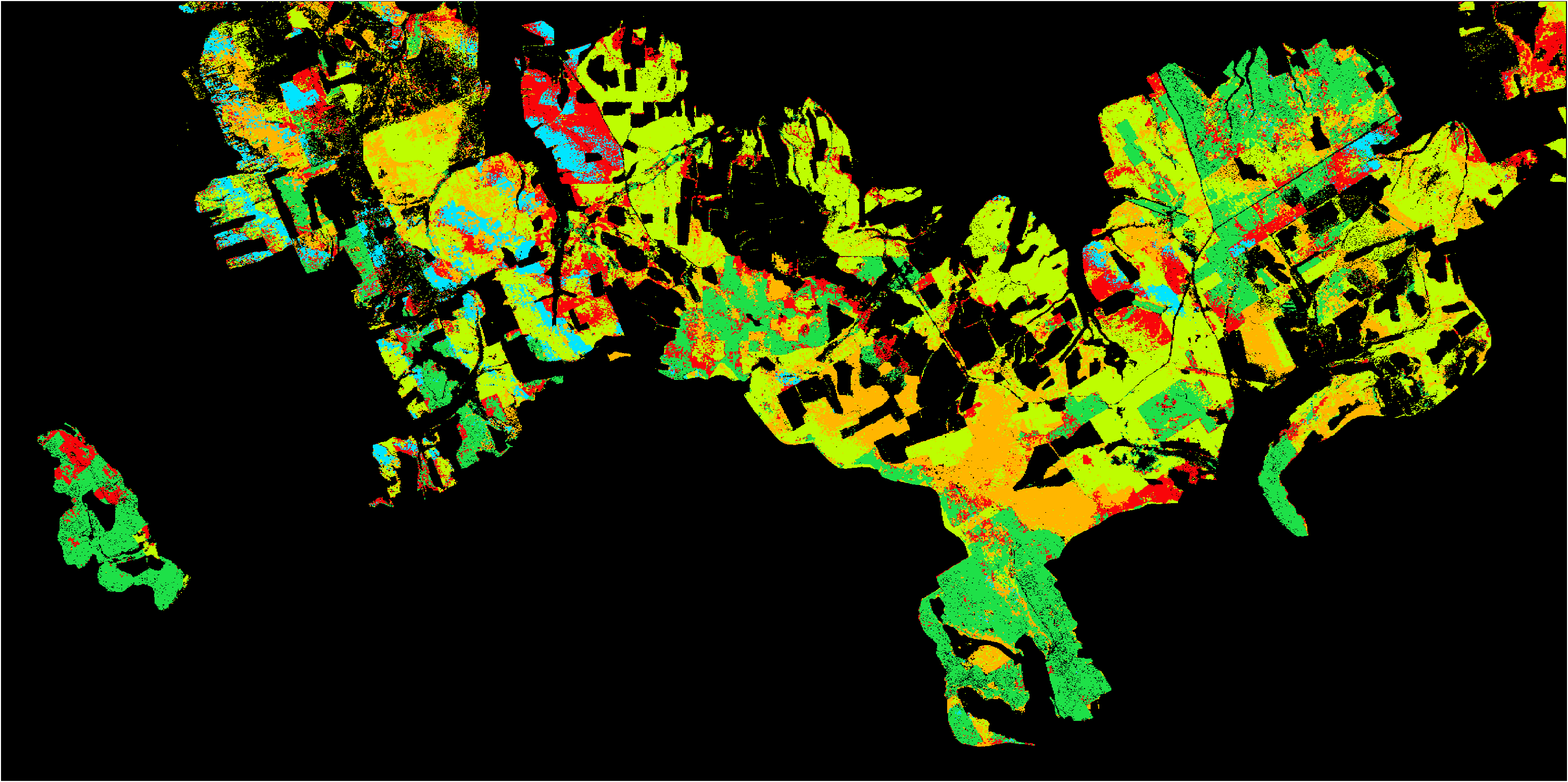}}    \\
		\multicolumn{3}{c}{(e)} & \multicolumn{3}{c}{(f)} & \\
		\multicolumn{3}{c}{\epsfig{width=0.5\figurewidth,file=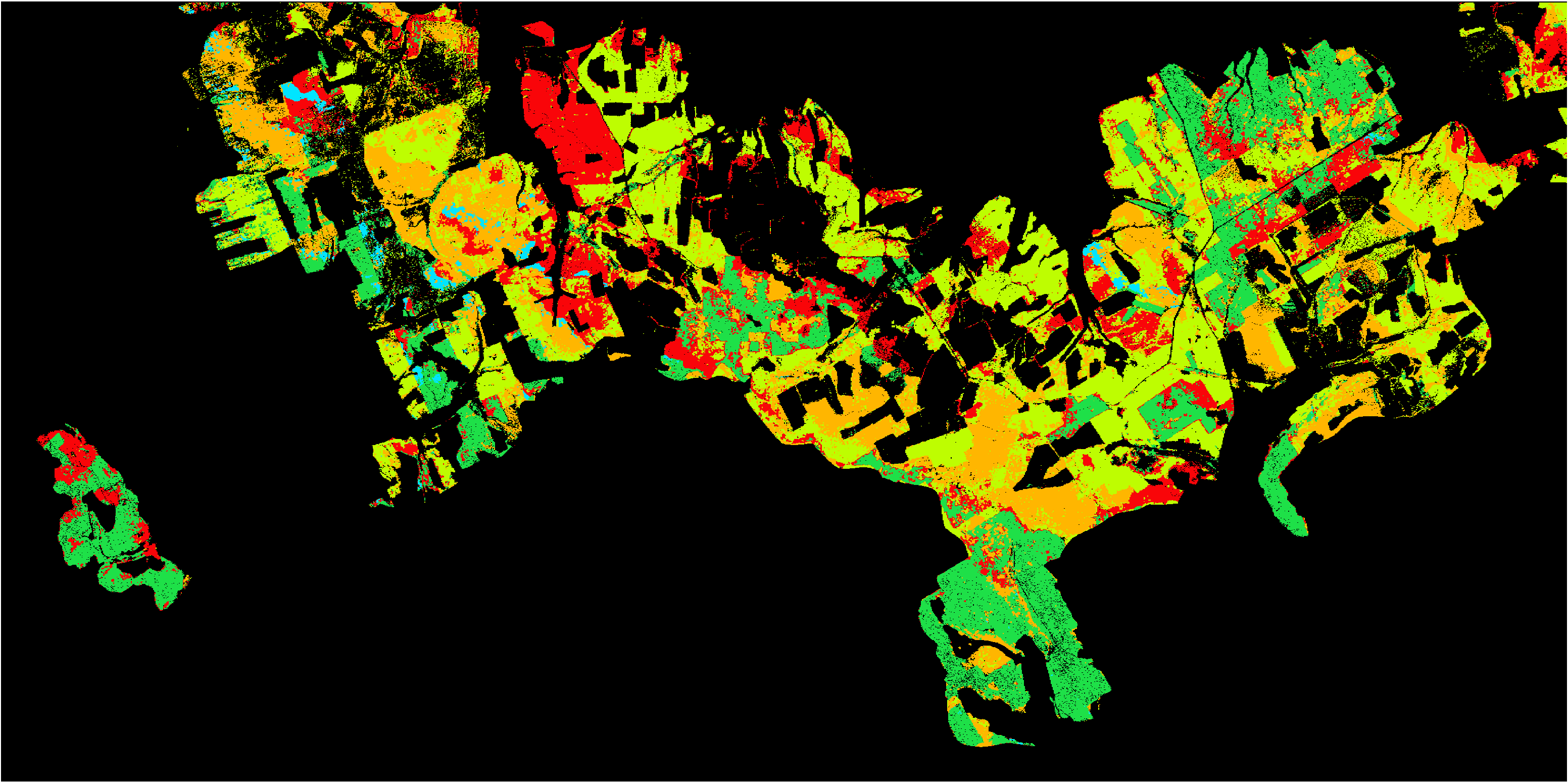}} &
		\multicolumn{3}{c}{\epsfig{width=0.5\figurewidth,file=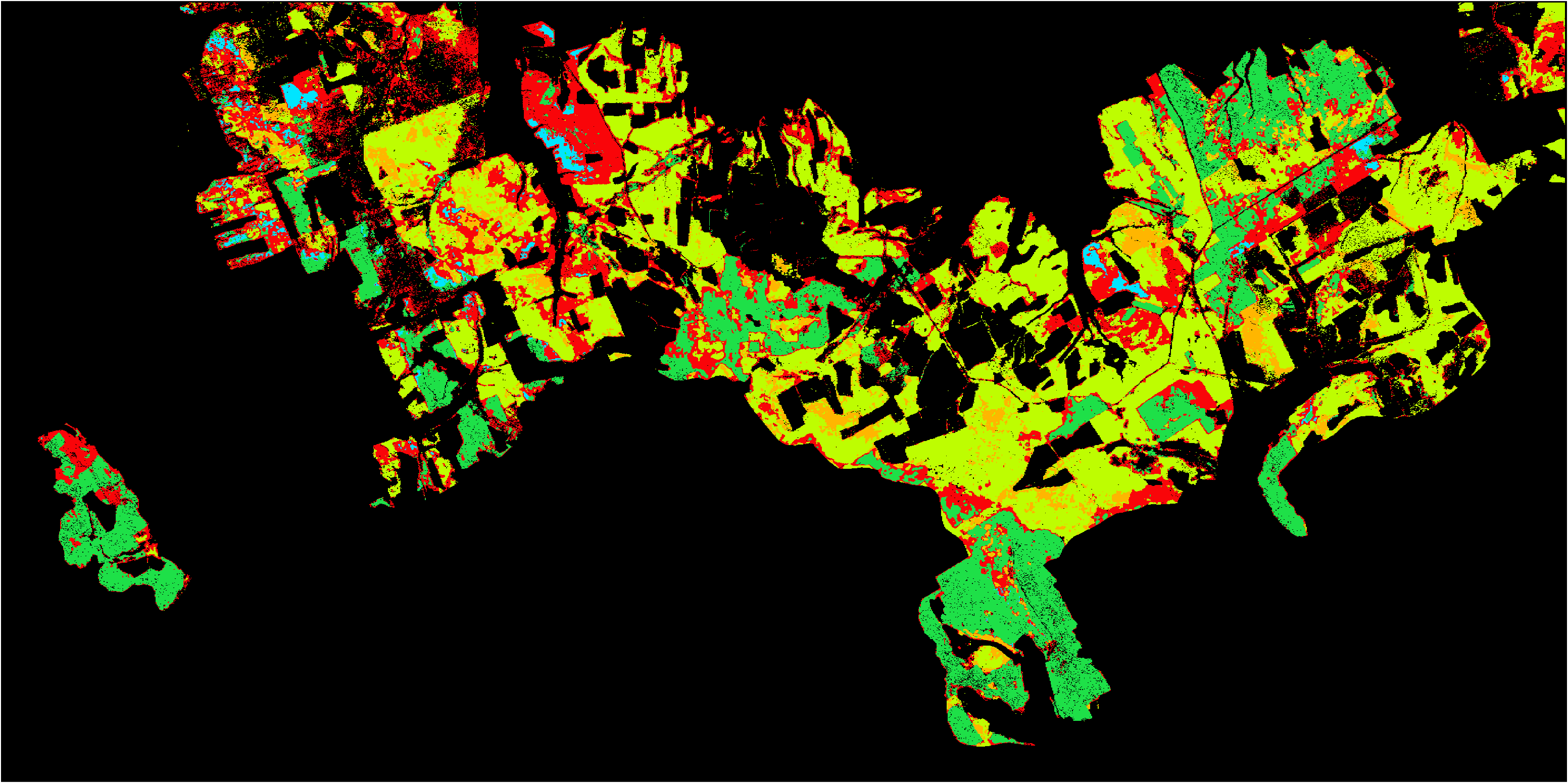}}    \\	
		\multicolumn{3}{c}{(g)}  & \multicolumn{3}{c}{(h)} &    \\
		\multicolumn{3}{c}{\epsfig{width=0.5\figurewidth,file=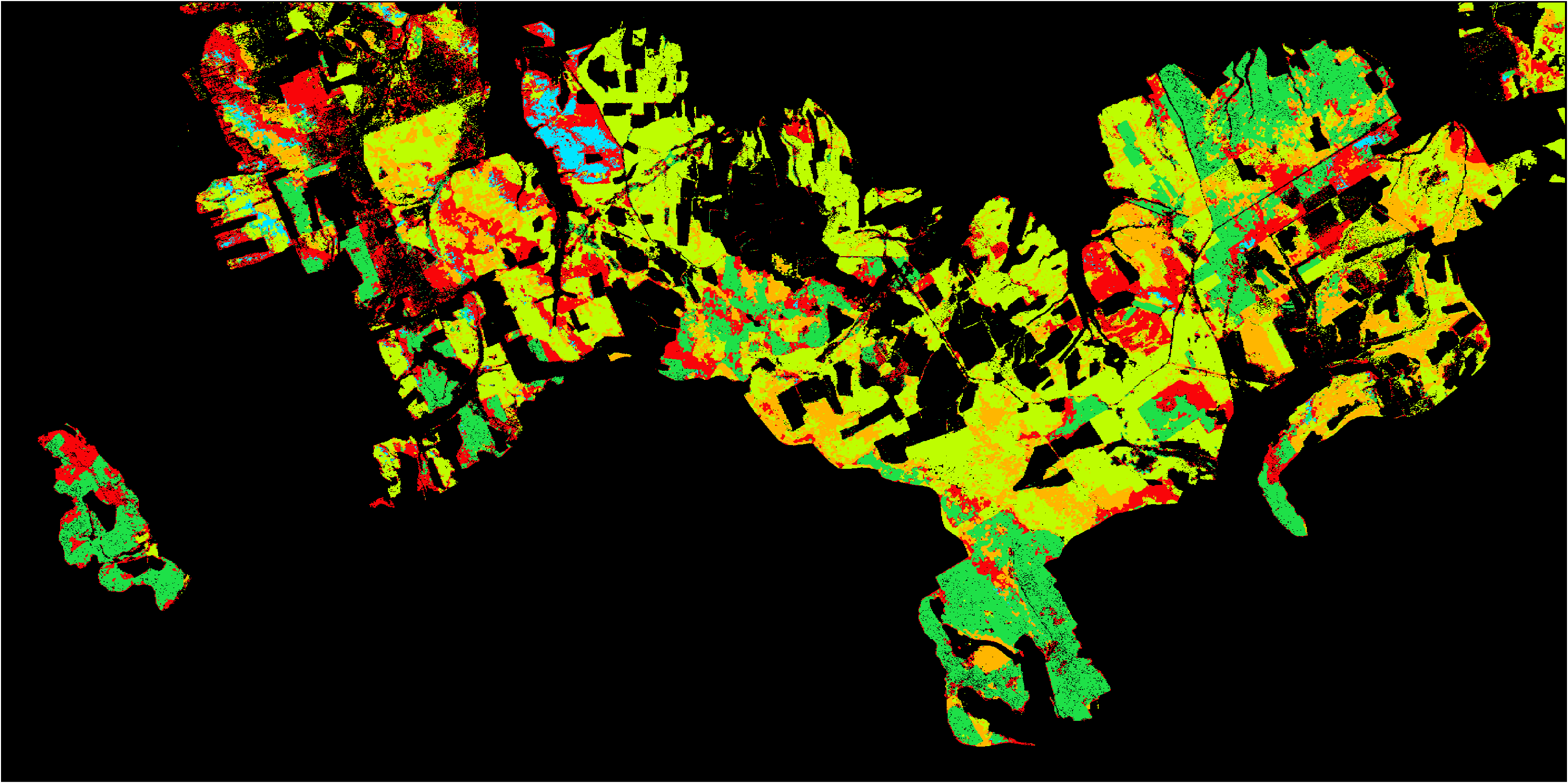}} & \multicolumn{3}{c}{\epsfig{width=0.5\figurewidth,file=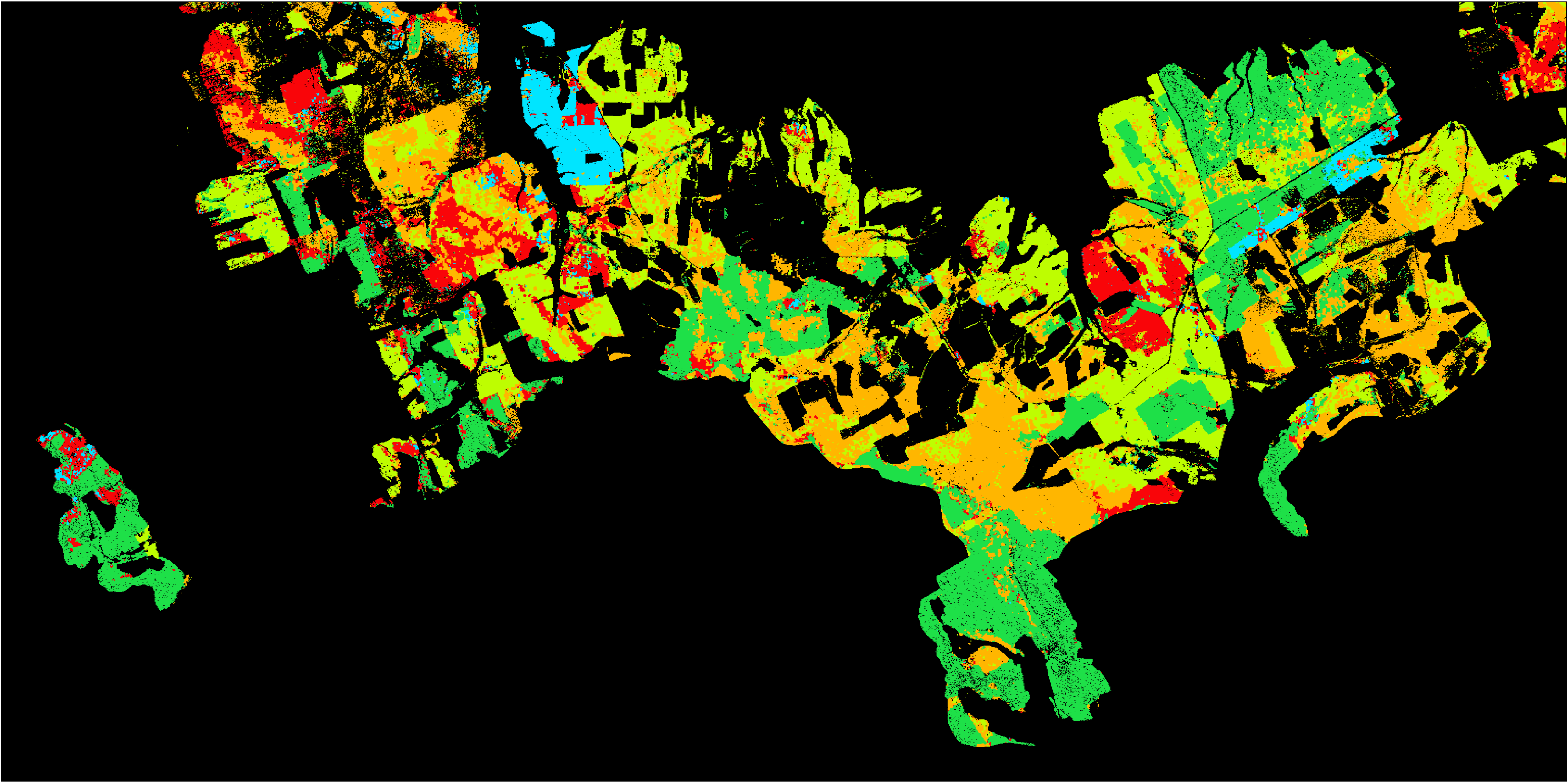}} \\
		\multicolumn{3}{c}{(i)}  & \multicolumn{3}{c}{(j)} &    \\
	\end{tabular}
	\caption{\label{fig:Map_TD1}
		Visualization and classification maps for the target scene MFF TD1 obtained with different methods including: (a) GAHT (67.38\%), (b) MLUDA (72.80\%), (c) MSDA (68.01\%), (d) TSTnet (68.21\%), (e) MDGTnet (65.24\%), (f) CLDA (68.26\%), (g) SCLUDA (62.72\%), (h) SSWADA (56.39\%), (i) CACL (66.86\%), (j) BiDA (77.40\%).}
\end{figure}

%

\begin{figure}[htp]
	\centering
	\begin{tabular}{ccccccccccccc}
	\multicolumn{3}{c}{\epsfig{width=0.5\figurewidth,file=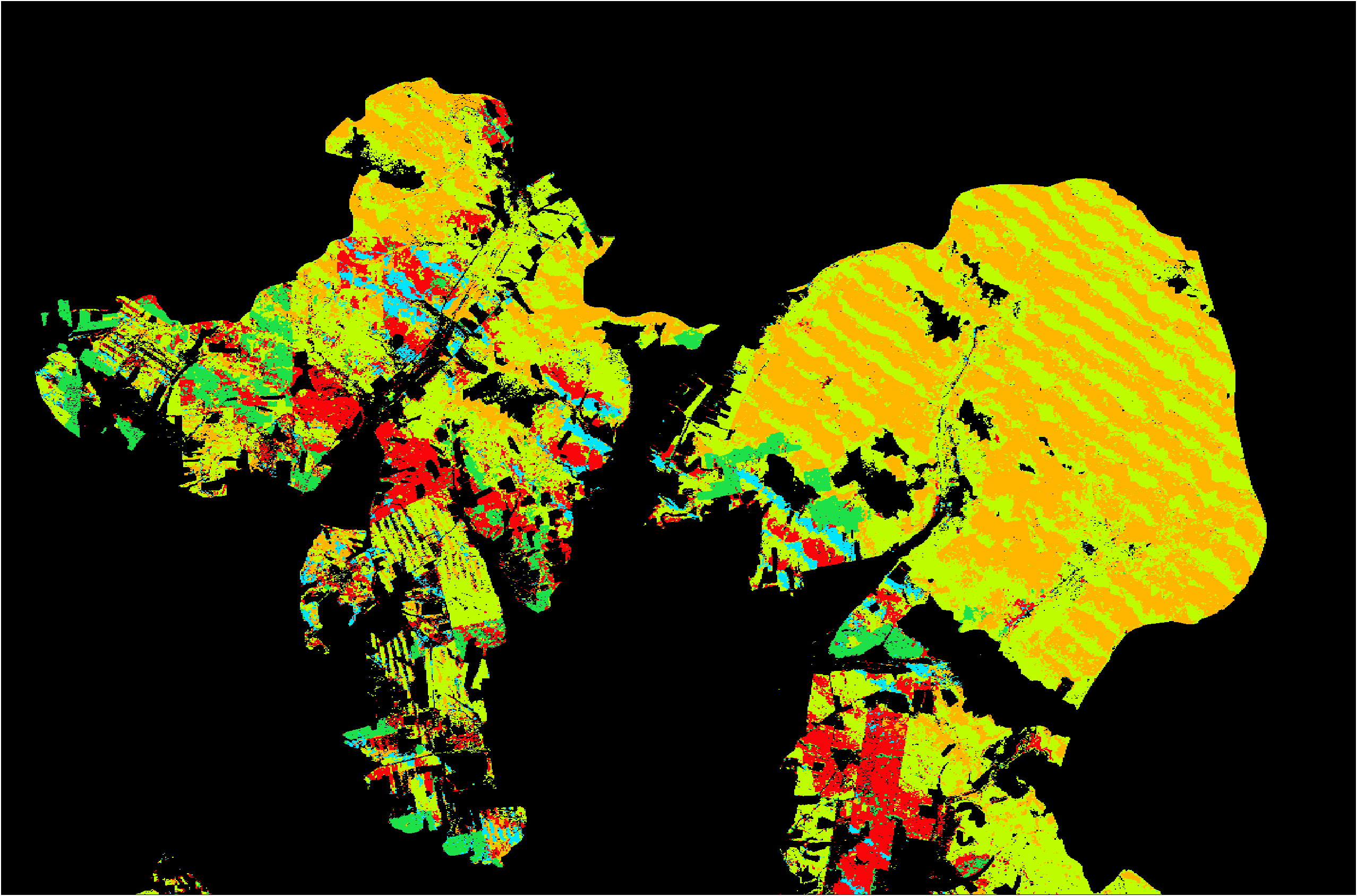}}     &
	\multicolumn{3}{c}{\epsfig{width=0.5\figurewidth,file=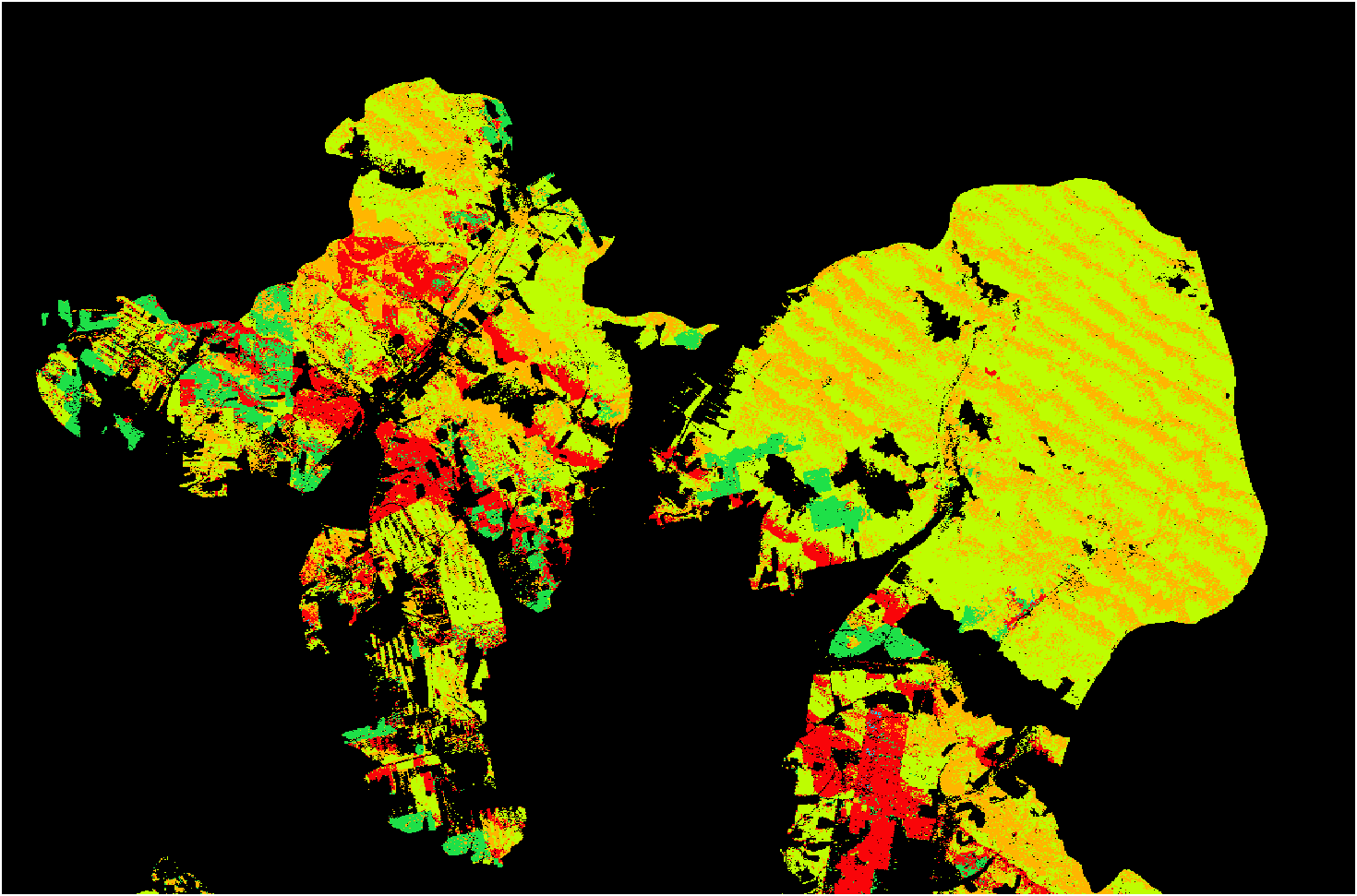}}  \\
	\multicolumn{3}{c}{(a)} & \multicolumn{3}{c}{(b)} & \\
	\multicolumn{3}{c}{\epsfig{width=0.5\figurewidth,file=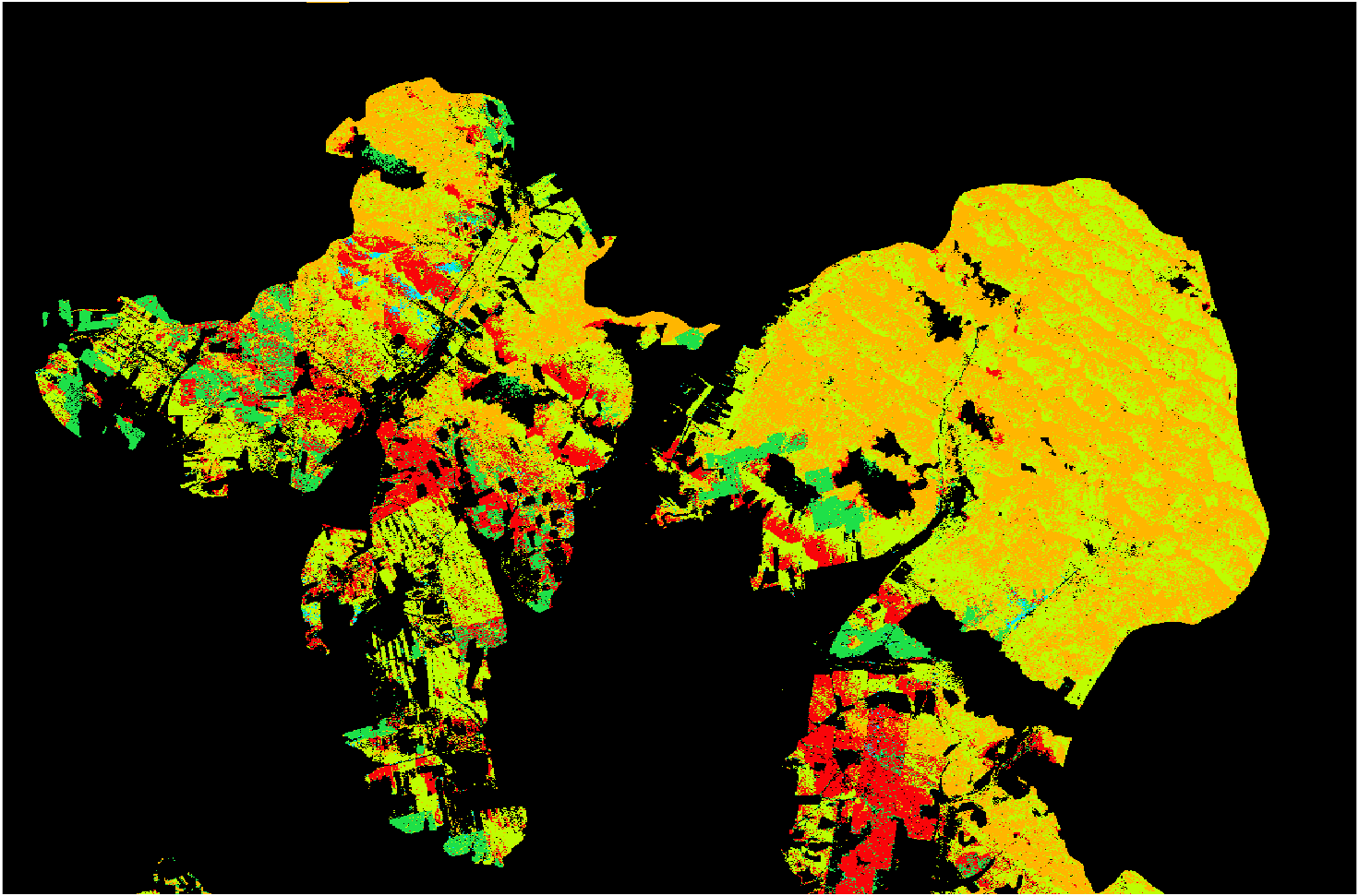}}    &
	\multicolumn{3}{c}{\epsfig{width=0.5\figurewidth,file=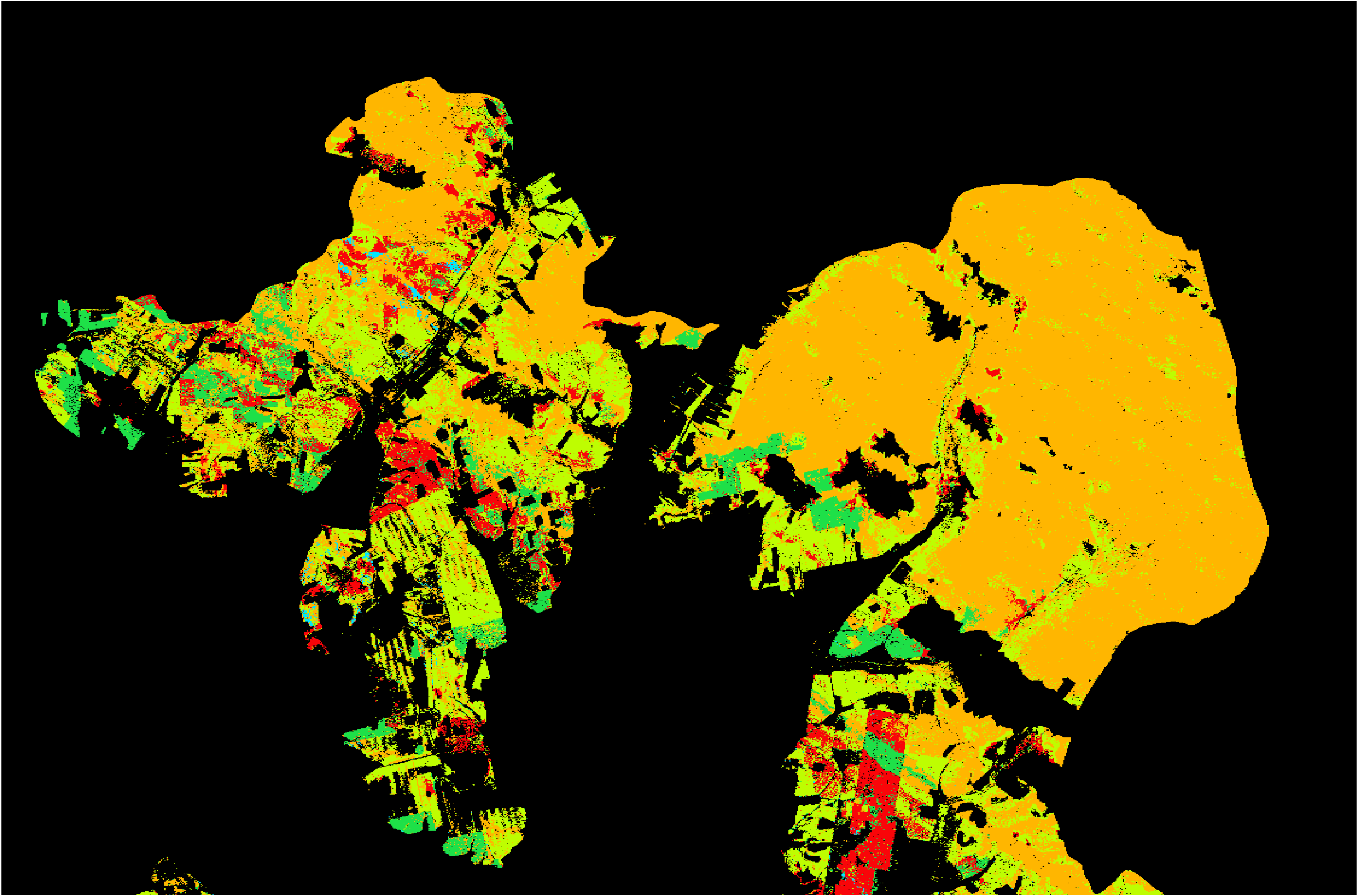}} \\				
	\multicolumn{3}{c}{(c)}  & \multicolumn{3}{c}{(d)} &    \\
	\multicolumn{3}{c}{\epsfig{width=0.5\figurewidth,file=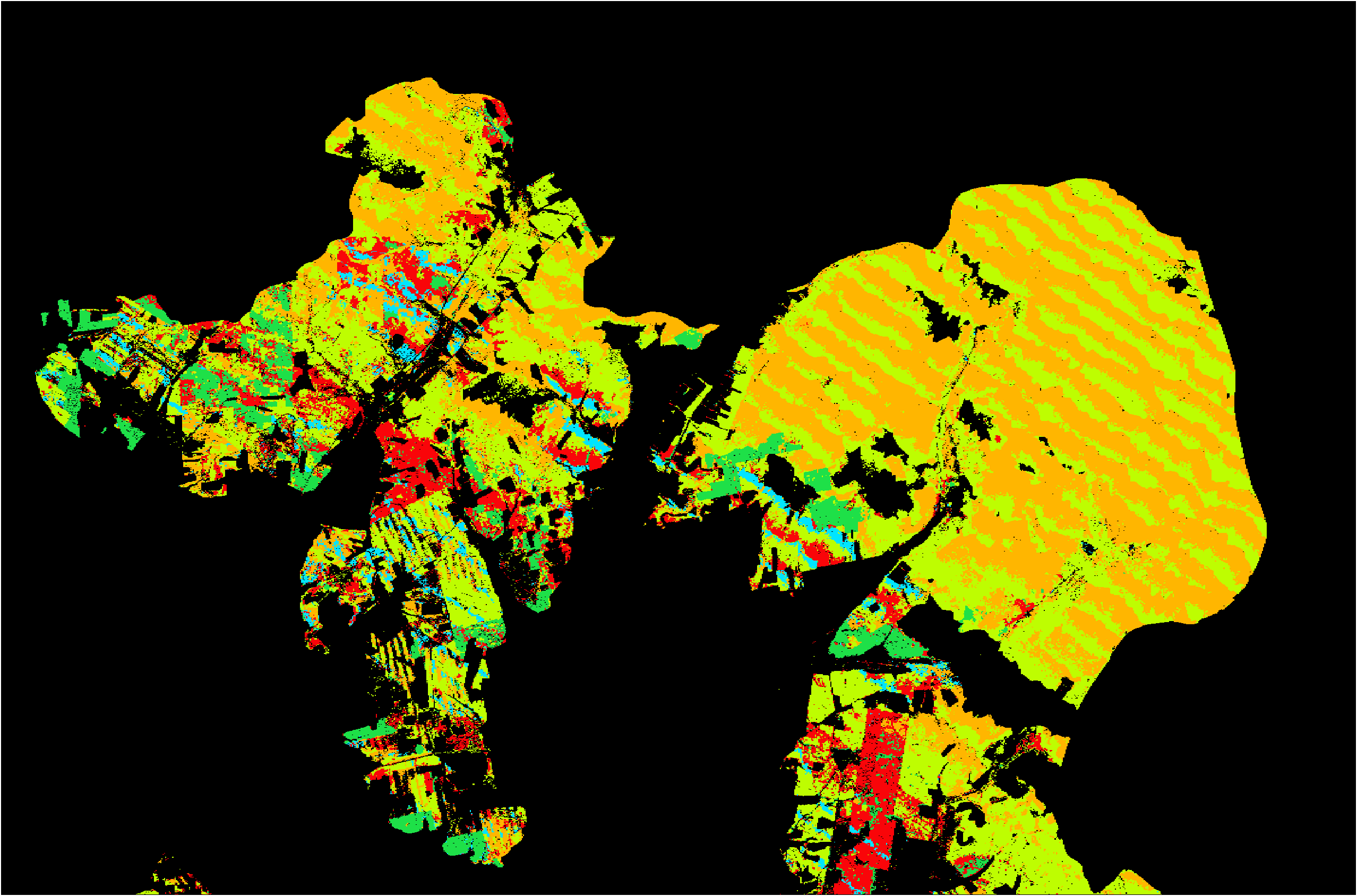}} &
	\multicolumn{3}{c}{\epsfig{width=0.5\figurewidth,file=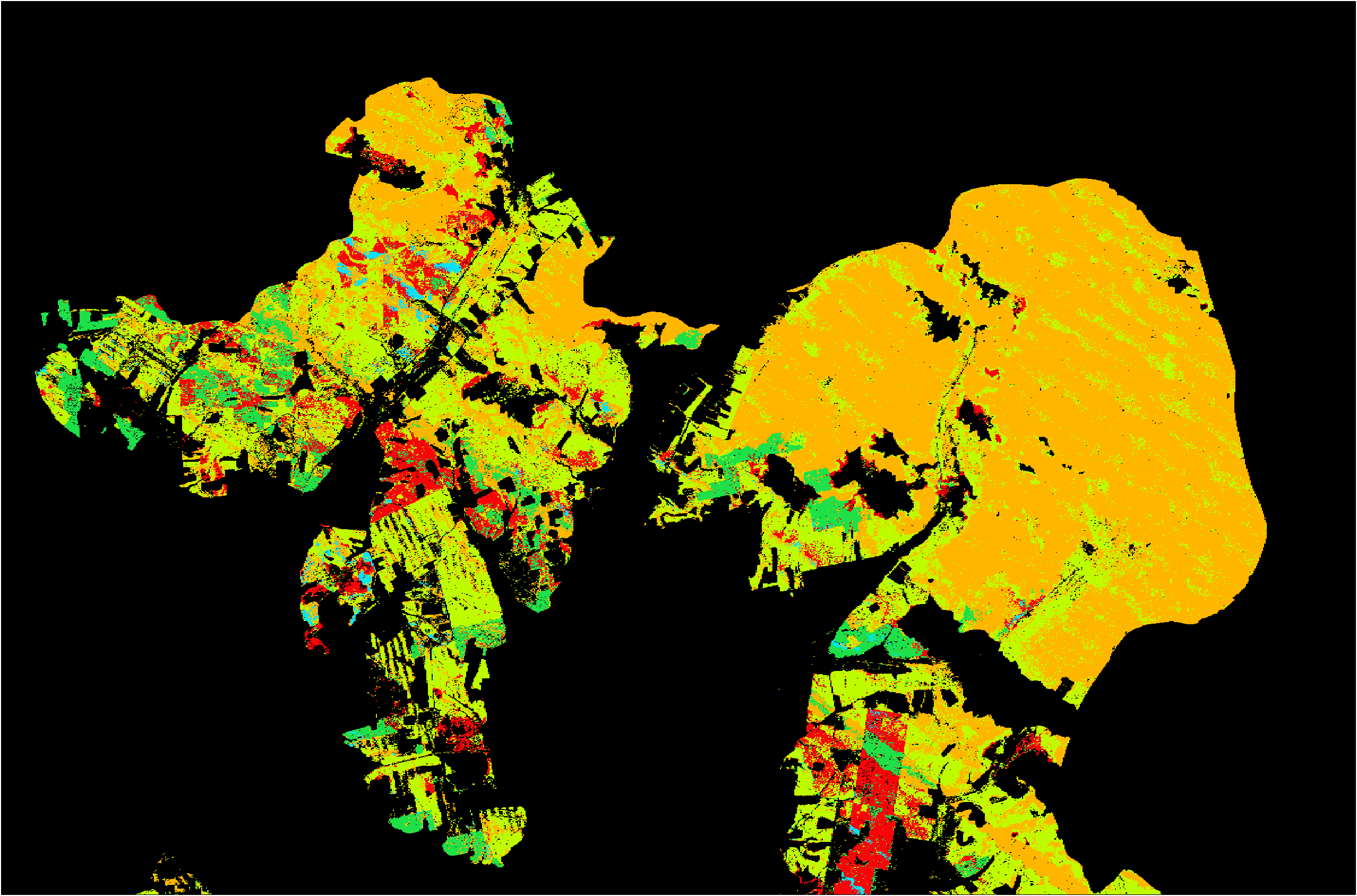}}    \\
	\multicolumn{3}{c}{(e)} & \multicolumn{3}{c}{(f)} & \\
	\multicolumn{3}{c}{\epsfig{width=0.5\figurewidth,file=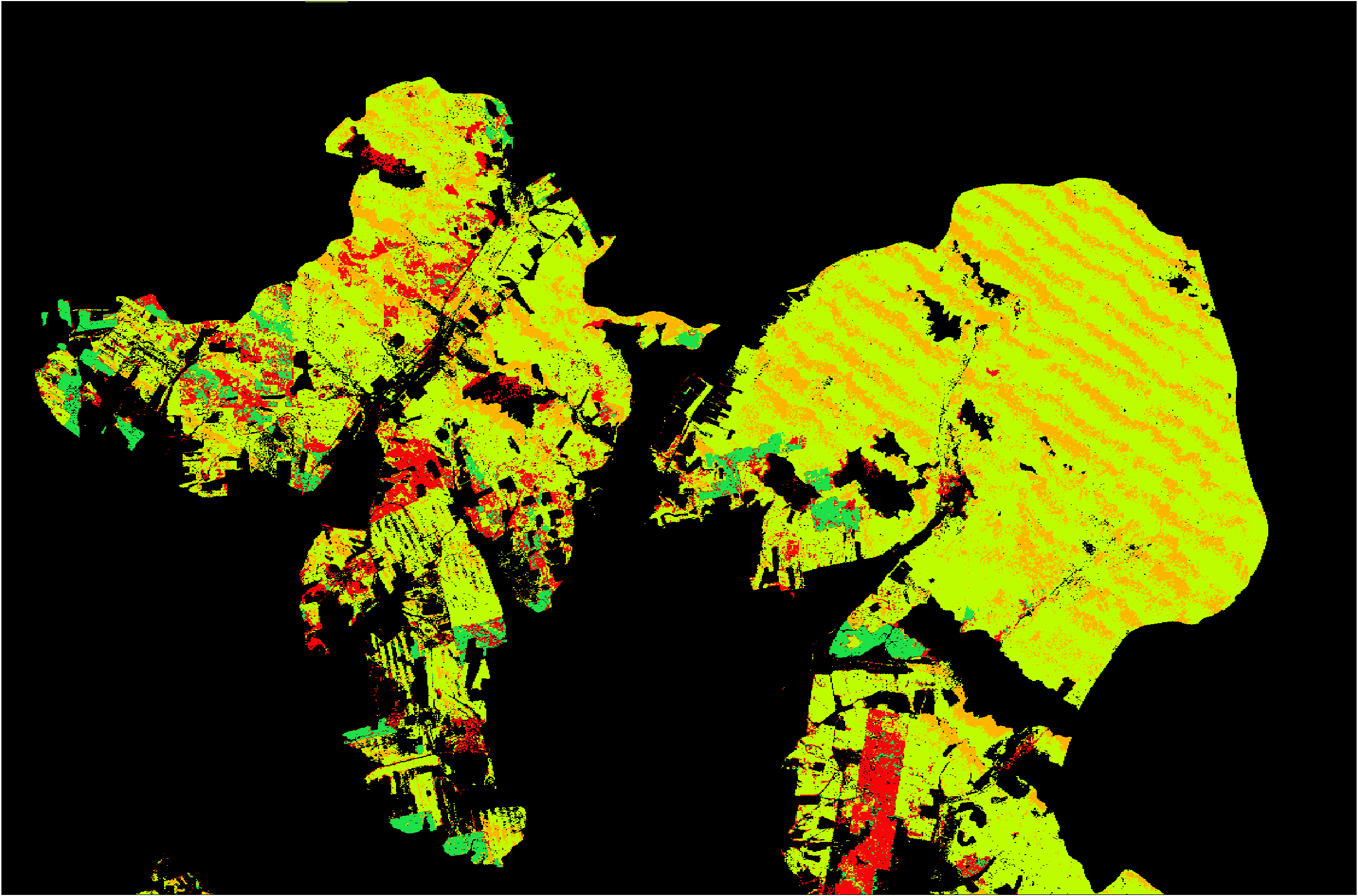}} &
	\multicolumn{3}{c}{\epsfig{width=0.5\figurewidth,file=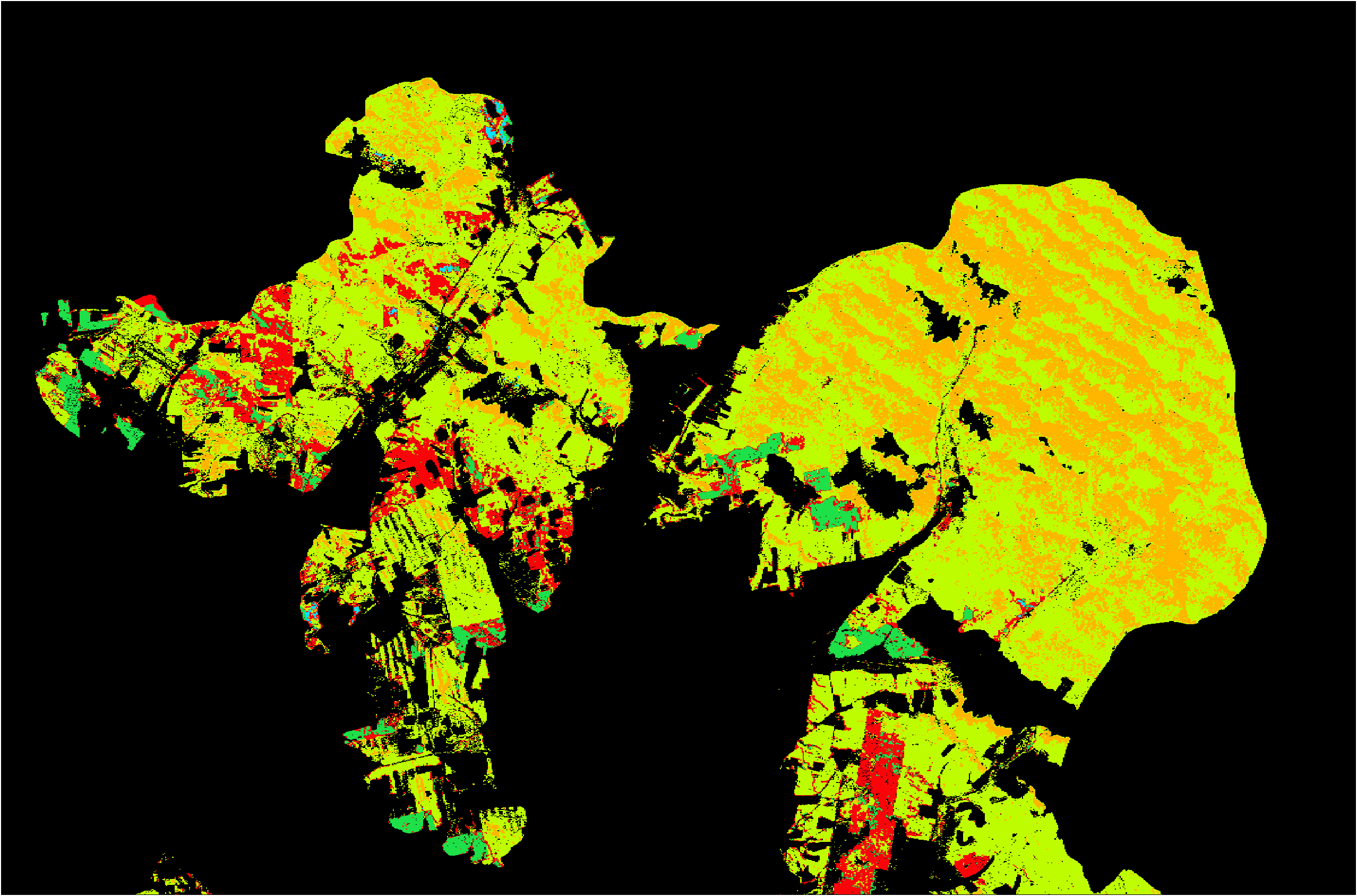}}    \\	
	\multicolumn{3}{c}{(g)}  & \multicolumn{3}{c}{(h)} &    \\
	\multicolumn{3}{c}{\epsfig{width=0.5\figurewidth,file=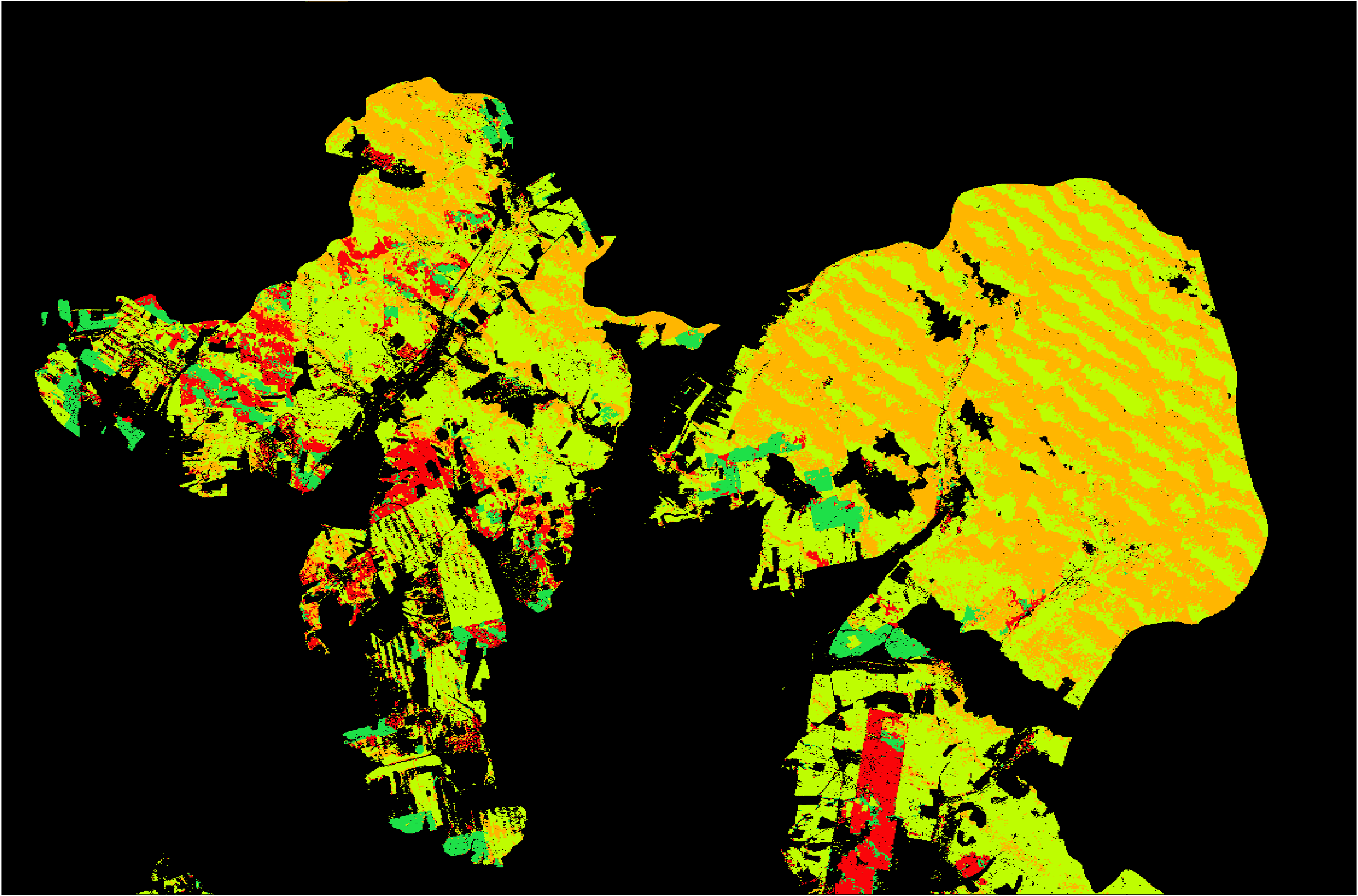}} & \multicolumn{3}{c}{\epsfig{width=0.5\figurewidth,file=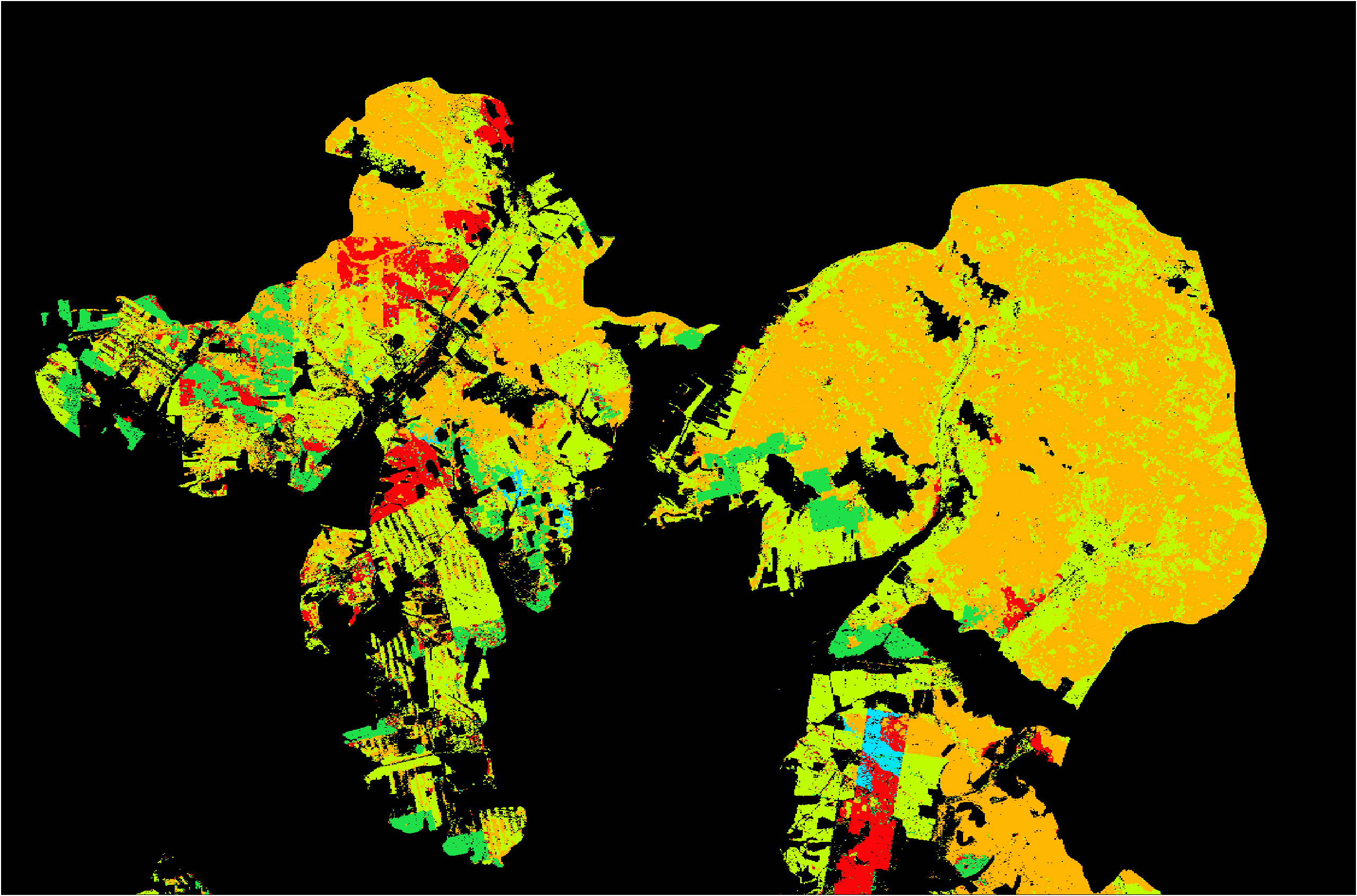}} \\
	\multicolumn{3}{c}{(i)}  & \multicolumn{3}{c}{(j)} &    \\
\end{tabular}
	\begin{tabular}{cc}
		\epsfig{width=1\figurewidth,file=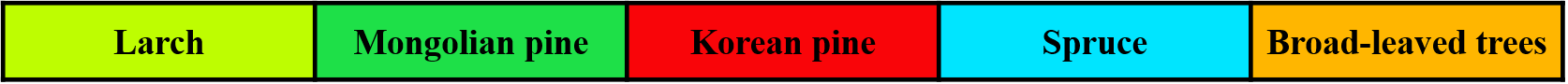}\\
	\end{tabular}
	\caption{\label{fig:Map_TD2}
		Visualization and classification maps for the target scene MFF TD2 obtained with different methods including: (a) GAHT (68.61\%), (b) MLUDA (70.86\%), (c) MSDA (72.41\%), (d) TSTnet (68.21\%), (e) MDGTnet (66.01\%), (f) CLDA (70.19\%), (g) SCLUDA (66.21\%), (h) SSWADA (65.19\%), (i) CACL (68.49\%), (j) BiDA (75.08\%).}
\end{figure}

\begin{figure*}[tp] \small
	\begin{center}
		\begin{tabular}{cccc}
			\epsfig{ width=0.48\figurewidth,file=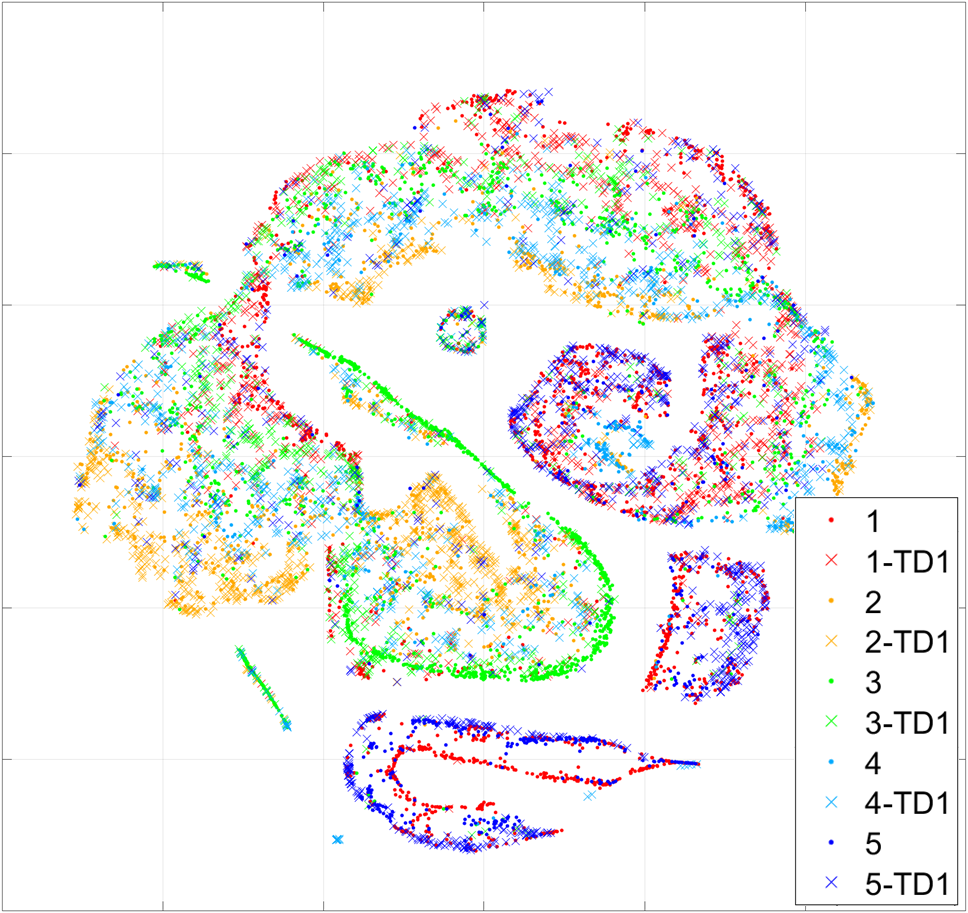} &
			\epsfig{width=0.5\figurewidth,file=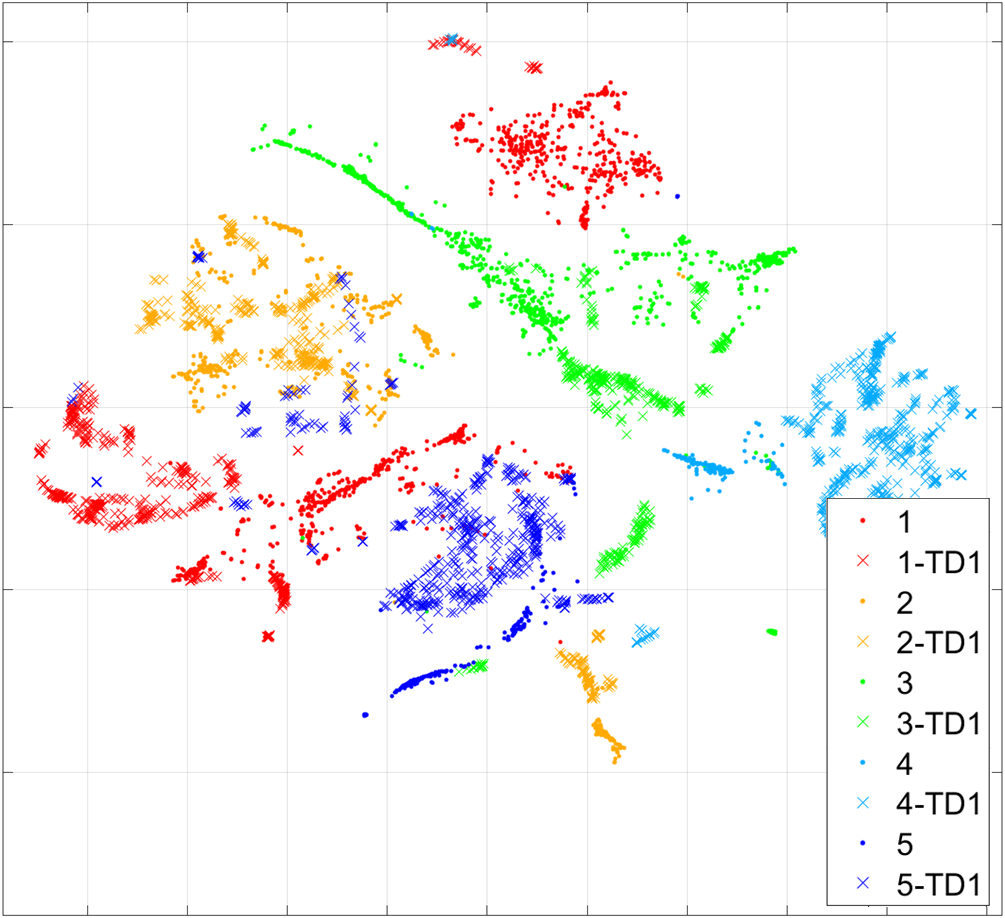}&
			\epsfig{width=0.5\figurewidth,file=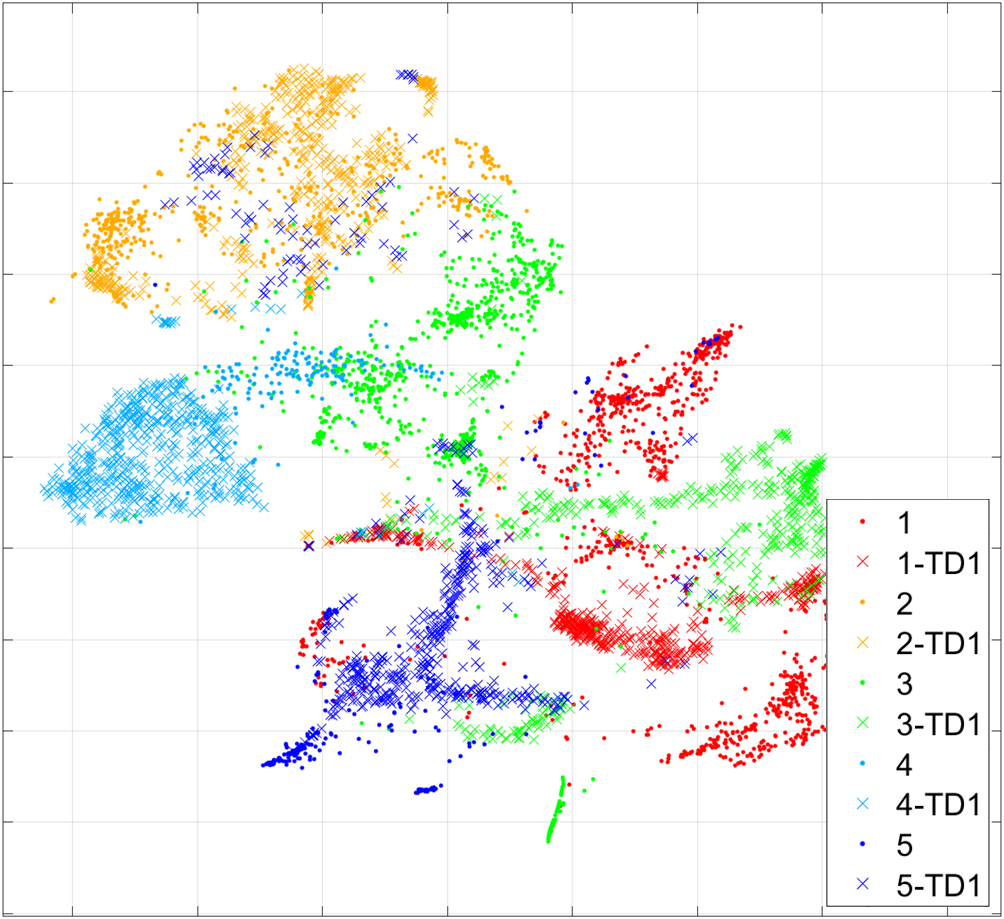}&
			\epsfig{width=0.48\figurewidth,file=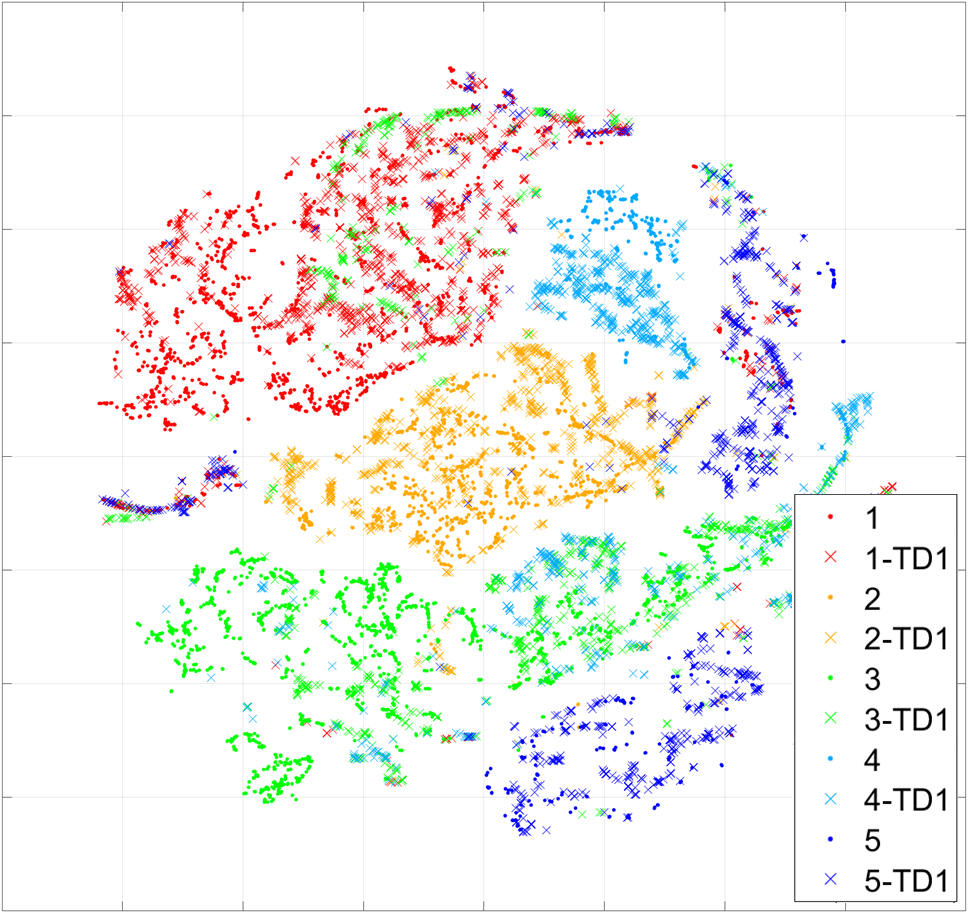}   \\
			(a) OS of the MFF SD\&TD1  & (b) AF from MLUDA & (c) AF from MSDA & (d) AF from BiDA\\  [0.5em]
		\end{tabular}
		\begin{tabular}{cccc}
		 \epsfig{width=0.48\figurewidth,file=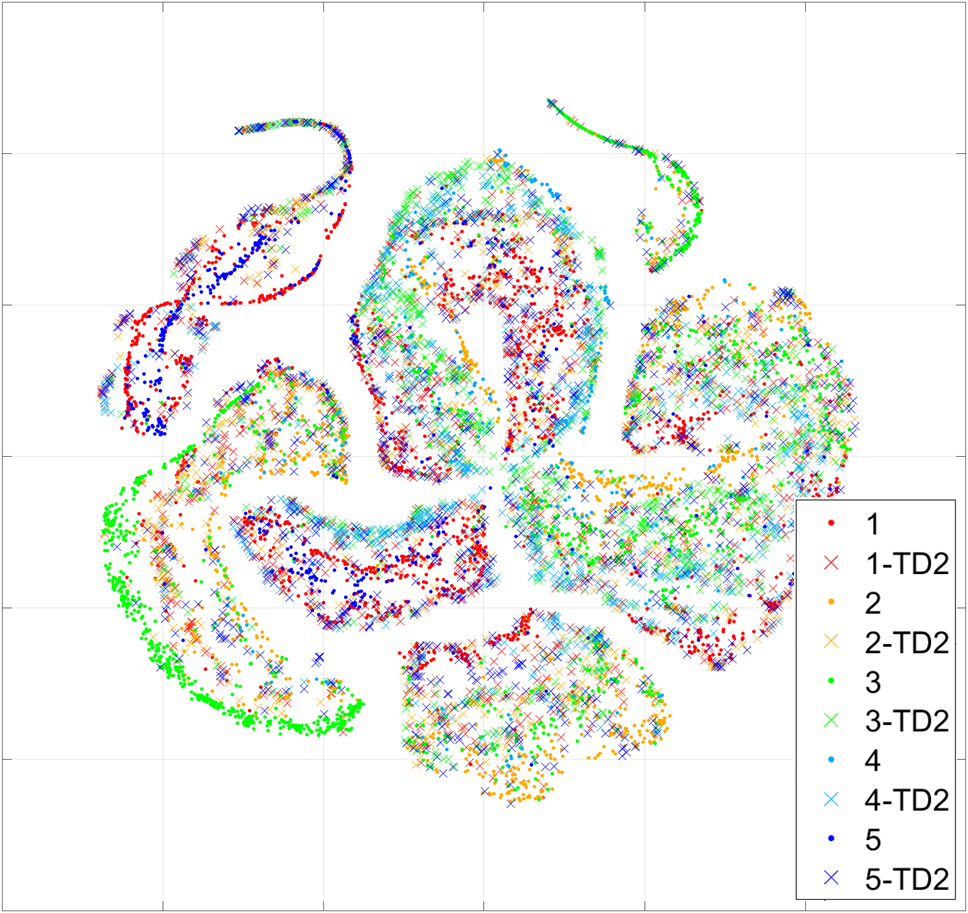} &
			\epsfig{width=0.5\figurewidth,file=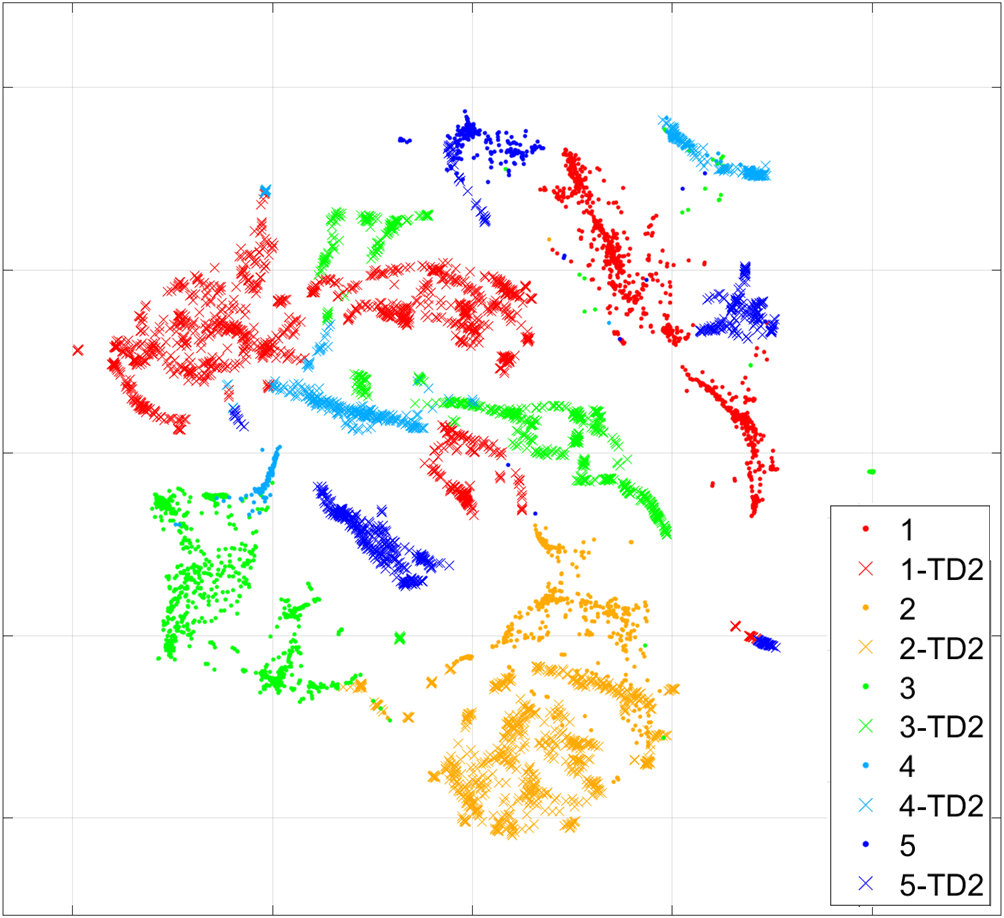}&
			\epsfig{width=0.5\figurewidth,file=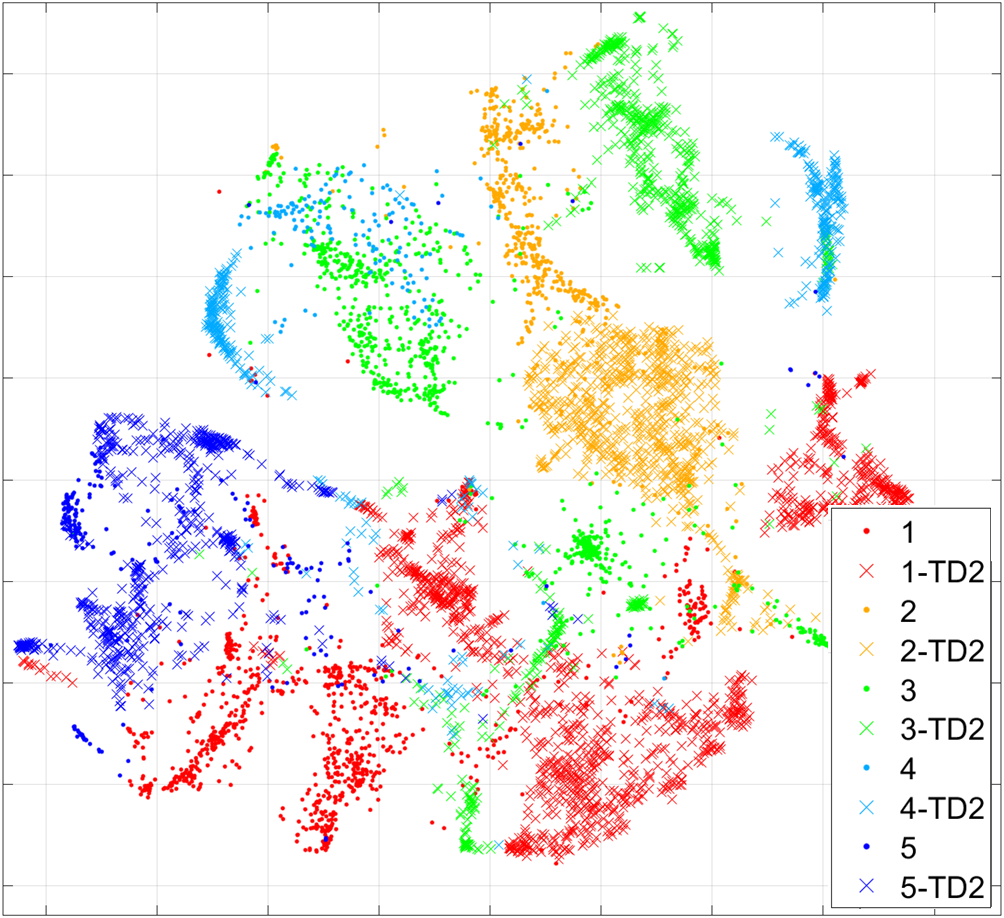}&
			\epsfig{width=0.488\figurewidth,file=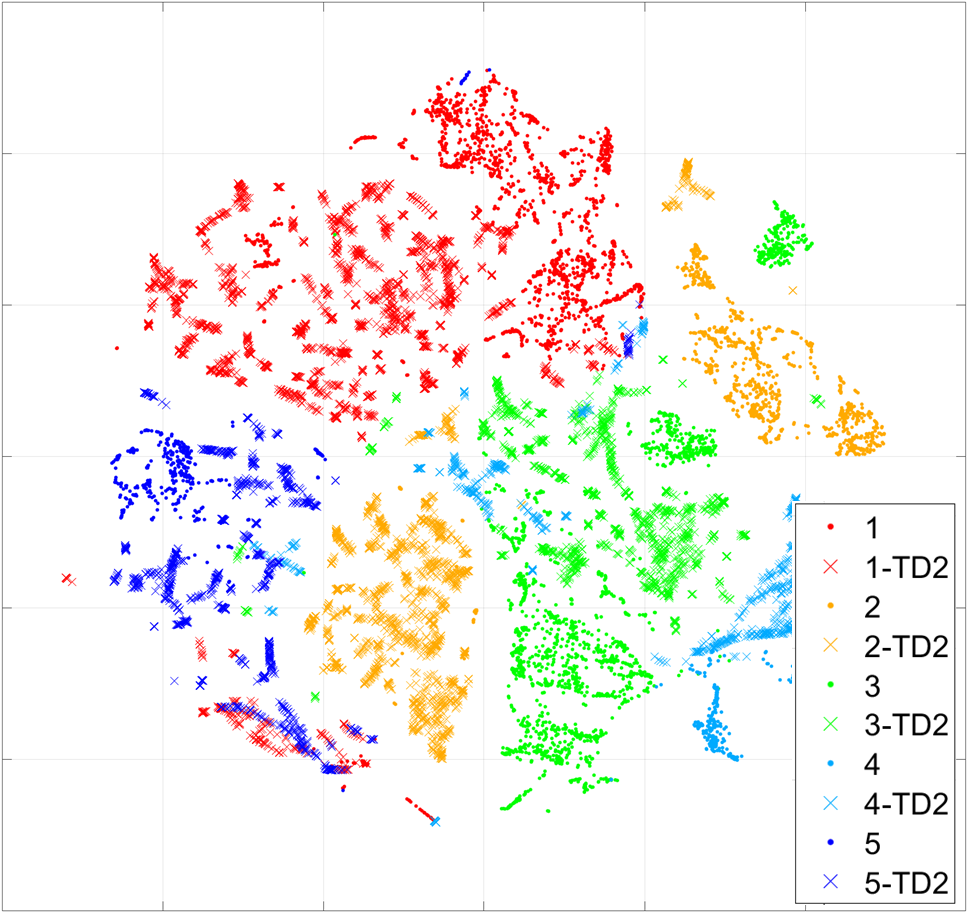}   \\
			(e) OS of the MFF SD\&TD2  & (f) AF from MLUDA & (g) AF from MSDA & (h) AF from BiDA\\
			\\  [0.5em]
		\end{tabular}
		\caption{\label{fig:analysis}
			 Alignment performance of the proposed BiDA using the MFF SD \& MFF TD1 and the MFF SD \& MFF TD2, $\bullet$ represents SD, × represents TD1/TD2, and the number represents class index (OS-Original samples, AF-Aligned features). In original space (a) and (e), there are obvious spectral shift between domains and significant overlap of the intra-domain samples of each class. MLUDA (b,f) and MSDA (c,g) alleviate spectral shift to a certain extent, but the separability was insufficient. BiDA (d,h) significantly improves inter-class separability and spectral shift in adaptive space.}  
	\end{center}
\end{figure*}

\begin{table*}[tp]
	\caption{\label{tab:accuracy_Hou}
	Class-specific and overall classification accuracy (\%) of different methods for the target scene Houston 2018 data.}
	\begin{center}
	\begin{tabular}{|c|c|c|c|c|c|c|c|c|c|c|}
		\hline \hline
		No.           & GAHT  & MLUDA & MSDA  & TSTnet & MDGTnet & CLDA  & SCLUDA & SSWADA & CACL  & BiDA           \\ \hline
		1             & 23.21 & 83.44 & 73.32 & 66.52  & 76.78   & 68.07 & 93.94  & 63.34  & 42.34 & 82.41          \\
		2             & 69.74 & 68.78 & 86.93 & 78.89  & 79.82   & 85.45 & 54.34  & 62.81  & 81.65 & 50.25          \\
		3             & 67.61 & 58.10 & 71.98 & 54.75  & 57.30   & 60.85 & 57.85  & 67.03  & 46.01 & 53.98          \\
		4             & 81.82 & 100   & 86.36 & 100    & 100     & 72.73 & 86.36  & 86.36  & 65.41 & 100            \\
		5             & 92.37 & 84.62 & 86.14 & 51.63  & 93.92   & 93.57 & 75.87  & 69.29  & 71.05 & 92.82          \\
		6             & 76.50 & 85.51 & 77.25 & 86.83  & 77.39   & 69.72 & 87.30  & 82.00  & 90.47 & 87.57          \\
		7             & 47.18 & 56.96 & 81.73 & 70.74  & 63.57   & 77.75 & 60.02  & 61.76  & 45.37 & 73.48          \\ \hline
		OA (\%)       & 72.15 & 78.97 & 79.41 & 79.26  & 76.57   & 74.02 & 78.61  & 75.29  & 79.10 & \textbf{81.11} \\ \hline
		KC ($\kappa$) & 57.11 & 64.86 & 65.28 & 65.11  & 63.37   & 61.54 & 64.63  & 62.69  & 65.03 & \textbf{69.48} \\ \hline \hline
	\end{tabular}
	\end{center}
\end{table*}

\begin{table*}[tp]
	\caption{\label{tab:accuracy_HyRANK}
	Class-specific and overall classification accuracy (\%) of different methods for the target scene Loukia data.}
		\begin{center}
	\begin{tabular}{|c|cccccccccc|}
		\hline \hline
		\multirow{2}{*}{Class} &
		\multicolumn{10}{c|}{Classification algorithms} \\ \cline{2-11} 
		&
		\multicolumn{1}{c|}{GAHT} &
		\multicolumn{1}{c|}{MLUDA} &
		\multicolumn{1}{c|}{MSDA} &
		\multicolumn{1}{c|}{TSTnet} &
		\multicolumn{1}{c|}{MDGTnet} &
		\multicolumn{1}{c|}{CLDA} &
		\multicolumn{1}{c|}{SCLUDA} &
		\multicolumn{1}{c|}{SSWADA} &
		\multicolumn{1}{c|}{CACL} &
		BiDA \\ \hline
		1 &
		\multicolumn{1}{c|}{14.56} &
		\multicolumn{1}{c|}{6.34} &
		\multicolumn{1}{c|}{56.80} &
		\multicolumn{1}{c|}{9.05} &
		\multicolumn{1}{c|}{14.56} &
		\multicolumn{1}{c|}{45.15} &
		\multicolumn{1}{c|}{52.69} &
		\multicolumn{1}{c|}{16.99} &
		\multicolumn{1}{c|}{1.94} &
		34.15 \\
		2 &
		\multicolumn{1}{c|}{0.00} &
		\multicolumn{1}{c|}{100} &
		\multicolumn{1}{c|}{100} &
		\multicolumn{1}{c|}{17.44} &
		\multicolumn{1}{c|}{1.85} &
		\multicolumn{1}{c|}{100} &
		\multicolumn{1}{c|}{1.47} &
		\multicolumn{1}{c|}{0.00} &
		\multicolumn{1}{c|}{16.67} &
		80.19 \\
		3 &
		\multicolumn{1}{c|}{75.12} &
		\multicolumn{1}{c|}{62.88} &
		\multicolumn{1}{c|}{83.57} &
		\multicolumn{1}{c|}{14.14} &
		\multicolumn{1}{c|}{11.27} &
		\multicolumn{1}{c|}{38.97} &
		\multicolumn{1}{c|}{88.60} &
		\multicolumn{1}{c|}{19.48} &
		\multicolumn{1}{c|}{48.36} &
		63.62 \\
		4 &
		\multicolumn{1}{c|}{44.3} &
		\multicolumn{1}{c|}{98.73} &
		\multicolumn{1}{c|}{81.01} &
		\multicolumn{1}{c|}{0.00} &
		\multicolumn{1}{c|}{0.00} &
		\multicolumn{1}{c|}{89.87} &
		\multicolumn{1}{c|}{23.33} &
		\multicolumn{1}{c|}{0.00} &
		\multicolumn{1}{c|}{6.33} &
		40.55 \\
		5 &
		\multicolumn{1}{c|}{70.10} &
		\multicolumn{1}{c|}{1.55} &
		\multicolumn{1}{c|}{54.65} &
		\multicolumn{1}{c|}{65.31} &
		\multicolumn{1}{c|}{34.78} &
		\multicolumn{1}{c|}{57.45} &
		\multicolumn{1}{c|}{86.26} &
		\multicolumn{1}{c|}{54.02} &
		\multicolumn{1}{c|}{81.12} &
		69.20 \\
		6 &
		\multicolumn{1}{c|}{35.55} &
		\multicolumn{1}{c|}{41.29} &
		\multicolumn{1}{c|}{2.13} &
		\multicolumn{1}{c|}{6.73} &
		\multicolumn{1}{c|}{8.53} &
		\multicolumn{1}{c|}{53.55} &
		\multicolumn{1}{c|}{97.23} &
		\multicolumn{1}{c|}{8.06} &
		\multicolumn{1}{c|}{5.92} &
		36.02 \\
		7 &
		\multicolumn{1}{c|}{64.95} &
		\multicolumn{1}{c|}{83.62} &
		\multicolumn{1}{c|}{68.42} &
		\multicolumn{1}{c|}{69.03} &
		\multicolumn{1}{c|}{75.47} &
		\multicolumn{1}{c|}{57.48} &
		\multicolumn{1}{c|}{76.03} &
		\multicolumn{1}{c|}{76.23} &
		\multicolumn{1}{c|}{79.11} &
		81.22 \\
		8 &
		\multicolumn{1}{c|}{28.50} &
		\multicolumn{1}{c|}{33.65} &
		\multicolumn{1}{c|}{45.91} &
		\multicolumn{1}{c|}{62.36} &
		\multicolumn{1}{c|}{67.77} &
		\multicolumn{1}{c|}{62.64} &
		\multicolumn{1}{c|}{57.31} &
		\multicolumn{1}{c|}{73.19} &
		\multicolumn{1}{c|}{64.59} &
		57.52 \\
		9 &
		\multicolumn{1}{c|}{70.68} &
		\multicolumn{1}{c|}{79.04} &
		\multicolumn{1}{c|}{22.31} &
		\multicolumn{1}{c|}{41.09} &
		\multicolumn{1}{c|}{54.39} &
		\multicolumn{1}{c|}{80.95} &
		\multicolumn{1}{c|}{36.26} &
		\multicolumn{1}{c|}{75.44} &
		\multicolumn{1}{c|}{38.10} &
		50.38 \\
		10 &
		\multicolumn{1}{c|}{43.27} &
		\multicolumn{1}{c|}{72.51} &
		\multicolumn{1}{c|}{71.74} &
		\multicolumn{1}{c|}{25.90} &
		\multicolumn{1}{c|}{0.22} &
		\multicolumn{1}{c|}{64.24} &
		\multicolumn{1}{c|}{0.00} &
		\multicolumn{1}{c|}{7.06} &
		\multicolumn{1}{c|}{0.00} &
		50.68 \\
		11 &
		\multicolumn{1}{c|}{100} &
		\multicolumn{1}{c|}{100} &
		\multicolumn{1}{c|}{100} &
		\multicolumn{1}{c|}{99.63} &
		\multicolumn{1}{c|}{100} &
		\multicolumn{1}{c|}{99.71} &
		\multicolumn{1}{c|}{90.63} &
		\multicolumn{1}{c|}{100} &
		\multicolumn{1}{c|}{100} &
		100 \\
		12 &
		\multicolumn{1}{c|}{100} &
		\multicolumn{1}{c|}{100} &
		\multicolumn{1}{c|}{100} &
		\multicolumn{1}{c|}{100} &
		\multicolumn{1}{c|}{100} &
		\multicolumn{1}{c|}{100} &
		\multicolumn{1}{c|}{61.56} &
		\multicolumn{1}{c|}{100} &
		\multicolumn{1}{c|}{100} &
		100 \\ \hline
		OA (\%) &
		\multicolumn{1}{c|}{60.31} &
		\multicolumn{1}{c|}{61.66} &
		\multicolumn{1}{c|}{63.61} &
		\multicolumn{1}{c|}{62.71} &
		\multicolumn{1}{c|}{61.95} &
		\multicolumn{1}{c|}{66.60} &
		\multicolumn{1}{c|}{64.65} &
		\multicolumn{1}{c|}{66.96} &
		\multicolumn{1}{c|}{67.92} &
		\textbf{68.89} \\ \hline
		KC ($\kappa$) &
		\multicolumn{1}{c|}{51.89} &
		\multicolumn{1}{c|}{52.68} &
		\multicolumn{1}{c|}{56.02} &
		\multicolumn{1}{c|}{53.12} &
		\multicolumn{1}{c|}{52.84} &
		\multicolumn{1}{c|}{60.58} &
		\multicolumn{1}{c|}{57.75} &
		\multicolumn{1}{c|}{60.79} &
		\multicolumn{1}{c|}{61.77} &
		\textbf{62.70} \\ \hline \hline
	\end{tabular}
	\end{center}
\end{table*}
\begin{figure*}[tp]
	\centering
	\begin{tabular}{cccccccccc}
		\multicolumn{1}{c}{ \epsfig{width=0.15\figurewidth,file=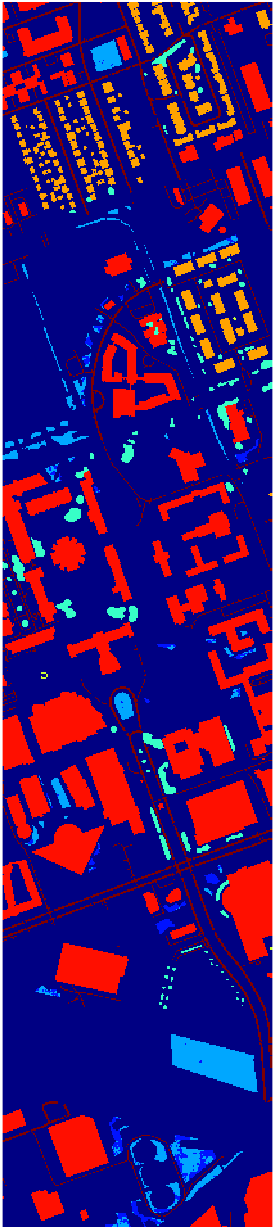}}&
		\multicolumn{1}{c}{
			\epsfig{width=0.15\figurewidth,file=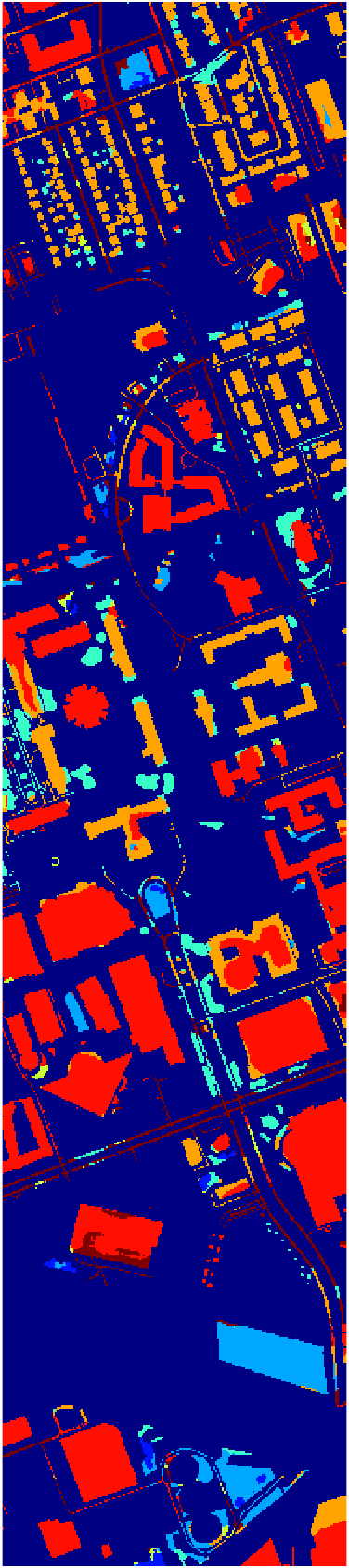}}  &
		\multicolumn{1}{c}{\epsfig{width=0.151\figurewidth,file=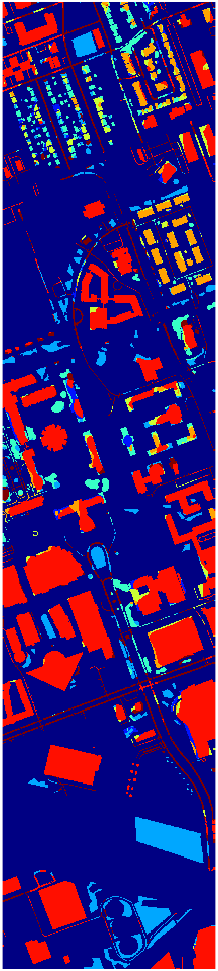}}   &		\multicolumn{1}{c}{\epsfig{width=0.151\figurewidth,file=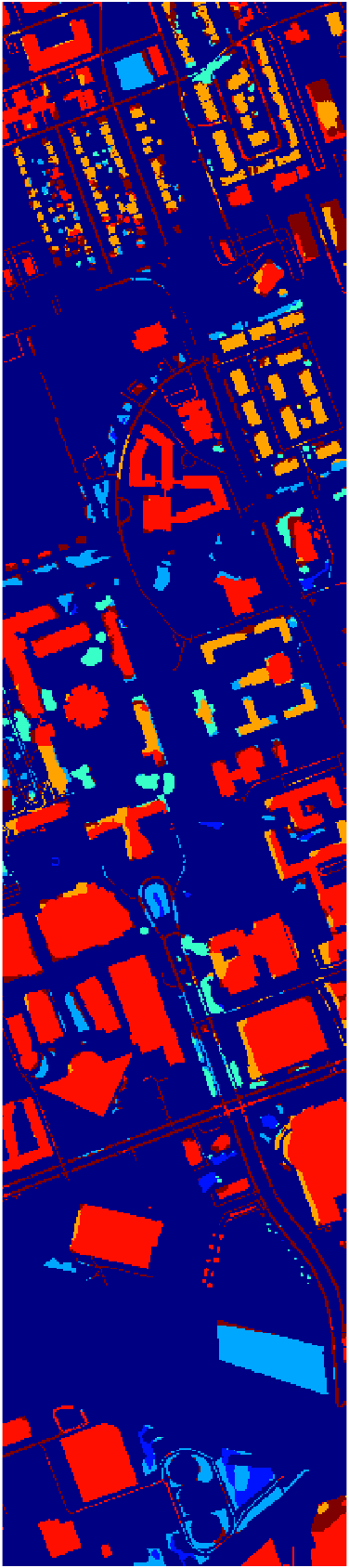}}   &		\multicolumn{1}{c}{\epsfig{width=0.151\figurewidth,file=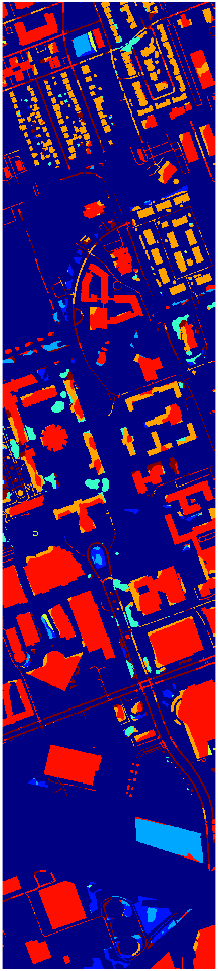}}   &
		\multicolumn{1}{c}{\epsfig{width=0.153\figurewidth,file=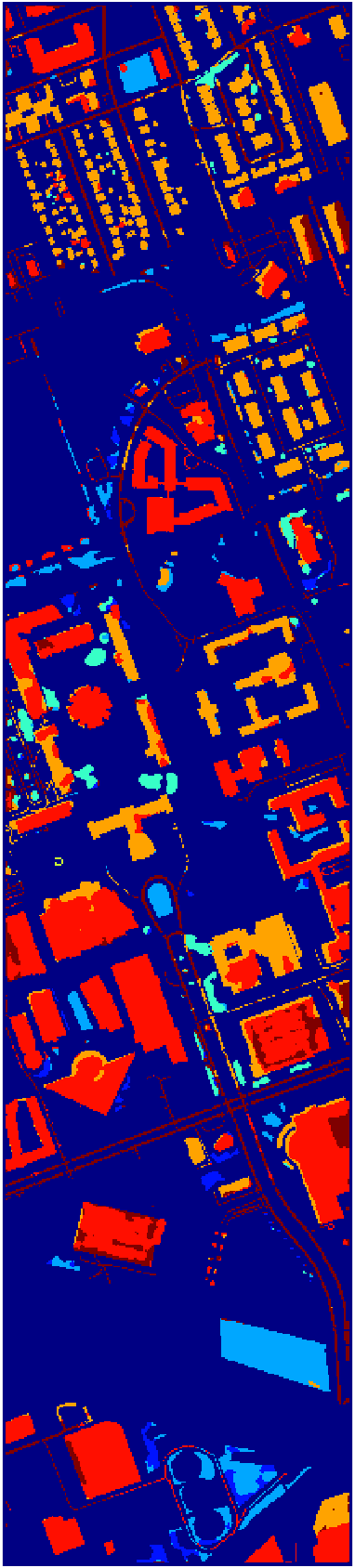}}  & 		\multicolumn{1}{c}{\epsfig{width=0.154\figurewidth,file=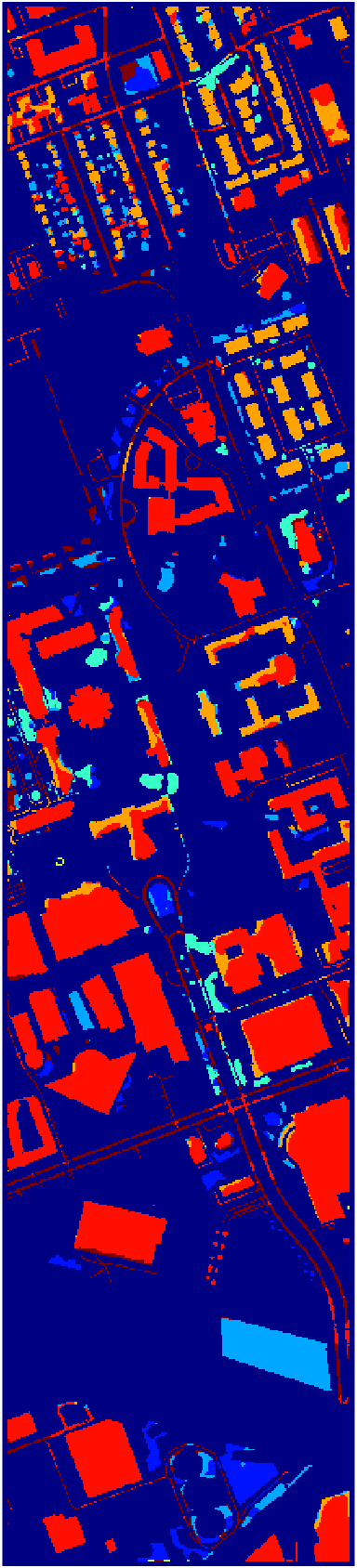}}   &		\multicolumn{1}{c}{\epsfig{width=0.15\figurewidth,file=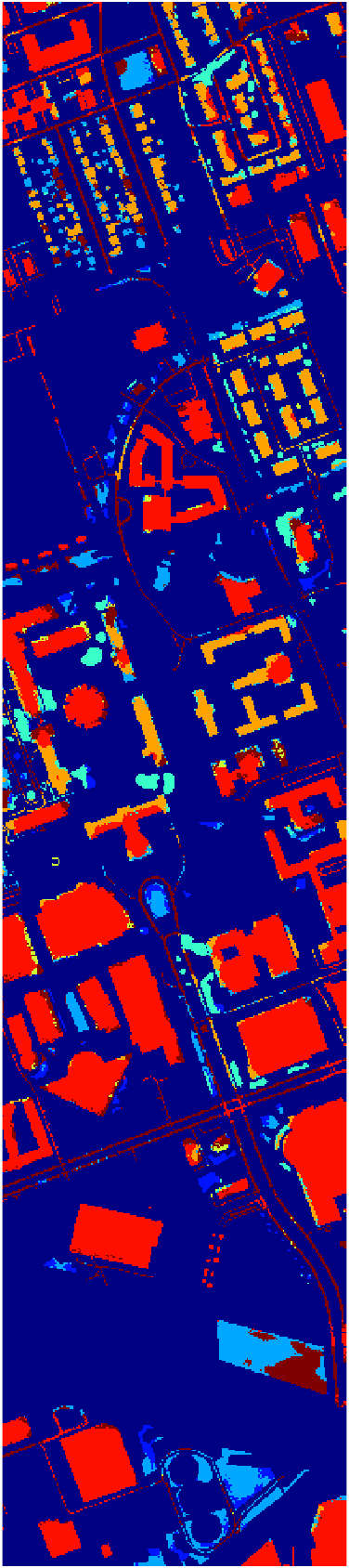}}   &		\multicolumn{1}{c}{\epsfig{width=0.15\figurewidth,file=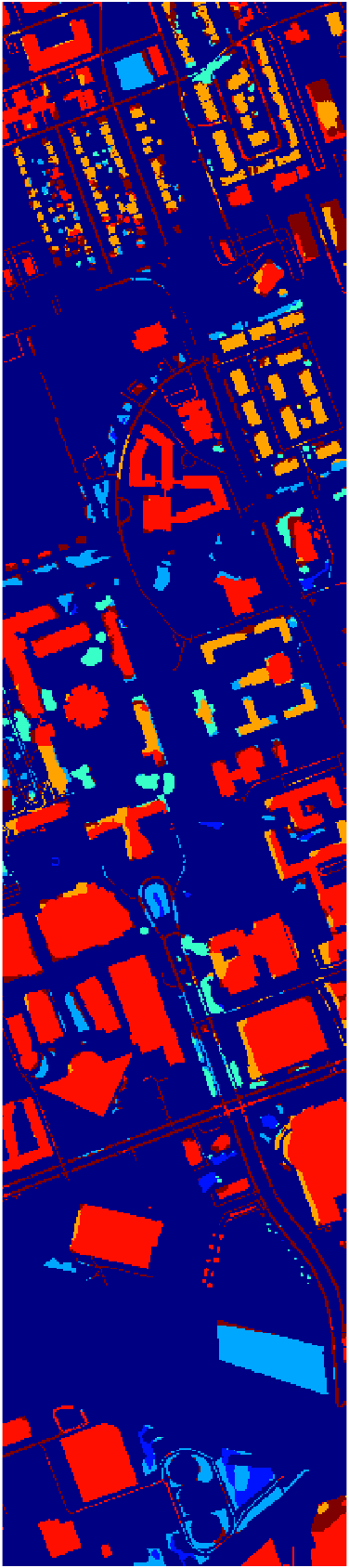}} &
		\multicolumn{1}{c}{\epsfig{width=0.15\figurewidth,file=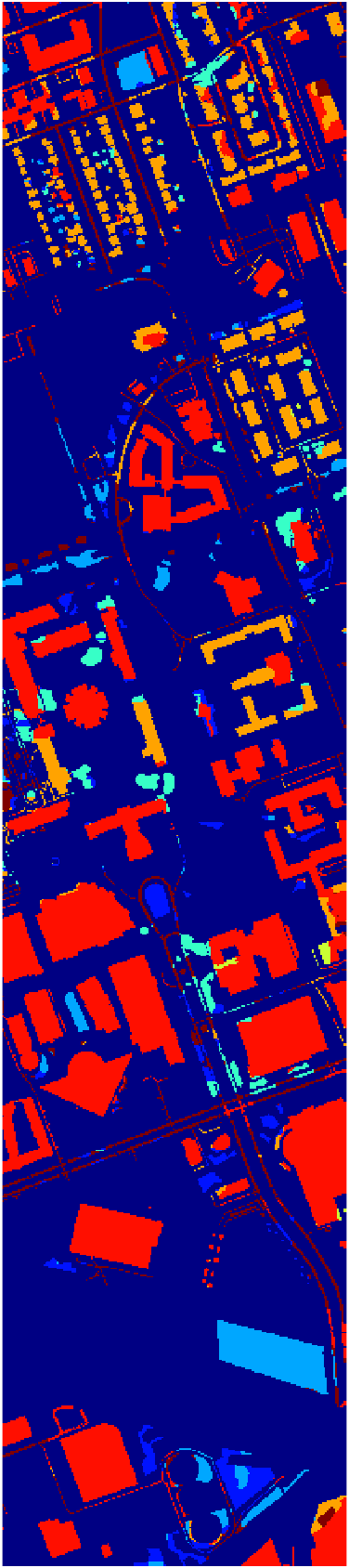}} \\
		\multicolumn{1}{c}{(a)} & \multicolumn{1}{c}{(b)} & \multicolumn{1}{c}{(c)}  & \multicolumn{1}{c}{(d)} & \multicolumn{1}{c}{(e)} & \multicolumn{1}{c}{(f)} & \multicolumn{1}{c}{(g)} & \multicolumn{1}{c}{(h)} & \multicolumn{1}{c}{(i)} &  \multicolumn{1}{c}{(j)}\\
	\end{tabular}
	\begin{tabular}{cc}
		\epsfig{width=1.5\figurewidth,file=Figures/Map/Houston_legend.pdf}\\
	\end{tabular}
	\caption{\label{fig:Map_Houston}
		Visualization and classification maps for the target scene Houston 2018 obtained with different methods including: (a) Ground truth map, (b) GAHT (72.15\%), (c) MLUDA (78.97\%), (d) MSDA (79.41\%), (e) MDGTnet (76.57\%) (f) CLDA (74.0\%), (g) SCLUDA (78.61\%), (h) SSWADA (75.29\%), (i) CACL (79.10\%), (j) BiDA (81.11\%).}
\end{figure*}

\subsection{Performance on MFF cross-temporal airborne dataset}

We verify the effectiveness of the proposed BiDA method for addressing MFF cross-temporal tree species classification. The comparison algorithms used in the experiments include GAHT, MLUDA, MSDA, MDGTnet, CLDA, SCLUDA, SSWADA, and CACL. The optimal base learning rate and regularization parameters of all comparison algorithms are selected from \{$1e-5$, $1e-4$, $1e-3$, $1e-2$, $1e-1$\} and \{$1e-3$, $1e-2$, $1e-1$, $1e+0$, $1e+1$, $1e+2$\}, respectively, and using cross-validation to find the corresponding optimal parameters. Tables \ref{tab:accuracy_SD2-TD1}-\ref{tab:accuracy_SD2-TD2} report the class-specific accuracy (CA), overall accuracy (OA) and Kappa coefficient (KC) of the above methods in the MFF TD1 and MFF TD2, and the analysis is as follows.

\begin{itemize}
	
	\item In transformer-based classification methods without DA strategies, GAHT performs the best, with OAs of 67.38\% and 68.61\% on MMF TD1 and MMF TD2, respectively. GAHT outperforms most unsupervised deep domain adaptation methods, including MLUDA, MSDA, SCLUDA, SSWADA, and CACL, where these methods are based on CNN as the baseline. It shows that a transformer-based framework designed for fine-grained tree species recognition is superior to CNN-based methods.
	
	\item Compared to GAHT, BiDA achieves improvements of 10.02\% and 6.47\% on MMF TD1 and MMF TD2, respectively. Without any domain alignment strategies in BiDA, referring to Table \ref{table:Ablation}, BiDA outperforms GAHT by 1.44\% and 0.5\%. It indicates that the transformer architecture consisting of semantic tokenizer, source branch, target branch, and coupled branch in BiDA is superior to GAHT constructed in a hierarchical manner. Furthermore, the overall architecture of BiDA combined with domain alignment strategies and ARS significantly outperforms single-scene classification methods based on transformers.
	
	\item Compared to MLUDA and MSDA, the best performing unsupervised deep domain adaptation methods on the MFF cross-temporal airborne dataset, BiDA achieves higher classification accuracy by 4.6\% and 2.8\% on MMF TD1 and MMF TD2, respectively. TThe main idea of MSDA is to design masked self-distillation-based DA strategies to generate pseudo-labels for target domain samples and learning shared adaptive space. This indicates that BiDA learns domain-invariant features in independent adaptive spaces, significantly better than MSDA and other DA methods in a shared space.
\end{itemize}
For further visual comparison, the classification maps of MMF TD1 and MMF TD2 are illustrated in Figs. \ref{fig:Map_TD1}-\ref{fig:Map_TD2}. We utilized the trained models to predict all pixels in MMF TD1 and MMF TD2. The classification maps generated by most of the comparison methods exhibit significant noise and large prediction errors. Particularly, in certain classes such as 3-th class (Korean pine), 4-th class (spruce), and 5-th class (Broad-leaved trees), the classification results show significant fragmentation, leading to inaccurate and disjointed identification of tree species regions. As illustrated in Fig. \ref{fig:Map_TD2}, especially in MMF TD2, the large area of 5-th class (Broad-leaved trees) is predicted to be striped, such as GAHT, MLUDA, SCLUDA, etc., where MLUDA and SCLUDA incorrectly predict large areas of 5-th class. The noise and errors in the predictions are primarily due to the large spatial and temporal span of the hyperspectral airborne imagery, resulting in significant spectral shifts of the same tree species across multiple flight lines. Consequently, under severe spectral shifts, the shared feature extraction and DA strategies of most methods exhibit significant limitations, as they fail to adequately capture subtle differences between different tree species, leading to inaccurate and unstable classification results. The classification maps of the proposed BiDA method, as shown in Fig. \ref{fig:Map_TD1}(h) and Fig. \ref{fig:Map_TD2}(h), demonstrate minimal influence from spectral shifts between flight lines, achieving regionally coherent and accurate tree species classification.

In cross-domain classification tasks, the most critical concern is how to effectively alleviate spectral shift. To further analyze the performance of BiDA in alleviating spectral shift and improving class separability, we output the original samples of the MFF SD, MFF TD1, and MFF TD2, as well as the aligned features obtained through BiDA. In Fig. \ref{fig:analysis}, we present two-dimensional distribution visualizations of the MFF SD \& MFF TD1 and the MFF SD \& MFF TD2, where the $\bullet$ represent SD, × represents TD1/TD2, and the numbers represent classes. From Fig. \ref{fig:analysis} (a) and (c), it can be observed that there is a distinct spectral shift between the MFF SD and MFF TD1/MFF TD2, such as the 1-th class (red $\bullet$ and ×) and the 3-th class (green $\bullet$ and ×), which exhibit a clear gap in distribution. Furthermore, observing the intra-domain samples (only considering $\bullet$ or ×), it can be seen that intra-domain samples of each class overlap significantly, indicating poor class separability, for example, the 1-th class (red ×), the 2-th class (yellow ×), and the 5-th class (dark blue ×). Through the transfer learning scheme in BiDA, SD and TD1/TD2 are projected into the adaptive space, as shown in Fig. \ref{fig:analysis} (d) and (h). Observing inter-domain samples of the same class, most same class samples from SD and TD1/TD2 are clustered together, especially the 1-th class (red $\bullet$ and ×) and the 4-th class (light blue $\bullet$ and ×). Furthermore, the inter-class separability of SD or TD1/TD2 is significantly improved (i.e., intra-domain same class samples are compactly clustered, while samples of different classes are discretely distributed). This indicates that the proposed ARS encourages the source branch and the target branch to capture intra-domain generalized features, better adapting to the intra-domain internal structure.

\subsection{Performance on cross-scene/temporal satellite dataset}

To verify the applicability of BiDA to satellite datasets, we conducted comparative experiments using Houston and HyRANK satellite datasets. Houston dataset contains 7 coarse-grained land cover classes. HyRANK dataset comprises 12 classes, in addition to typical land cover classes, it also includes multiple tree species, such as Fruit Trees, Olive Groves, Coniferous Forest and Sderophyllous Vegetation. Compared to the MFF cross-temporal airborne dataset, it is more difficult to achieve high-precision cross-scene interpretation on HyRANK.

The CA, OA and KC of the tested methods on the Houston 2018 data and Loukia data are presented in Tables \ref{tab:accuracy_Hou}-\ref{tab:accuracy_HyRANK}. All methods incorporating DA strategies outperform single-scene classification methods GAHT. Compared with better performance methods TSTnet and CACL, BiDA demonstrates an improvement of 1.85\% and 0.77\% in transfer performance on the Houston 2018 data and Loukia data. The classification maps generated by all methods for the Houston 2018 are displayed in Figs. \ref{fig:Map_Houston}. Compared to the ground truth maps in Fig. \ref{fig:Map_Houston} (a), BiDA achieves more accurate predictions with less noise in multiple regions.

%

%
\section{Conclusions}
\label{sec:conclusions}
In this paper, the Bi-directional Domain Adaptation (BiDA) for cross-scene/temporal hyperspectral image (HSI) classification is proposed. Firstly, a transformer encoder comprising source branch, target branch and coupled branch is constructed, and based on the characteristics of HSI, a semantic tokenizer suitable for the transformer structure is designed. Specifically, the Coupled Multi-head Cross-attention (CMCA) mechanism is devised for bi-directional feature alignment. Furthermore, a bi-directional distillation loss is introduced to guide the bi-directional supervision training for the independent adaptive space learning of source and target branches. Lastly, an Adaptability Reinforcement Strategy (ARS) is proposed to address a problem of overlooking the extraction of domain-specific generalized features within the target domain. Extensive experimental results demonstrate that the proposed BiDA outperforms some state-of-the-art domain adaptation approaches across three cross-temporal/scene airborne and satellite datasets. In the cross-temporal tree species classification task, the proposed BiDA is more than 3\%$\sim$5\% higher than the most advanced method.

\bibliographystyle{IEEEtran}
\bibliography{bibfile_zyx}

\end{document}